\pdfoutput=1
\documentclass[twocolumn]{next++}

\usepackage{amsmath,amssymb,amsthm}
\usepackage{tabularx,colortbl,stfloats,needspace}
\usepackage{algorithm,algorithmic,float}
\newtheorem{lemma}{Lemma}[section]
\newtheorem{proposition}{Proposition}[section]
\newtheorem{corollary}{Corollary}[section]

\newcolumntype{Y}{>{\centering\arraybackslash}X}
\definecolor{resultshade}{RGB}{235,244,249}
\definecolor{cvprblue}{rgb}{0.21,0.49,0.74}
\definecolor{deltaneg}{rgb}{0.78,0.22,0.18}
\makeatletter
\newcommand{\fullrowband}[1]{%
  \noalign{\begingroup\color{#1}%
    \hrule width\linewidth height\dimexpr\ht\@arstrutbox+\dp\@arstrutbox\relax
    \kern-\dimexpr\ht\@arstrutbox+\dp\@arstrutbox\relax
  \endgroup}}
\makeatother

\AtBeginEnvironment{table}{\renewcommand{\textbf}[1]{{\bfseries #1}}}
\AtBeginEnvironment{table*}{\renewcommand{\textbf}[1]{{\bfseries #1}}}

\title{PRAM: Activating Latent Multimodal Knowledge through Partwise Reasoning}

\author[1,6]{Haidong Chen}
\author[1]{Ying Guo}
\author[2]{Shangze Li}
\author[3,6]{Zhongkuan Mao}
\author[4]{\\Linrui Xu}
\author[5]{Canran Xiao}
\author[6,\dagger]{Fei Shen}
\author[6]{Tat-Seng Chua}

\affiliation[1]{North China University of Technology}
\affiliation[2]{Nanjing University of Science and Technology}
\affiliation[3]{Sichuan University}
\affiliation[4]{Central South University}
\affiliation[5]{Sun Yat-sen University}
\affiliation[6]{National University of Singapore, NExT++ Research Centre}

\abstract{
Emergent cross-modal classification and retrieval enable transfer between modalities without direct paired supervision.
However, existing methods often prioritize sample-level alignment, overlooking local discriminative information and partwise statistical relations.
To this end, we propose partwise reasoning across modalities (PRAM), which activates latent multimodal knowledge in frozen encoders by composing partwise statistical relations through a shared pivot modality.  
Specifically, we first present a multi-granularity covariance regularization module that estimates covariances from global and local vectors in two collections of modality--pivot pairs, stabilizing local estimates through adaptive blending with their cross-part means. 
Second, we devise a pivot-based spectral relational reasoning module that infers global and partwise statistical relations between target modalities through regularized covariance composition and extracts paired projection directions and their correlation strengths via whitening and spectral decomposition. 
Finally, we introduce a multi-granularity score fusion module that standardizes and fuses matching scores from global and local relations, incorporating task-specific information on pivot associations and class semantics for class prediction and candidate ranking. 
Extensive experiments across ten datasets demonstrate that PRAM achieves state-of-the-art performance in 16 classification settings and 19 retrieval settings, while exhibiting highly competitive performance in three sensing scenarios.
\vspace{-0.1cm}
}

\checkdata[Resources]{
  \href{https://luo1yan.github.io/PRAM/}{\raisebox{-1ex}{\includegraphics[height=4.2ex]{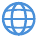}}~Project Page}\hspace{1em}
  \href{https://github.com/luo1yan/PRAM}{\raisebox{-1ex}{\includegraphics[height=4.2ex]{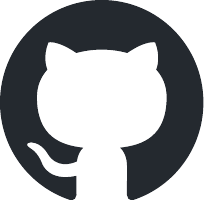}}~Code}\hspace{1em}
  \href{https://huggingface.co/datasets/Rowance/PRAM-features}{\raisebox{-1ex}{\includegraphics[height=4.2ex]{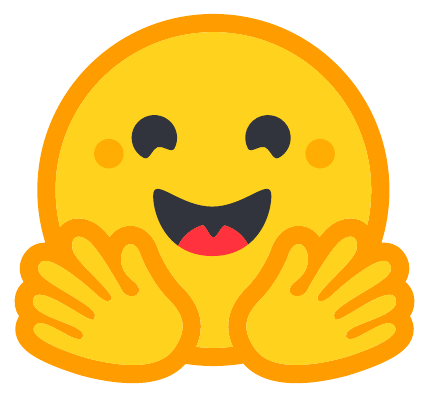}}~Features}
  \vspace{0.2cm}
}

\checkdata[Date]{
  November 25, 2026
}

\begin{document}
\maketitle

\begingroup
  \renewcommand{\thefootnote}{\fnsymbol{footnote}}
  \footnotetext[2]{Corresponding Author (\href{mailto:shenfei29@nus.edu.sg}{shenfei29@nus.edu.sg}).}
\endgroup

\begin{figure}[t]
\centering
\includegraphics[width=0.98\linewidth]{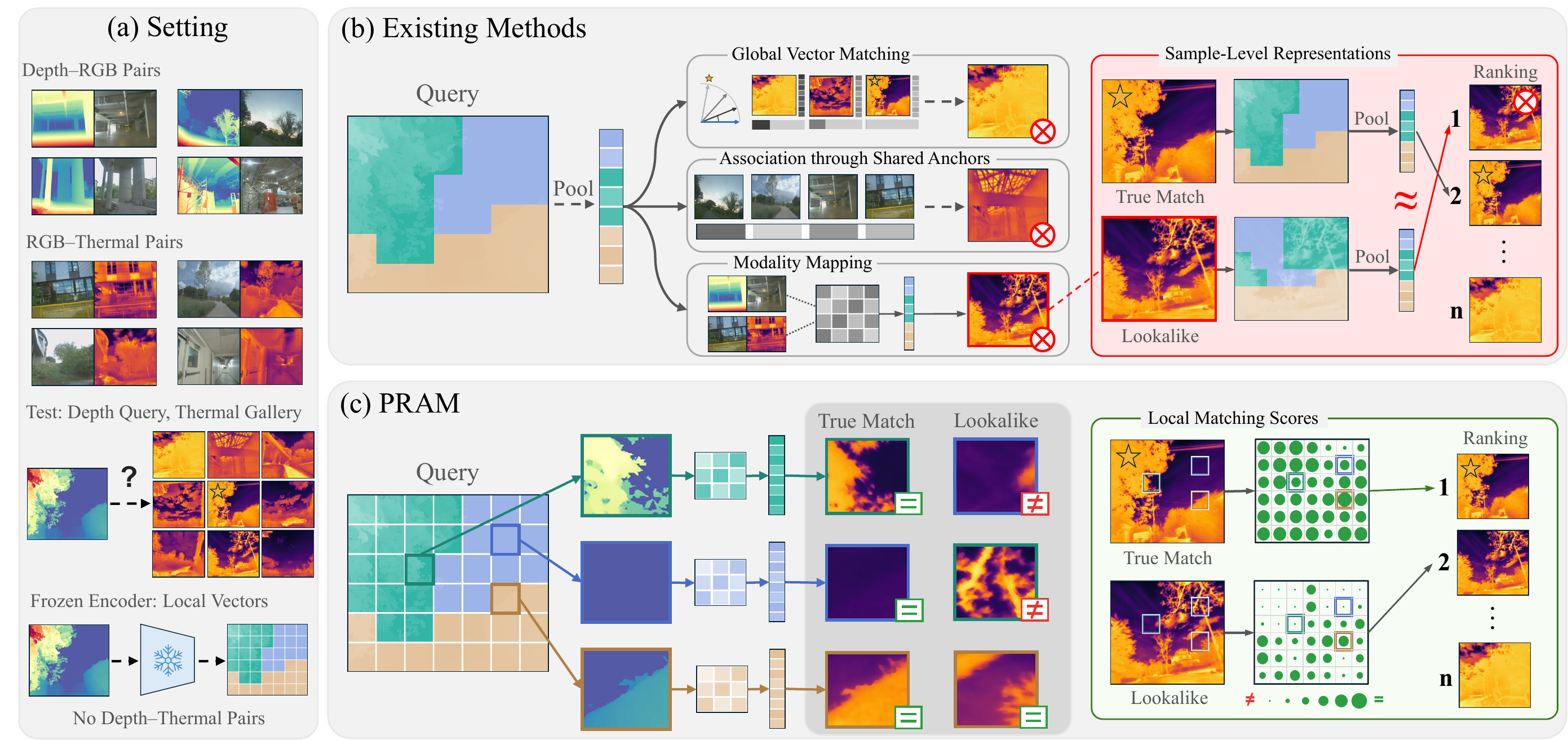}
\vspace{-0.2cm}
\caption{\textbf{Partwise Reasoning Activates Latent Multimodal Knowledge.}
(a) Separate modality--pivot collections connect target modalities without direct pairing.
(b) Sample-level matching can rank a lookalike above the true match.
(c) PRAM composes relations through corresponding pivot parts and fuses global and local matching scores to rank the true match first.}
\label{fig:teaser}
\end{figure}

\section{Introduction}
\label{sec:introduction}

Cross-modal classification and retrieval~\cite{clip,align,siglip} connect modalities to exploit complementary information for class prediction and sample matching.
Emergent cross-modal classification~\cite{imagebind} and retrieval~\cite{emergentbridge} extend these capabilities through existing modality--pivot pairs without direct pairing between target modalities (Figure~\ref{fig:teaser} (a)).
Reusing these pairs and knowledge from pretrained models helps reduce additional paired data requirements, supporting multimodal applications on robots~\cite{anyloc}, vehicles~\cite{liploc}, and wearable devices~\cite{imu_video_ood} under data constraints.

Existing research~\cite{omnibind,ucl,mmbind,emergentbridge} on emergent cross-modal classification and retrieval primarily relies on a shared pivot modality for modality alignment and pretrained representations for cross-modal relation modeling.
For example, ImageBind~\cite{imagebind} and LanguageBind~\cite{languagebind} use image and language as their respective shared pivot modalities, learning modality alignment from modality--pivot pairs and demonstrating emergent retrieval between target modalities without direct paired supervision.
Building on these pretrained representations, related methods~\cite{latenttranslation,exmcr,structure} compute matching scores between target modalities through global vector matching~\cite{dbsn}, association through shared anchors~\cite{asif}, or modality mapping~\cite{textme}, as illustrated in Figure~\ref{fig:teaser}~(b).
MaskCLIP~\cite{maskclip} further shows that frozen CLIP features support local semantic prediction beyond standard image-level outputs, providing evidence for exploiting local semantic information in pretrained encoders.
We refer to information that is already present in frozen encoders yet remains incompletely expressed in their standard outputs as latent multimodal knowledge.
However, sample-level cross-modal relation modeling often neglects to preserve local discriminative information and compose partwise statistical relations, limiting the use of latent multimodal knowledge for classification and retrieval.

Just as a friend of a friend is not necessarily a friend, knowing that entities are connected does not mean knowing how they are related.
For emergent cross-modal tasks, statistical relations for both modality--pivot pairs do not guarantee full recovery of partwise statistical relations between target modalities.
Relation recovery depends on preserving local discriminative information in the pivot representation and maintaining correspondence between partwise statistical relations on both sides.
Therefore, representation granularity and pivot capacity limit which partwise statistical relations can be recovered and how effectively latent multimodal knowledge supports classification and retrieval.

For that, we propose partwise reasoning across modalities (PRAM), which activates latent multimodal knowledge in frozen encoders by composing partwise statistical relations through a shared pivot modality to support classification and retrieval, from Figure~\ref{fig:teaser}~(c).
First, we present a multi-granularity covariance regularization (MCR) module that estimates covariances from global and local vectors in two collections of modality--pivot pairs and adaptively blends local estimates with cross-part means per modality pair, improving stability with limited samples.
Second, we devise a pivot-based spectral relational reasoning (PSRR) module inferring global and partwise statistical relations between modalities via regularized covariance composition globally and at corresponding pivot parts, extracting paired projection directions and correlation strengths through whitening and spectral decomposition.
Finally, we introduce a multi-granularity score fusion (MSF) module that standardizes and fuses global and local matching scores derived from projection responses and correlation strengths.
For classification, these scores are augmented with pivot associations and class semantics as applicable; for retrieval, they rank candidates.
Throughout this process, encoders remain frozen, and relation estimation and score standardization use only modality--pivot pairs, without any direct pairs between target modalities.
Experiments across ten datasets using ImageBind and LanguageBind, together with evaluations in robot localization, cross-device vehicle localization, and wearable action recognition, demonstrate PRAM's effectiveness and competitive performance.
The main contributions are summarized as follows:
\begin{itemize}
    \item We propose PRAM for emergent cross-modal classification and retrieval between unpaired modalities, activating frozen encoders' latent multimodal knowledge by composing partwise statistical relations via a shared pivot.

    \item We devise MCR, PSRR, and MSF to stabilize covariance estimates, infer partwise statistical relations, and fuse global and local matching scores, using only modality--pivot pairs to estimate relations and standardize scores.

    \item Across ImageBind and LanguageBind, PRAM achieves the best results in 16 of 17 classification and 19 of 20 retrieval settings across ten datasets, while demonstrating its effectiveness in three sensing scenarios.
\end{itemize}

\section{Related Work}
\label{sec:related}

\noindent\textbf{Multimodal Alignment.}
Contrastive image--text pretraining~\cite{clip} and locked-encoder alignment~\cite{lit} establish shared multimodal representations, while ImageBind~\cite{imagebind} and LanguageBind~\cite{languagebind} extend modality alignment through image and language pivots.
FreeBind~\cite{freebind} and OmniBind~\cite{omnibind} integrate pretrained spaces; UniBind~\cite{unibind} introduces modality-agnostic alignment centers, and Point-Bind~\cite{pointbind} connects point clouds to the shared space.
Other approaches connect pretrained spaces through shared anchors~\cite{asif}, semantic correspondences~\cite{latenttranslation}, overlapping modalities~\cite{exmcr}, text descriptions~\cite{textme}, relational geometry~\cite{gromov}, or sample neighborhoods~\cite{structure}; CSA~\cite{csa} aligns unimodal features via canonical correlations from limited pairs.
Representation-similarity analyses~\cite{platonic} characterize geometric agreement between independently learned spaces.
Further strategies synthesize bridge embeddings~\cite{emergentbridge}, infer connections through intermediate variables~\cite{ucl}, learn modality pairs sequentially~\cite{cmcl}, or construct pseudo-pairs through shared modalities~\cite{mmbind}; Bridge CorrNet~\cite{bridgecorrnet} correlates unpaired views via a shared pivot.
However, formulations that rely on sample-level representations can obscure spatial or temporal distinctions, limiting the use of local discriminative information in emergent cross-modal classification and retrieval.

\noindent\textbf{Local Relation Modeling.}
Vision Transformers~\cite{vit} and self-supervised encoders~\cite{dinov2} provide spatial features, while dense correspondence learning~\cite{densecl}, region alignment~\cite{regionclip}, and semantic grouping~\cite{groupvit} support local representation learning.
Image-level supervision also supports dense prediction~\cite{maskclip} and patch--language alignment~\cite{dinotxt}.
Local cross-modal interactions~\cite{denseav} and token-level late interaction~\cite{colbert} enable fine-grained matching.
For statistical relation estimation, canonical correlation analysis~\cite{hotelling} and its probabilistic~\cite{probcca} and deep~\cite{dcca} variants identify shared coordinates, while partial least squares~\cite{plsr} emphasizes cross-covariance.
Covariance shrinkage~\cite{ledoitwolf} and ridge regularization~\cite{hoerl} stabilize estimation and inversion, respectively.
Hubness~\cite{hubness} motivates local scaling~\cite{csls} and dual-bank normalization~\cite{dbsn}, and embedding standardization~\cite{i0t} reduces the CLIP modality gap.
However, these methods do not directly address how to stably compose partwise statistical relations from separate collections of modality--pivot pairs and fuse global and local matching scores with frozen encoders for classification and retrieval.

\begin{figure*}[t]
\centering
\includegraphics[width=0.98\linewidth]{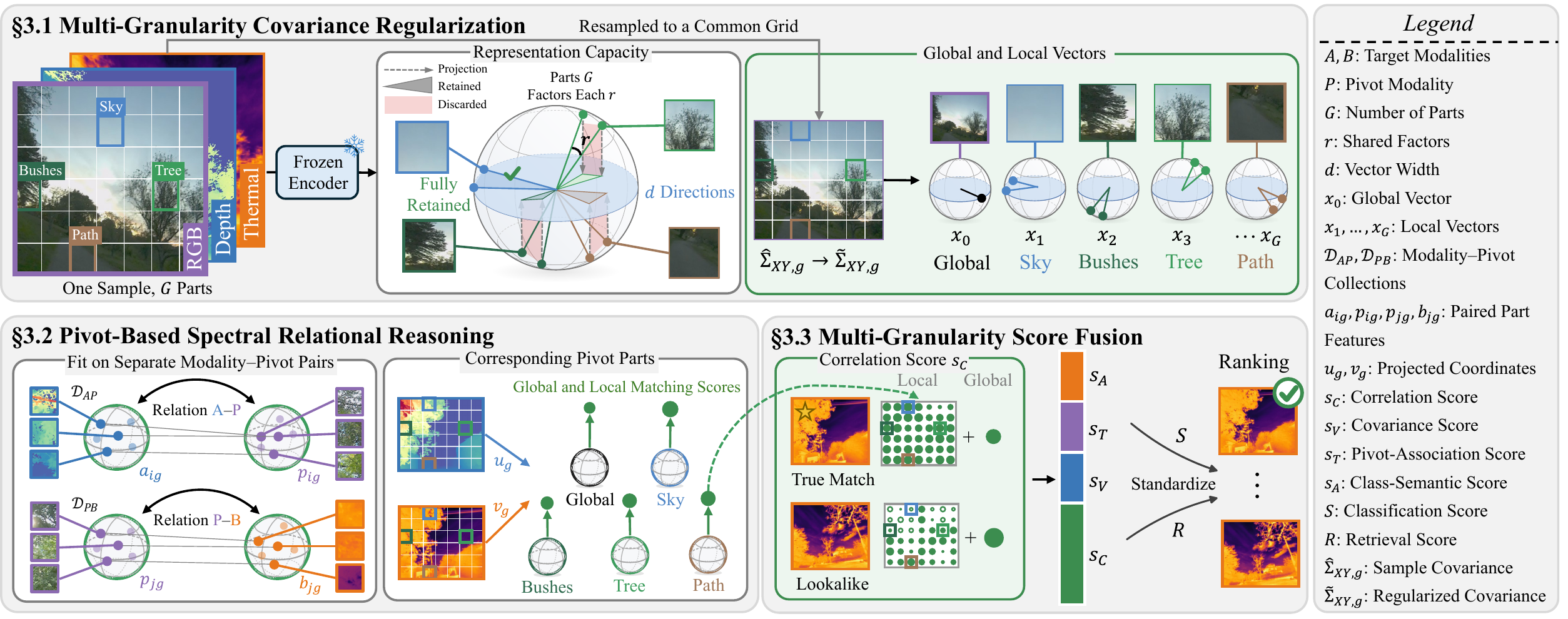}
\vspace{-0.27cm}
\caption{\textbf{Overview of PRAM.}
MCR retains global and local vectors $x_0,\ldots,x_G$ and stabilizes covariance estimation.
Using separate modality--pivot pairs, PSRR composes $A$--$P$ and $P$--$B$ relations globally and through corresponding pivot parts, yielding matching scores.
MSF aggregates global and local scores and fuses standardized correlation and covariance scores for retrieval ($R$), incorporating pivot-association and class-semantic scores for classification ($S$).
Filled and hollow dots denote positive and negative matching scores.}
\label{fig:method}
\end{figure*}

\section{Method}
\label{sec:method}
\noindent\textbf{Overview.}
Given two separately provided collections $\mathcal D_{AP}=\{(a_i,p_i)\}_{i=1}^{n_A}$ and $\mathcal D_{PB}=\{(p_j,b_j)\}_{j=1}^{n_B}$, PRAM infers statistical relations between target modalities $A$ and $B$ for classification and retrieval using frozen encoders, without direct $A$--$B$ paired supervision.
The collections define the observed edges $E_A=AP$ and $E_B=PB$, with $A,B$ as endpoints and $P$ as the shared pivot modality.
As shown in Figure~\ref{fig:method}, MCR (Sec.~\ref{sec:mcr}) retains a global vector and local vectors for corresponding parts of each sample and stabilizes covariance estimates on both edges.
PSRR (Sec.~\ref{sec:psrr}) composes these covariances at the global level and at corresponding pivot parts, extracting paired projection directions and correlation strengths to compute global and local matching scores.
MSF (Sec.~\ref{sec:msf}) aggregates these matching scores, then standardizes and fuses the correlation and covariance scores.
For classification, it further incorporates pivot associations and class semantics as applicable; for retrieval, it ranks candidates using the fused scores.
Algorithm~\ref{alg:pram} summarizes the complete procedure.

\subsection{Multi-Granularity Covariance Regularization}
\label{sec:mcr}
MCR retains global and local vectors and stabilizes covariance estimation from limited modality--pivot pairs.

\noindent\textbf{Multi-Granularity Representation.}
Binding models~\cite{imagebind,languagebind} compare one width-$d$ global vector $x_0$ per sample across modalities.
Such global representations can obscure local discriminative information (Figure~\ref{fig:method}).
To characterize this limitation, consider an idealized linear--Gaussian model with $G$ independent parts, each containing $r$ shared factors, and independent, uniformly oriented projection spaces.
A width-$d$ global vector and one width-$d$ local vector per part retain the fractions
\begin{equation}
\eta_{\rm global}=\min\!\left\{1,\frac{d}{Gr}\right\},\qquad
\eta_{\rm part}=\min\!\left\{1,\frac{d}{r}\right\}.
\label{eq:capacity}
\end{equation}
At equal width per vector and $r\leq d<Gr$, the global vector retains $d/(Gr)$, whereas local vectors fully retain each part.
Their content and the population cross-modal relations derived from them retain these fractions under this model (proof in Appendix~\ref{sec:app-capacity}).
MCR therefore retains $x_0$ and local vectors $x_1,\ldots,x_G$ already computed by frozen encoders.
Spatial features are resampled to a common grid using registered coordinates; temporal segments use synchronized intervals, and single-vector modalities broadcast their vector across parts.
The resulting paired features $(a_{ig},p_{ig})$ and $(p_{jg},b_{jg})$ preserve within-sample distinctions and link each target part to its corresponding pivot part.

\noindent\textbf{Covariance Regularization.}
On each edge, MCR centers the features of every part and divides each modality by scalar training standard deviations, computed on its own edge for the endpoints and on both collections for the pivot.
To stabilize estimation with limited pairs, MCR adaptively blends each local sample covariance block with its cross-part mean on the same edge ($X,Y\in\{A,P\}$ or $X,Y\in\{P,B\}$), following covariance shrinkage~\cite{ledoitwolf}:
\begin{equation}
\widetilde\Sigma_{XY,g}
=(1-\alpha)\widehat\Sigma_{XY,g}
+\frac{\alpha}{G}\sum_{h=1}^{G}\widehat\Sigma_{XY,h}.
\label{eq:shrink}
\end{equation}
MCR sets $\alpha$ to the clipped ratio of the part blocks' estimated sampling variance to their squared deviation from the cross-part mean, pooled over both edges (Appendix~\ref{sec:app-covariance}); the global component uses $\alpha=0$.

\begin{algorithm}[t]
\caption{Partwise Reasoning across Modalities}
\label{alg:pram}
\footnotesize
\begin{algorithmic}[1]
\REQUIRE Separate $\mathcal D_{AP},\mathcal D_{PB}$; frozen encoders; task protocol, query, and candidates (labels if needed).
\ENSURE Class prediction or retrieval ranking.

\item[] \textit{Fitting: modality--pivot pairs only; no direct $A$--$B$ supervision.}
\STATE \textbf{MCR:} Extract global/local vectors; center parts per edge, scale endpoints per edge and the pivot on both edges, and fit sample covariances $\widehat\Sigma$.
\STATE Estimate $\alpha$ from both edges; obtain $\widetilde\Sigma$ by Eq.~\ref{eq:shrink} ($\alpha=0$ for $g=0$).
\STATE \textbf{PSRR:} Count pivot states $m$; compute whitening maps and retained $(U_g,\sigma_g,V_g)$ for $g=0,\ldots,G$ (Eq.~\ref{eq:compose}).
\STATE \textbf{MSF:} Identify the shared support $\mathcal K_\cap$; set $\omega$ and $w,o$ (Eq.~\ref{eq:gates}).
\STATE Build $\phi_{X,g}$ from unshrunk moments (Eq.~\ref{eq:covmap}); fit $s_T$ posteriors for classification if $o>0$.
\STATE Set $\tau_A=0$; for class-prompt classification, fit the $s_A$ route and set $\tau_A=w_A\chi_A$.
\STATE Select training reference banks $\mathcal B_A,\mathcal B_B$; cache overall moments of active $s_T,s_A$ on $\mathcal B_A\times\mathcal B_B$.

\item[] \textit{Pair scoring: fitted quantities fixed; evaluate active components only.}
\FOR{each query--candidate pair $(a,b)$ in modality order $(A,B)$}
    \STATE Encode and normalize $(a,b)$; compute the required projections and $\ell_g$ (Eq.~\ref{eq:llr}).
    \STATE Form $s_C=\ell_0+\omega\sum_{g=1}^G\ell_g$; compute $s_V$ as mean cosine in pivot coordinates over $g=0,\ldots,G$ (Eq.~\ref{eq:covmap}).
    \STATE For classification, evaluate active $s_T,s_A$ in the form of Eq.~\ref{eq:state}.
    \STATE Standardize $s_C,s_V$ by Eq.~\ref{eq:bank}; obtain active $\overline{\mathcal Z}_T,\overline{\mathcal Z}_A$ using cached overall moments.
    \STATE Compute the classification score $S$ or retrieval score $R$ by Eq.~\ref{eq:readouts}.
\ENDFOR

\item[] \textit{Task decisions.}
\STATE Return the class of the highest-$S$ prompt, the majority label of the ten highest-$S$ labeled candidates, or the $R$ ranking, following the protocol.
\end{algorithmic}
\end{algorithm}

\subsection{Pivot-Based Spectral Relational Reasoning}
\label{sec:psrr}
PSRR infers global and partwise statistical relations between $A$ and $B$ from two observed edges and scores pairs.

\noindent\textbf{Relation Composition.}
PSRR composes regularized $A$--$P$ and $P$--$B$ covariances globally ($g=0$) and through corresponding pivot parts ($g=1,\ldots,G$).
Let $H_g=\tfrac12(\widetilde\Sigma_{PP,g}^{(AP)}+\widetilde\Sigma_{PP,g}^{(PB)})$.
For $X\in\{A,B\}$, composition, whitening, and spectral decomposition give
\begin{equation}
\begin{aligned}
K_g&=\widetilde\Sigma_{AP,g}(H_g+\rho_{P,g}I)^{-1}
                   \widetilde\Sigma_{PB,g},\\
W_{X,g}&=(\widetilde\Sigma_{XX,g}+\rho_{X,g}I)^{-1/2},\\
W_{A,g}K_gW_{B,g}
&=U_g\,\operatorname{diag}(\sigma_g)\,V_g^\top.
\end{aligned}
\label{eq:compose}
\end{equation}
We use ridge regularization~\cite{hoerl} with $\rho_{P,g}=\rho(H_g)$ and $\rho_{X,g}=\rho(\widetilde\Sigma_{XX,g})$, where $\rho(M)=\lambda\max\{\operatorname{tr}(M)/\dim(M),10^{-12}\}$ and $\lambda=0.5$.
In the whitened feature spaces, the columns of $U_g$ and $V_g$ define paired projection directions, and $\sigma_g$ contains their correlation strengths.
Retain the $k_g\leq256$ leading directions, subject to the additional single-vector-pivot rank cap specified in Appendix~\ref{sec:estimator-details}.
The projected coordinates are $u_g=U_g^\top W_{A,g}\bar a_g$ and $v_g=V_g^\top W_{B,g}\bar b_g$, where bars denote features centered and scaled in MCR.

\noindent\textbf{Matching Scores.}
For jointly Gaussian part feature vectors $A_g,B_g,P_g$ with $A_g\perp B_g\mid P_g$, population whitening, and no regularization, nonsingular retained coordinates form independent correlated pairs whose log-likelihood ratios add.
For regularized frozen features, we apply this form as a composite matching score~\cite{composite}:
\begin{equation}
\ell_g(a,b)=\sum_{k=1}^{k_g}
\frac{c_{gk}u_{gk}v_{gk}
-\tfrac12c_{gk}^{2}(u_{gk}^{2}+v_{gk}^{2})}
{1-c_{gk}^{2}},
\label{eq:llr}
\end{equation}
where $c_{gk}=\operatorname{clip}(\sigma_{gk},0,1-10^{-6})$; the omitted determinant term is constant across pairs.
Positive-semidefinite edge covariances and equal-edge averaging imply $\sigma_{gk}\leq1$; clipping avoids the singular boundary (proof in Appendix~\ref{sec:app-bound}).
Thus, $\ell_0$ and $\ell_1,\ldots,\ell_G$ provide the global and local matching scores passed to MSF, preserving each part's contribution before fusion.

\begin{table*}[t]
\centering
\caption{\textbf{Cross-Modal Classification.} Top-1 accuracy (\%) on 17 settings. The gray row marks Frozen Cosine as the default reference; bold and underline mark the best and second-best pivot-only results. $^*$: class names on the text side.}
\label{tab:classification}
\fontsize{8.0}{9.6}\selectfont
\setlength{\tabcolsep}{2.4pt}
\renewcommand{\arraystretch}{1.08}
\resizebox{\textwidth}{!}{\begin{tabular}{@{}l*{17}{c}@{}}
\toprule
 & \multicolumn{8}{c}{ImageBind~\cite{imagebind}} & \multicolumn{9}{c}{LanguageBind~\cite{languagebind}} \\
\cmidrule(lr){2-9}\cmidrule(lr){10-18}
\shortstack{\vphantom{Ag$^*$}\phantom{M}\\\vphantom{Ag$^*$}Method} & \shortstack{\vphantom{Ag$^*$}BV\\\vphantom{Ag$^*$}{\fontsize{7}{8.4}\selectfont Au--D}} & \shortstack{\vphantom{Ag$^*$}E4D\\\vphantom{Ag$^*$}{\fontsize{7}{8.4}\selectfont Au--I}} & \shortstack{\vphantom{Ag$^*$}UTD\\\vphantom{Ag$^*$}{\fontsize{7}{8.4}\selectfont D--I}} & \shortstack{\vphantom{Ag$^*$}TR\\\vphantom{Ag$^*$}{\fontsize{7}{8.4}\selectfont D--Th}} & \shortstack{\vphantom{Ag$^*$}VGG$^*$\\\vphantom{Ag$^*$}{\fontsize{7}{8.4}\selectfont Tx--Au}} & \shortstack{\vphantom{Ag$^*$}E4D\\\vphantom{Ag$^*$}{\fontsize{7}{8.4}\selectfont Tx--I}} & \shortstack{\vphantom{Ag$^*$}TR$^*$\\\vphantom{Ag$^*$}{\fontsize{7}{8.4}\selectfont Tx--Th}} & \shortstack{\vphantom{Ag$^*$}NYU$^*$\\\vphantom{Ag$^*$}{\fontsize{7}{8.4}\selectfont Tx--D}} & \shortstack{\vphantom{Ag$^*$}BV\\\vphantom{Ag$^*$}{\fontsize{7}{8.4}\selectfont D--Au}} & \shortstack{\vphantom{Ag$^*$}VGG\\\vphantom{Ag$^*$}{\fontsize{7}{8.4}\selectfont Im--Au}} & \shortstack{\vphantom{Ag$^*$}NYU\\\vphantom{Ag$^*$}{\fontsize{7}{8.4}\selectfont Im--D}} & \shortstack{\vphantom{Ag$^*$}TR\\\vphantom{Ag$^*$}{\fontsize{7}{8.4}\selectfont Im--Th}} & \shortstack{\vphantom{Ag$^*$}UCF\\\vphantom{Ag$^*$}{\fontsize{7}{8.4}\selectfont Im--V}} & \shortstack{\vphantom{Ag$^*$}TR\\\vphantom{Ag$^*$}{\fontsize{7}{8.4}\selectfont Th--D}} & \shortstack{\vphantom{Ag$^*$}MSR\\\vphantom{Ag$^*$}{\fontsize{7}{8.4}\selectfont V--Au}} & \shortstack{\vphantom{Ag$^*$}TR\\\vphantom{Ag$^*$}{\fontsize{7}{8.4}\selectfont V--D}} & \shortstack{\vphantom{Ag$^*$}TR\\\vphantom{Ag$^*$}{\fontsize{7}{8.4}\selectfont V--Th}} \\
\midrule
\fullrowband{gray!13}
Frozen Cosine~\cite{languagebind} & 24.7 & 10.4 & 2.8 & 5.8 & 55.9 & 5.8 & 2.0 & 26.6 & 29.9 & 44.8 & 61.0 & 89.9 & 82.8 & 78.1 & 21.9 & 86.0 & 92.3 \\
\midrule
UCL~\cite{ucl} & 18.3 & 14.7 & 3.7 & 6.8 & 51.0 & 10.5 & 2.9 & 27.7 & 34.0 & 39.5 & 56.2 & 70.0 & 64.2 & 71.5 & 20.9 & 71.8 & 73.0 \\
TextME~\cite{textme} & 22.6 & 10.9 & 3.1 & 8.5 & 49.4 & 8.7 & 2.9 & 22.0 & 32.7 & 44.0 & 61.9 & 89.9 & 81.3 & 83.8 & 23.0 & 87.3 & 90.3 \\
DBSN~\cite{dbsn} & 28.8 & 11.6 & 3.3 & 7.7 & \underline{58.6} & 6.5 & 1.9 & 27.5 & 38.4 & 54.1 & 60.6 & 89.9 & 85.3 & 83.5 & 24.4 & 86.3 & 89.2 \\
Ex-MCR~\cite{exmcr} & 39.5 & 10.3 & 6.7 & 21.2 & 54.2 & 10.6 & 12.7 & 32.0 & 43.9 & 60.9 & 64.7 & 91.6 & 86.1 & 77.5 & 33.4 & 82.6 & 91.3 \\
ASIF~\cite{asif} & 67.2 & \underline{24.3} & 11.5 & 44.0 & 47.6 & 23.6 & 31.4 & 25.9 & 61.1 & 57.0 & 69.1 & 92.7 & 88.0 & 91.2 & 38.3 & 92.1 & \underline{92.8} \\
EmergentBridge~\cite{emergentbridge} & 68.3 & 23.7 & 16.9 & 46.3 & 58.5 & \underline{24.5} & 44.1 & 35.8 & 69.0 & \underline{65.9} & 71.8 & 92.3 & 90.7 & \textbf{91.7} & 37.7 & \underline{92.3} & 92.3 \\
Latent Translation~\cite{latenttranslation} & 71.6 & 22.3 & 13.7 & 42.2 & 57.3 & 17.6 & 40.8 & \textbf{41.9} & 68.7 & 63.2 & \underline{72.1} & \underline{93.0} & 89.7 & \underline{91.6} & 33.0 & 91.9 & 92.5 \\
MMBind~\cite{mmbind} & \underline{72.0} & 23.4 & \underline{17.0} & \underline{50.4} & 54.9 & 23.5 & \underline{48.5} & 37.6 & \underline{70.2} & 64.4 & 70.0 & 92.1 & \underline{91.0} & 90.1 & \underline{38.9} & 90.4 & 91.5 \\
\midrule
\fullrowband{resultshade}
\textbf{PRAM (Ours)} & \textbf{74.4} & \textbf{30.6} & \textbf{50.7} & \textbf{51.9} & \textbf{64.7} & \textbf{26.4} & \textbf{65.7} & \underline{40.4} & \textbf{75.8} & \textbf{68.8} & \textbf{72.2} & \textbf{93.4} & \textbf{91.6} & \textbf{91.7} & \textbf{42.6} & \textbf{92.4} & \textbf{92.9} \\
\bottomrule
\end{tabular}}
\end{table*}

\subsection{Multi-Granularity Score Fusion}
\label{sec:msf}
MSF aggregates global and local matching scores and combines complementary scores for classification and retrieval.

\noindent\textbf{Base Scores.}
The correlation score is $s_C=\ell_0+\omega\sum_{g=1}^G\ell_g$, where $\omega=1/G$ when exactly one of $A,P,B$ has multiple local vectors before broadcasting, and $\omega=1$ otherwise.
The two single-vector modalities then enter every local score through broadcasting; averaging counts them once, while $\ell_0$ keeps the separate global relation.
To complement endpoint-whitened correlations, MSF maps features into pivot coordinates using unshrunk moments.
For $X\in\{A,B\}$ with edge $E_X$,
\begin{equation}
\phi_{X,g}(x)=
(\widehat\Sigma_{PP,g}^{(E_X)}+\rho(\widehat\Sigma_{PP,g}^{(E_X)})I)^{-1/2}
\widehat\Sigma_{PX,g}^{(E_X)}\bar x_g.
\label{eq:covmap}
\end{equation}
The mean cosine similarity between $\phi_{A,g}(a)$ and $\phi_{B,g}(b)$ over $g=0,\ldots,G$ defines the covariance score $s_V$, retaining covariance weighting to complement $s_C$.

\noindent\textbf{Auxiliary Scores.}
Global pivot vectors rounded to four decimal places define pivot states, with shared support $\mathcal K_\cap$ of states present on both edges.
For each endpoint, separate shrinkage linear discriminant analysis predictors~\cite{fisher,friedman,ledoitwolf} use the global vector and, when multiple local parts are available, their mean vector to predict pivot states.
Geometric pooling~\cite{heskes} combines voter posteriors into $q_X$, with nonnegative unit-sum weights fitted by two-fold held-out loss~\cite{wolpert}.
For pooled shared-support frequency $\pi_\kappa$, the pivot-association score is
\begin{equation}
s_T(a,b)=\log\sum_{\kappa\in\mathcal K_\cap}
\frac{q_A(\kappa\mid a)q_B(\kappa\mid b)}{\pi_\kappa}.
\label{eq:state}
\end{equation}
For class semantics, repeated class prompts define $A$ states even when pivot samples differ.
The class-semantic route fits a predictor from pivot features to class-prompt states on $\mathcal D_{AP}$, assigns predicted states to samples in $\mathcal D_{PB}$, and fits endpoint posteriors.
Their compatibility, computed in the same form as Eq.~\ref{eq:state}, defines the class-semantic score $s_A$.

\noindent\textbf{Score Fusion.}
MSF uses reference banks $\mathcal B_A,\mathcal B_B$ of up to 2,000 training samples per edge.
For $s\in\{s_C,s_V\}$, symmetric standardization gives
\begin{equation}
\mathcal Z_s(a,b)=\frac12\left[
\frac{s(a,b)-\mu_s^A(a)}{\varsigma_s^A(a)}+
\frac{s(a,b)-\mu_s^B(b)}{\varsigma_s^B(b)}\right].
\label{eq:bank}
\end{equation}
The first mean and standard deviation are computed over $\mathcal B_B$ for fixed $a$; the second over $\mathcal B_A$ for fixed $b$.
For $s_T,s_A$, let $\overline{\mathcal Z}_T,\overline{\mathcal Z}_A$ denote standardization using their overall means and standard deviations on $\mathcal B_A\times\mathcal B_B$.
Let $m$ count distinct pivot states among $n=n_A+n_B$ training samples, with state assignment $\kappa(p)$.
The repeat index $w$ and shared-state coverage $o$ are
\begin{equation}
\begin{aligned}
w&=\max\{0,1-\log m/\log n\},\\
o&=\frac12\sum_{E\in\{AP,PB\}}
\frac{\#\{p\in E:\kappa(p)\in\mathcal K_\cap\}}{n_E}.
\end{aligned}
\label{eq:gates}
\end{equation}
Here, $n_{AP}=n_A$, $n_{PB}=n_B$, and $w=0$ for $n<3$.
For class-prompt states in $\mathcal D_{AP}$, compute $w_A$ by Eq.~\ref{eq:gates}, counting their $m_A$ distinct states among the $n_A$ samples.
Let $\chi_A$ be the two-fold held-out accuracy of predicting these states from pivot features, linearly rescaled and clipped to $[0,1]$ with uniform chance mapped to zero and perfect accuracy to one.
Set $\tau_A=w_A\chi_A$ for class-prompt classification with above-chance pivot accuracy, and $\tau_A=0$ otherwise (details in Appendix~\ref{sec:app-class-semantics}).
The task scores are
\begin{equation}
\begin{aligned}
R(a,b)&=\mathcal Z_{s_C}(a,b)+w\mathcal Z_{s_V}(a,b),\\
S_0(a,b)&=(1-o)R(a,b)+o\overline{\mathcal Z}_T(a,b),\\
S(a,b)&=(1-\tau_A)S_0(a,b)+\tau_A\overline{\mathcal Z}_A(a,b).
\end{aligned}
\label{eq:readouts}
\end{equation}
Zero-weight components are skipped.
All fitted predictors, weights, and reference banks remain fixed at inference; the conditional moments in Eq.~\ref{eq:bank} depend on the current endpoints.
Classification selects the class prompt maximizing $S$ or uses majority voting over the ten labeled samples with the highest $S$ in the other modality, depending on the evaluation protocol.
Retrieval ranks gallery samples by $R$.
The same fitted relations thus support both tasks.

\section{Experiments}
\label{sec:experiments}

\noindent\textbf{Datasets.}
We evaluate 17 classification and 20 retrieval settings across ten datasets: BatVision~\cite{batvision} (BV), UTD-MHAD~\cite{utdmhad} (UTD), TartanRGBT~\cite{anythermal} (TR), VGGSound~\cite{vggsound} (VGG), NYUv2~\cite{nyuv2} (NYU), UCF101~\cite{ucf101} (UCF), MSR-VTT~\cite{msrvtt} (MSR), Ego4D~\cite{ego4d} (E4D), CART~\cite{cart} (CT), and MAVD~\cite{mavd} (MAV).
We further evaluate PRAM in three sensing scenarios on MS2~\cite{ms2}, OBV~\cite{obv}, ViViD++~\cite{vivid}, UTD, and CZU-MHAD~\cite{czumhad} (CZU).
Au, D, Th, Tx, Im, V, and I denote audio, depth, thermal, text, RGB image, video, and inertial measurement unit (IMU) signals, respectively.
In the pivot-based setting, fitting, validation, registration, and score standardization use only modality--pivot training pairs.
NYU Tx--D uses class names for classification and per-image text for retrieval.

\noindent\textbf{Metrics.}
Following~\cite{clip,dinov2}, classification uses class-prompt matching or ten-neighbor label transfer, with supervision from text pivots or labeled galleries.
Following~\cite{imagebind,emergentbridge}, retrieval reports mean bidirectional Recall@1 and Recall@10 (R@1 and R@10), with expected recall for tied scores.
Retrieval settings marked with $^*$ treat same-class samples as matches.
Both directions reuse the same fitted relations, swapping only the query and gallery roles.

\noindent\textbf{Parameters.}
Frozen ImageBind~\cite{imagebind} (IB) and LanguageBind~\cite{languagebind} (LB) provide their official sample-level embeddings as global vectors.
For IB on TR, validation on modality--pivot training pairs selects depth, image, and thermal blocks 6, 12, and 4 (thermal block 6 for unseen trajectories) on a $6\times7$ grid; other IB spatial settings use $7\times7$ grids.
LB uses block-12 features on an $8\times8$ grid.
Temporal settings use 4--8 synchronized segments; single-vector features are broadcast across parts.
The main tables report nine sample-level baselines.
Fitting takes 1--175\,s for the timed configurations on one NVIDIA RTX 4070 SUPER.
All 15 comparators, full metrics, dataset splits, adaptations, and detailed costs are provided in the Appendix.

\subsection{Main Results}
\label{sec:main-results}

\begin{figure}[t]
\centering
\includegraphics[width=0.98\linewidth]{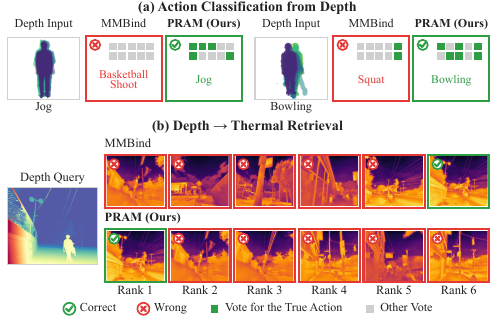}
\vspace{-0.2cm}
\caption{\textbf{Qualitative Classification and Retrieval.} (a) IB UTD action classification from depth; (b) IB TR depth--thermal retrieval. Examples are drawn at random from test cases where PRAM is correct and MMBind is not.}
\label{fig:qualitative}
\end{figure}

\noindent\textbf{Classification.}
Among pivot-only methods in Table~\ref{tab:classification}, PRAM leads 16 of 17 settings, one of them tied, and ranks second in the other.
Its mean accuracy exceeds the best baseline average by 9.7 points on IB and 2.0 on LB.
MMBind~\cite{mmbind} is the strongest IB baseline; PRAM leads it in every IB setting, including UTD D--I (50.7 vs.\ 17.0).
In the two UTD cases in Figure~\ref{fig:qualitative}~(a), 5 and 6 of PRAM's ten label-transfer neighbors vote for jog and bowling, respectively, compared with 0 and 2 for MMBind.
The UTD cases illustrate how local relations support label transfer.

\begin{figure}[t]
\centering
\includegraphics[width=0.98\linewidth]{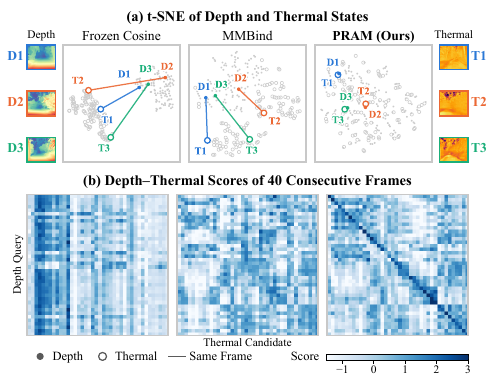}
\vspace{-0.2cm}
\caption{\textbf{Internal States in IB TR Depth--Thermal Retrieval.} (a) PRAM aligns paired states while separating frames; (b) Scores for 40 consecutive frames reveal PRAM's distinct true-pair diagonal compared with Frozen Cosine and MMBind. D$k$, T$k$: depth and thermal samples from frame $k$.}
\label{fig:internal-states}
\end{figure}

\noindent\textbf{Retrieval.}
Among pivot-only methods in Table~\ref{tab:retrieval}, PRAM leads 19 of 20 settings and ranks second in the other.
Its mean R@10 exceeds the best baseline average by 11.3 points on IB and 10.4 on LB.
MMBind~\cite{mmbind} is the strongest IB baseline; PRAM leads it in every IB setting, including TR D--Th (75.1 vs.\ 17.5).
In Figure~\ref{fig:qualitative}~(b), MMBind ranks five thermal frames of similar street scenes above the true match of a TR depth query, whereas PRAM ranks it first.
Figure~\ref{fig:internal-states}~(a) shows PRAM aligning paired states and separating frames, unlike sample-level methods; Figure~\ref{fig:internal-states}~(b) shows a distinct PRAM diagonal: true matches rank highest in 35 of 40 rows versus 6 for MMBind.
On trajectories unseen in fitting, PRAM reaches 62.7 R@10 on IB TR, compared with 10.5 for MMBind.
This helps distinguish similar scenes and generalize to unseen trajectories.

\begin{table*}[t]
\centering
\caption{\textbf{Cross-Modal Retrieval.} Bidirectional R@10 (\%) across 20 settings. The gray row marks Frozen Cosine as the default reference; bold and underline mark the best and second-best pivot-only results. $^*$: class names on the text side.}
\label{tab:retrieval}
\label{tab:retrieval-lb}
\fontsize{8.0}{9.6}\selectfont
\setlength{\tabcolsep}{2.0pt}
\renewcommand{\arraystretch}{1.08}
\resizebox{\textwidth}{!}{\begin{tabular}{@{}l*{20}{c}@{}}
\toprule
 & \multicolumn{10}{c}{ImageBind~\cite{imagebind}} & \multicolumn{10}{c}{LanguageBind~\cite{languagebind}} \\
\cmidrule(lr){2-11}\cmidrule(lr){12-21}
\shortstack{\vphantom{Ag$^*$}\phantom{M}\\\vphantom{Ag$^*$}Method} & \shortstack{\vphantom{Ag$^*$}BV\\\vphantom{Ag$^*$}{\fontsize{7}{8.4}\selectfont Au--D}} & \shortstack{\vphantom{Ag$^*$}E4D\\\vphantom{Ag$^*$}{\fontsize{7}{8.4}\selectfont Au--I}} & \shortstack{\vphantom{Ag$^*$}MAV\\\vphantom{Ag$^*$}{\fontsize{7}{8.4}\selectfont Au--Th}} & \shortstack{\vphantom{Ag$^*$}UTD\\\vphantom{Ag$^*$}{\fontsize{7}{8.4}\selectfont D--I}} & \shortstack{\vphantom{Ag$^*$}TR\\\vphantom{Ag$^*$}{\fontsize{7}{8.4}\selectfont D--Th}} & \shortstack{\vphantom{Ag$^*$}VGG$^*$\\\vphantom{Ag$^*$}{\fontsize{7}{8.4}\selectfont Tx--Au}} & \shortstack{\vphantom{Ag$^*$}E4D\\\vphantom{Ag$^*$}{\fontsize{7}{8.4}\selectfont Tx--I}} & \shortstack{\vphantom{Ag$^*$}TR$^*$\\\vphantom{Ag$^*$}{\fontsize{7}{8.4}\selectfont Tx--Th}} & \shortstack{\vphantom{Ag$^*$}CT\\\vphantom{Ag$^*$}{\fontsize{7}{8.4}\selectfont Th--I}} & \shortstack{\vphantom{Ag$^*$}NYU\\\vphantom{Ag$^*$}{\fontsize{7}{8.4}\selectfont Tx--D}} & \shortstack{\vphantom{Ag$^*$}BV\\\vphantom{Ag$^*$}{\fontsize{7}{8.4}\selectfont D--Au}} & \shortstack{\vphantom{Ag$^*$}VGG\\\vphantom{Ag$^*$}{\fontsize{7}{8.4}\selectfont Im--Au}} & \shortstack{\vphantom{Ag$^*$}NYU\\\vphantom{Ag$^*$}{\fontsize{7}{8.4}\selectfont Im--D}} & \shortstack{\vphantom{Ag$^*$}TR\\\vphantom{Ag$^*$}{\fontsize{7}{8.4}\selectfont Im--Th}} & \shortstack{\vphantom{Ag$^*$}UCF\\\vphantom{Ag$^*$}{\fontsize{7}{8.4}\selectfont Im--V}} & \shortstack{\vphantom{Ag$^*$}MAV\\\vphantom{Ag$^*$}{\fontsize{7}{8.4}\selectfont Th--Au}} & \shortstack{\vphantom{Ag$^*$}TR\\\vphantom{Ag$^*$}{\fontsize{7}{8.4}\selectfont Th--D}} & \shortstack{\vphantom{Ag$^*$}MSR\\\vphantom{Ag$^*$}{\fontsize{7}{8.4}\selectfont V--Au}} & \shortstack{\vphantom{Ag$^*$}TR\\\vphantom{Ag$^*$}{\fontsize{7}{8.4}\selectfont V--D}} & \shortstack{\vphantom{Ag$^*$}TR\\\vphantom{Ag$^*$}{\fontsize{7}{8.4}\selectfont V--Th}} \\
\midrule
\fullrowband{gray!13}
Frozen Cosine~\cite{languagebind} & 2.0 & 1.7 & 1.6 & 3.1 & 0.3 & 76.0 & 1.5 & 7.7 & 4.1 & 14.4 & 2.7 & 43.6 & 50.7 & 36.9 & 96.6 & 2.0 & 25.2 & 7.0 & 25.0 & 37.2 \\
\midrule
UCL~\cite{ucl} & 2.0 & 2.5 & 1.8 & 2.2 & 0.1 & 66.5 & 1.7 & 15.7 & 4.2 & 5.8 & 2.5 & 27.4 & 9.1 & 5.6 & 22.6 & 2.3 & 5.2 & 6.2 & 5.6 & 7.3 \\
DBSN~\cite{dbsn} & 2.3 & 3.1 & 2.1 & 2.4 & 0.4 & 78.5 & 1.5 & 8.7 & 2.0 & \underline{14.8} & 2.5 & \textbf{56.0} & \underline{81.9} & \underline{56.3} & \underline{99.2} & 1.3 & \underline{35.7} & 10.4 & \underline{38.3} & \underline{48.7} \\
TextME~\cite{textme} & 2.5 & 2.0 & 1.2 & 2.3 & 0.3 & 73.2 & 2.0 & 10.4 & 2.4 & 9.9 & 2.0 & 41.3 & 54.7 & 36.1 & 94.5 & 1.7 & 24.9 & 8.4 & 23.7 & 33.4 \\
Ex-MCR~\cite{exmcr} & 4.5 & 2.8 & 3.9 & 2.7 & 2.4 & 75.6 & 2.8 & 40.5 & 4.5 & 12.4 & 3.2 & 53.8 & 14.4 & 15.1 & 57.4 & 3.1 & 10.0 & 15.3 & 13.5 & 13.5 \\
Latent Translation~\cite{latenttranslation} & 7.3 & 5.0 & \underline{4.5} & 3.7 & 8.0 & 78.7 & 3.5 & 70.4 & 5.9 & 13.5 & 4.0 & 49.5 & 14.1 & 22.0 & 74.3 & 3.6 & 16.7 & 15.0 & 14.8 & 20.1 \\
MMBind~\cite{mmbind} & 8.0 & 5.4 & 4.1 & \underline{5.9} & \underline{17.5} & 77.5 & 4.5 & \underline{74.2} & 4.1 & 14.4 & 5.4 & 47.7 & 12.9 & 21.4 & 59.1 & \underline{5.2} & 16.1 & 17.0 & 17.0 & 18.8 \\
ASIF~\cite{asif} & 8.2 & \underline{5.7} & 4.0 & 3.8 & 9.6 & 71.1 & \underline{4.7} & 64.9 & \underline{6.5} & 10.7 & 4.7 & 48.1 & 12.1 & 20.9 & 38.3 & 3.9 & 13.8 & 13.4 & 18.0 & 19.2 \\
EmergentBridge~\cite{emergentbridge} & \underline{8.3} & \underline{5.7} & 2.8 & 4.7 & 11.4 & \underline{78.8} & 4.1 & 72.1 & \underline{6.5} & 14.4 & \underline{5.7} & 50.5 & 20.2 & 27.9 & 77.9 & 3.0 & 18.3 & \underline{20.2} & 17.3 & 15.1 \\
\midrule
\fullrowband{resultshade}
\textbf{PRAM (Ours)} & \textbf{11.1} & \textbf{10.9} & \textbf{5.1} & \textbf{22.7} & \textbf{75.1} & \textbf{82.1} & \textbf{6.9} & \textbf{76.3} & \textbf{19.8} & \textbf{18.7} & \textbf{6.3} & \underline{53.9} & \textbf{98.5} & \textbf{76.1} & \textbf{99.6} & \textbf{5.3} & \textbf{54.6} & \textbf{26.4} & \textbf{51.8} & \textbf{61.5} \\
\bottomrule
\end{tabular}}
\end{table*}

\subsection{Ablation Study}
\label{sec:ablation}

\noindent\textbf{Representation.}
In Table~\ref{tab:ablation}, Global-only PRAM reduces mean R@10 from 32.9 to 22.4 on IB and from 53.4 to 30.1 on LB.
Removing the separate global component lowers classification by 1.1 points on IB and 5.2 on LB, and LB retrieval by 2.4.
Local vectors improve instance discrimination; the global component adds complementary evidence.

\noindent\textbf{Covariance Regularization.}
With 2\% of the pairs, removing cross-part shrinkage ($\alpha=0$ in Eq.~\ref{eq:shrink}) lowers R@10 from 59.0 to 21.7 on IB TR, from 46.4 to 25.0 on its unseen trajectories, and from 94.0 to 41.6 on LB TR Th--D with an RGB pivot.
These means come from the two subsets of Sec.~\ref{sec:mechanisms}; all ten random subsets shared by both variants show each drop (Table~\ref{tab:app-budget-ten}).
MCR's cross-part shrinkage matters with scarce modality--pivot pairs.

\noindent\textbf{Relation Scoring.}
Partwise ASIF and Partwise MMBind use the same local vectors and average 41.4 and 43.2 versus 54.3 for PRAM; Table~\ref{tab:app-local-methods} compares four more methods on these vectors.
Removing the quadratic term that penalizes individual coordinates in Eq.~\ref{eq:llr} lowers R@10 from 32.9 to 30.5 on IB and from 53.4 to 47.1 on LB.
The same vectors yield stronger results with PRAM, and its quadratic term further improves retrieval.

\noindent\textbf{Score Fusion.}
Removing the covariance score or standardization lowers LB R@10 from 53.4 to 45.6 or 46.2, respectively; top-1 changes by at most 1.2 points on either backbone.
Removing pivot associations lowers LB top-1 from 80.2 to 74.6, and removing class semantics lowers IB top-1 from 50.6 to 46.7, with retrieval unchanged.
The covariance score and standardization strengthen instance matching for retrieval, whereas pivot associations and class semantics mainly support category prediction.

\begin{table}[t]
\centering
\caption{\textbf{Ablations.} Mean retrieval R@10 and classification accuracy (\%) across settings; Avg.\ averages the four columns.}
\label{tab:ablation}
\fontsize{8.0}{9.6}\selectfont
\setlength{\tabcolsep}{1.5pt}
\renewcommand{\arraystretch}{1.08}
\begin{tabularx}{\linewidth}{@{}>{\raggedright\arraybackslash}p{.36\linewidth}*{5}{Y}@{}}
\toprule
 & \multicolumn{2}{c}{ImageBind~\cite{imagebind}} & \multicolumn{2}{c}{\makebox[0pt]{LanguageBind~\cite{languagebind}}} & \\
\cmidrule(lr){2-3}\cmidrule(lr){4-5}
Variant & R@10 & Top-1 & R@10 & Top-1 & Avg. \\
\midrule
Frozen Cosine & 11.2 & 16.7 & 32.7 & 65.2 & 31.5 \\
Global-only PRAM & 22.4 & 43.1 & 30.1 & 79.5 & 43.8 \\
Partwise ASIF & 26.5 & 35.6 & 31.3 & 72.1 & 41.4 \\
Partwise MMBind & 28.0 & 44.8 & 27.5 & 72.7 & 43.2 \\
\fullrowband{resultshade}\textbf{PRAM (Ours)} & 32.9 & 50.6 & 53.4 & 80.2 & \textbf{54.3} \\
\midrule
w/o global component & 32.8 & 49.5 & 51.0 & 75.0 & 52.1 \\
w/o cross-part shrinkage & 32.2 & 50.6 & 53.4 & 80.2 & 54.1 \\
w/o quadratic term & 30.5 & 48.6 & 47.1 & 80.3 & 51.6 \\
w/o covariance score & 32.9 & 50.6 & 45.6 & 80.2 & 52.3 \\
w/o standardization & 32.6 & 49.4 & 46.2 & 80.1 & 52.1 \\
w/o pivot associations & 32.9 & 50.6 & 53.4 & 74.6 & 52.9 \\
w/o class semantics & 32.9 & 46.7 & 53.4 & 80.2 & 53.3 \\
\bottomrule
\end{tabularx}
\end{table}

\begin{table*}[t]
\centering
\caption{\textbf{Mechanism Analysis.} R@10 (\%) with frozen features; $\Delta$: change in points. (a) Parts on IB TR; $\Delta$ against one part; GB: peak GPU memory; the best baseline reaches 17.5. (b) Merged features versus the best of 4, 12, and 42 parts on TR, with 12 parts for IB and 4 for LB, both under a total cap of 256 directions. (c) Full pivots versus pivot parts shuffled within each sample, and state labels.}
\label{tab:mechanisms}
\fontsize{6.8}{8.4}\selectfont
\setlength{\tabcolsep}{1.6pt}
\renewcommand{\arraystretch}{1.15}
\begin{tabularx}{\textwidth}{@{}r*{4}{Y}@{\hspace{7pt}}l*{3}{Y}@{\hspace{7pt}}l*{4}{Y}@{}}
\toprule
\multicolumn{5}{c}{(a) Spatial Granularity} & \multicolumn{4}{c}{(b) Equal Direction Budget} & \multicolumn{5}{c}{(c) Pivot Structure} \\
\cmidrule(r){1-5}\cmidrule(lr){6-9}\cmidrule(l){10-14}
Parts & PRAM & $\Delta$ & R@1 & GB & Setting & Merged & Partwise & $\Delta$ & Setting & Full & Shuffled & $\Delta$ & State \\
\midrule
1 & 21.7 & -- & 3.9 & 3.57 & IB D--Th & 22.9 & 62.7 & \textcolor{cvprblue}{+39.8} & IB TR D--Th & 75.1 & 23.5 & \textcolor{deltaneg}{\textminus 51.6} & 12.9 \\
4 & 49.6 & \textcolor{cvprblue}{+27.9} & 15.5 & 3.58 & IB Im--Th & 30.5 & 72.2 & \textcolor{cvprblue}{+41.7} & IB UTD D--I & 22.7 & 13.6 & \textcolor{deltaneg}{\textminus 9.1} & 27.6 \\
16 & 68.3 & \textcolor{cvprblue}{+46.6} & 35.1 & 3.59 & IB Im--D & 36.0 & 81.6 & \textcolor{cvprblue}{+45.6} & LB TR Th--D & 98.5 & 69.0 & \textcolor{deltaneg}{\textminus 29.5} & 53.2 \\
\cellcolor{resultshade}42 & \cellcolor{resultshade}\textbf{75.1} & \cellcolor{resultshade}\textcolor{cvprblue}{+53.4} & \cellcolor{resultshade}46.8 & \cellcolor{resultshade}3.62 & LB V--Th & 75.2 & 84.6 & \textcolor{cvprblue}{+9.4} & LB TR V--D & 89.3 & 60.2 & \textcolor{deltaneg}{\textminus 29.1} & 50.8 \\
168 & 74.1 & \textcolor{cvprblue}{+52.4} & 48.4 & 8.21 & LB V--D & 75.9 & 86.4 & \textcolor{cvprblue}{+10.5} & LB TR V--Th & 89.9 & 64.4 & \textcolor{deltaneg}{\textminus 25.5} & 59.4 \\
\bottomrule
\end{tabularx}
\end{table*}

\subsection{Further Analysis}
\label{sec:mechanisms}

\begin{figure}[t]
\centering
\includegraphics[width=0.98\linewidth]{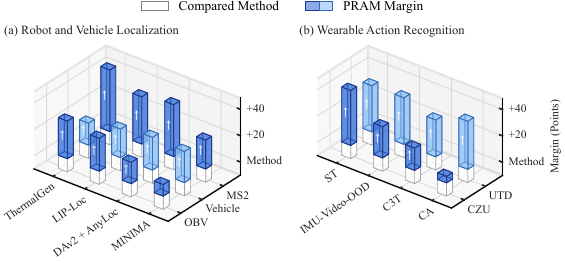}
\vspace{-0.2cm}
\caption{\textbf{Knowledge Activation in Sensing Scenarios.} Blue boxes show percentage-point gains over white bases (compared methods): (a) Localization success; (b) Top-1 accuracy.}
\label{fig:applications}
\end{figure}

\noindent\textbf{Sensing Scenarios.}
\label{sec:applications}
Figure~\ref{fig:applications} uses task-specific frozen encoders and protocols beyond main-table settings. Robot localization uses AnyThermal~\cite{anythermal} and DINOv2-B~\cite{dinov2} features for thermal-to-depth retrieval via RGB pivots, with pose-distance success thresholds.
With MINIMA~\cite{minima}, Depth Anything V2 (DAv2)~\cite{depthanythingv2} with AnyLoc~\cite{anyloc}, LIP-Loc~\cite{liploc}, and ThermalGen~\cite{thermalgen} as baselines, PRAM reaches 55.3\% within 5\,m on MS2 and 31.9\% within 0.25\,m on OBV, versus MINIMA's 33.4\% and 23.1\%, respectively.
For cross-device vehicle localization, PRAM fits relations on daytime ViViD++ RGB--thermal pairs and MS2 depth--RGB pairs across vehicles and sensors, localizing nighttime ViViD++ thermal queries against MS2 depth with 26.2\% success within 5\,m versus ThermalGen's 9.6\%.
Following C3T~\cite{c3t}, wearable action labels link to IMU via labeled videos (pivot) and unlabeled video--IMU pairs; testing uses IMU alone.
For wearables, we compare to IMU-Video-OOD-HAR (IMU-Video-OOD)~\cite{imu_video_ood} and reimplement C3T and its contrastive-alignment (CA) and student--teacher (ST) baselines on PRAM's splits.
PRAM's IMU-only top-1, averaged over seeds 1--3, reaches 80.6\% on UTD versus C3T's 52.7\%, and 90.8\% on CZU versus CA's 85.9\%.
Activated relations support cross-sensor localization and IMU-only recognition.

\begin{figure*}[t]
\centering
\begin{minipage}[t]{0.485\textwidth}
\centering
\includegraphics[height=0.525\dimexpr0.98\linewidth\relax]{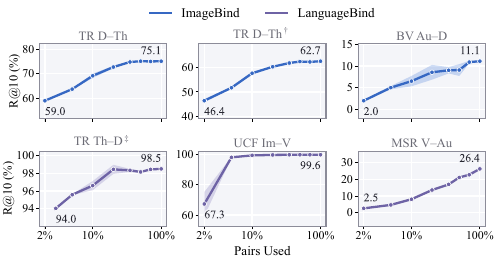}
\vspace{-0.2cm}
\caption{\textbf{Data Efficiency of Knowledge Activation.} Pair budgets vary at fixed features and scores. Points and bands show subset means and ranges. $^\dagger$: unseen trajectories; $^\ddagger$: RGB pivot.}
\label{fig:pivot-budget}
\end{minipage}\hfill
\begin{minipage}[t]{0.485\textwidth}
\centering
\includegraphics[width=0.98\linewidth]{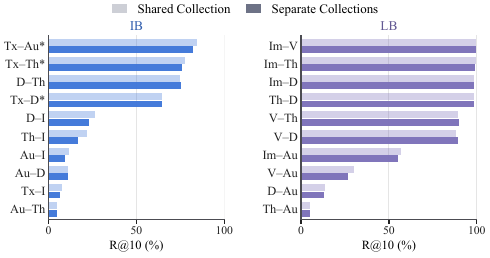}
\vspace{-0.2cm}
\caption{\textbf{Learning Relations across Fitting Collections.} One shared versus two separate modality--pivot collections. Bars show bidirectional R@10; $^*$: class names on the text side.}
\label{fig:unseen-observations}
\end{minipage}
\end{figure*}

\noindent\textbf{Spatial Granularity.}
\label{sec:granularity-study}
Table~\ref{tab:mechanisms}~(a) divides fixed IB features on TR into 1--168 parts, retaining the separate global component.
With 42 parts, R@10 rises from 21.7 to 75.1, and R@1 from 3.9 to 46.8.
With 168 parts, R@10 is 74.1, but peak memory exceeds twice that at 42 parts (8.21 vs.\ 3.62\,GB); the main tables balance cost and accuracy with a $6\times7$ grid for IB TR.
Spatial structure strengthens instance matching with the same frozen layers, and the gain saturates around 42 parts on IB TR.

\noindent\textbf{Direction Budget.}
\label{sec:direction-study}
Table~\ref{tab:mechanisms}~(b) compares merged and partwise features in five TR settings by local correlation alone, with a 256-direction total cap, averaging adjacent cells grouped alike at the endpoints and pivot.
Every setting has a partition that outperforms merging; twelve parts reach 62.7 versus 22.9 on IB depth--thermal retrieval.
Across the five settings, the best partition exceeds merging by 9.4--45.6 points and by 9.3--45.2 points without the quadratic term of Eq.~\ref{eq:llr}.
At 42 parts, the total cap leaves about six directions per part; with 256 per part, all five settings gain a further 14.9--18.1 points (Table~\ref{tab:app-direction}).
Partwise gains persist under a common cap, and finer partitions require enough capacity for each part.

\noindent\textbf{Pivot Structure.}
\label{sec:pivot-content-study}
Table~\ref{tab:mechanisms}~(c) tests within-sample pivot-part shuffling and state-label pivots in five settings with endpoint features fixed, both endpoints partitioned, and $\alpha<1$ (Eq.~\ref{eq:shrink}); LB uses RGB pivots. Table~\ref{tab:supp-pivot} adds part means and all thirteen settings.
Shuffling keeps the pivot vectors but breaks part correspondence, lowering R@10 on IB TR from 75.1 to 23.5 and on the three LB TR settings by 25.5--29.5 points.
State labels alone give 12.9 on IB TR but 27.6 on UTD, where action categories carry the relation.
Local correspondence aids spatial instance matching; state semantics support coarse action relations on UTD.

\noindent\textbf{Data Efficiency.}
\label{sec:pivot-budget-study}
Figure~\ref{fig:pivot-budget} fixes features, scoring rules and test samples at eight nominal modality--pivot budgets from 2\% to 100\%; lower budgets average two random subsets of training pairs, drawn by frame for TartanRGBT.
The 50-pair floor raises the nominal 2\% budget of LB TR Th--D to 2.9\% of pairs.
With 2\% of pairs, IB TR scores 59.0 R@10 on 8,601 random-split test samples, above the best full-data sample-level baseline's 17.5, and 46.4 on unseen trajectories.
LB TR Th--D with an RGB pivot reaches 94.0 using 50 of 1,733 training pairs and 95.6 with 5\%.
IB TR reaches 72.8 R@10 with 20\% of pairs and 75.1 with 50\%.
IB TR on unseen trajectories reaches 60.4 and 62.5 at these budgets.
LB TR Th--D reaches 98.5 with 20\% of pairs, matching its full-budget result.
Small pair collections thus suffice for strong spatial correspondence, also on unseen trajectories.

\noindent\textbf{Separate Collections.}
Figure~\ref{fig:unseen-observations} compares fits on one shared or two separate modality--pivot collections, matching per-edge counts, reference banks and test samples; the main tables always fit both edges on disjoint samples.
Six LB settings use frame-level RGB or depth image pivots because their main-table text prompts are shared by each recording's frames, so shared collections would link $A$ and $B$ via the same frames, bypassing the pivot.
IB NYU Tx--D uses class names.
In 20 settings over ten random splits with both assignments, separate collections lower mean R@10 by 1.0 point and retain 96.6\% of the mean gain over Frozen Cosine.
Separate collections keep at least 95\% of the gain in 11 of the 20 settings and raise R@10 in three TR settings.
Relations are thus recovered across distinct sets of fitting samples, as the pivot-based setting requires.

\section{Conclusion}
\label{sec:conclusion}
We present PRAM for emergent cross-modal classification and retrieval without direct paired supervision between target modalities.
Using only modality--pivot pairs, PRAM activates latent multimodal knowledge in frozen encoders by preserving local discriminative information and composing partwise statistical relations through a shared pivot modality.
MCR stabilizes covariance estimates, PSRR recovers global and partwise statistical relations, and MSF combines standardized matching scores with task-specific pivot associations and class semantics for classification and candidate ranking.
Experiments on ImageBind and LanguageBind across ten datasets show that PRAM achieves the best results in 16 of 17 classification and 19 of 20 retrieval settings, with mean bidirectional R@10 reaching 32.9\% and 53.4\%, respectively.
Evaluations in three sensing scenarios demonstrate its effectiveness across visual and inertial encoders.

\FloatBarrier
\bibliographystyle{plainnat}
\bibliography{main}

\clearpage
\beginappendix
\suppressfloats[t]
\numberwithin{equation}{section}
\setcounter{table}{0}
\setcounter{figure}{0}
\setcounter{algorithm}{0}
\renewcommand{\thetable}{S\arabic{table}}
\renewcommand{\thefigure}{S\arabic{figure}}
\renewcommand{\thealgorithm}{S\arabic{algorithm}}

\section{Notation}
\label{sec:app-setting}
\label{sec:app-notation}
Table~\ref{tab:app-notation} collects the main symbols used in the derivations, implementation details, and full result tables.

\begin{table}[t]
\centering
\caption{\textbf{Notation.} The index $g=0$ is reserved for the global component; $g=1,\ldots,G$ indexes local parts.}
\label{tab:app-notation}
\fontsize{7.2}{8.6}\selectfont
\renewcommand{\arraystretch}{0.94}
\setlength{\tabcolsep}{3pt}
\begin{tabularx}{\linewidth}{@{}p{.36\linewidth}X@{}}
\toprule
Symbol & Meaning\\
\midrule
$A,B,P,X,Y$ & Target modalities $A,B$ and shared pivot modality $P$; $X,Y$ denote modalities on the same edge, either identical (within-modality block) or distinct\\
$\mathcal D_{AP},\mathcal D_{PB}$ & Separate modality--pivot fitting collections, with $n_A,n_B$ samples\\
$x_0,x_g$ & Official global vector and frozen local vector of sample $x$\\
$d,r,G$ & Vector width, shared factors per part, and number of local parts\\
$\eta$ & Retained fraction of content and composed-relation energy\\
$\bar x_g$ & Part-centered feature divided by its modality's training scalar scale\\
$\widehat\Sigma_{XY,g}^{(E)}$ & Empirical covariance block on edge $E$\\
$E,n_E$ & Edge ($AP$ or $PB$) and its number of rows\\
$\widetilde\Sigma_{XY,g},\alpha$ & Cross-part-shrunk covariance block and shared shrinkage intensity\\
$H_g,K_g$ & Equal-edge pivot covariance and composed endpoint cross-covariance\\
$W_{X,g},\rho_{X,g},\lambda$ & Endpoint inverse square root, absolute ridge, and relative ridge coefficient\\
$U_g,V_g,\sigma_g,k_g$ & Singular vectors, singular values, and retained direction count\\
$u_g,v_g,\ell_g$ & Endpoint singular coordinates and component correlation score (Eq.~4 of the main text)\\
$c_{gk}$ & Clipped singular value in $\ell_g$; $k$ indexes retained directions\\
$\omega$ & Structural coefficient of the local correlation sum\\
$\phi_{X,g},s_C,s_V,s_T$ & Covariance map; correlation, covariance, and pivot-association scores\\
$\kappa,\mathcal K_\cap,\pi_\kappa$ & Pivot state, shared state support, and shared-support frequency\\
$q_X,\gamma_{Xe}$ & Pooled endpoint state posterior and voter weight\\
$\mathcal B_A,\mathcal B_B$ & Training reference banks for score standardization\\
$\mathcal Z_s,\mu_s,\varsigma_s$ & Bank-standardized score $s$, bank mean, and bank standard deviation\\
$m,n$ & Distinct pivot states and training samples, $n=n_A+n_B$\\
$\nu,\mathcal A_\cap,\pi_\nu^A$ & Class-prompt state, shared class-prompt support, and pooled frequency\\
$s_A,\tau_A$ & Class-semantic score and its weight $\tau_A=w_A\chi_A$\\
$w_A,\chi_A$ & Class-prompt repeat index and chance-corrected held-out pivot accuracy\\
$w,o,R,S_0,S$ & Repeat index, coverage, retrieval score, and intermediate/final classification scores (Eq.~\ref{eq:app-readouts})\\
\bottomrule
\end{tabularx}
\end{table}

\Needspace*{3\baselineskip}

\section{Population Relations and Capacity}
\label{sec:app-capacity}
\label{sec:population-details}
We first specify the relation determined by two compatible observed edges.
The capacity calculation then isolates how pivot dimension and representation granularity constrain that relation under the stated block model.

\subsection{Compatible Edge Populations}
Let $(A,P,B)$ have a joint distribution with finite second moments.
The two fitting collections independently sample its $AP$ and $PB$ marginals, using the same pivot representation and part coordinates.
They share population pivot mean $\mu_P$ and covariance $\Sigma_{PP}$.
Assume $\Sigma_{PP}\succ0$ for the identities below.
Sample separation follows the data protocol; the assumption concerns compatible edge distributions.

For centered features, the linear predictions from the pivot are
\begin{equation}
\widehat A(P)=\Sigma_{AP}\Sigma_{PP}^{-1}P,\qquad
\widehat B(P)=\Sigma_{BP}\Sigma_{PP}^{-1}P.
\end{equation}
Both residuals $\varepsilon_A=A-\widehat A(P)$ and $\varepsilon_B=B-\widehat B(P)$ are uncorrelated with $P$.
Expanding the endpoint covariance $\operatorname{Cov}(A,B)$ therefore gives
\Needspace{7\baselineskip}
\begin{equation}
\begin{aligned}
\Sigma_{AB}
&=\underbrace{\Sigma_{AP}\Sigma_{PP}^{-1}\Sigma_{PB}}_{\Sigma_{AB}^{(P)}}
+\Sigma_{AB\cdot P},\\
\Sigma_{AB\cdot P}
&=\operatorname{Cov}(\varepsilon_A,\varepsilon_B).
\end{aligned}
\label{eq:supp-residual}
\end{equation}
Finite second moments suffice for this decomposition.
Under joint Gaussianity, the residual is also the conditional cross-covariance~\cite{bishop}, and it vanishes exactly when $A\perp B\mid P$.
The composed term is thus the cross-covariance between the two linear predictions from $P$.
Partwise reasoning across modalities (PRAM) estimates a regularized version of this term for each corresponding part.

\subsection{Identifiability and Correspondence}
The observed marginals determine $\Sigma_{AB}^{(P)}$.
They allow multiple residual endpoint relations.
For example, let $P,\xi,\zeta$ be independent standard Gaussian scalars, and set
\begin{equation}
A=\xi,\qquad
B=t\xi+\sqrt{1-t^2}\,\zeta,\qquad -1<t<1.
\end{equation}
Every $t$ gives the same $AP$ and $PB$ marginals, whereas $\operatorname{Cov}(A,B)=t$ varies.
This example fixes the estimand supported by pivot-only samples.

At part $g$, the same decomposition applies to $(A_g,P_g,B_g)$.
The local route uses $A_g\leftrightarrow P_g\leftrightarrow B_g$.
Cross-part endpoint dependencies and routes through other pivot parts lie outside this component.
The global component supplies an additional sample-level route with its own moments.
Recovering knowledge through these routes requires corresponding coordinates and compatible pivot populations on the two observed edges.

\subsection{A Global Random-Subspace Pivot}
\label{sec:capacity-proof}
In this model, with $L=Gr$ latent factors $z\sim\mathcal N(0,I_L)$, a width-$d$ pivot carries rank $\delta_P=\min(d,L)$; when it is written as $p=Q_Pz$ with $Q_P\in\mathbb R^{\delta_P\times L}$ having orthonormal rows, it retains the fraction $\mathbb E\|p\|^2/\mathbb E\|z\|^2=\delta_P/L=\min\{1,d/(Gr)\}$ of content energy; the following proposition shows that relations composed from it retain the same fraction.
The idealized model concerns a ratio of expected covariance energies:
\Needspace{5\baselineskip}
\begin{equation}
\eta=1-\frac{\mathbb E\|\Sigma_{AB}-\Sigma_{AB}^{(P)}\|_F^2}{\mathbb E\|\Sigma_{AB}\|_F^2},
\label{eq:supp-eta-definition}
\end{equation}
where $\Sigma_{AB}^{(P)}$ is the pivot-composed term of Eq.~\ref{eq:supp-residual}.
Positive endpoint ranks ensure a nonzero denominator.
The expectation is over random projection spaces.

\begin{proposition}[Global pivot capacity]
\label{prop:app-global}
Let $z\sim\mathcal N(0,I_L)$, where $L=Gr$.
Let $a=Q_Az$, $b=Q_Bz$, and $p=Q_Pz$.
Each $Q_X$ has orthonormal rows, with mutually independent, uniformly oriented row spaces in $\mathbb R^L$.
For endpoint ranks $\delta_A,\delta_B>0$ and pivot rank $\delta_P=\min(d,L)$,
\begin{equation}
\eta_{\rm global}=\frac{\delta_P}{L}
=\min\{1,d/(Gr)\}.
\end{equation}
\end{proposition}
\begin{proof}
Define $\Pi_X=Q_X^\top Q_X$.
Then $\Sigma_{AB}=Q_AQ_B^\top$, $\Sigma_{AB}^{(P)}=Q_A\Pi_PQ_B^\top$, and
$\mathbb E\Pi_X=(\delta_X/L)I_L$ by rotational symmetry.
The expected endpoint energy is
\begin{equation}
\mathbb E\|\Sigma_{AB}\|_F^2
=\mathbb E\operatorname{tr}(\Pi_A\Pi_B)
=\delta_A\delta_B/L.
\end{equation}
Writing $T=I_L-\Pi_P$, the residual energy is
\begin{equation}
\begin{aligned}
\mathbb E\|\Sigma_{AB}-\Sigma_{AB}^{(P)}\|_F^2
&=\mathbb E\operatorname{tr}(\Pi_A T\Pi_B T)\\
&=\frac{\delta_A\delta_B(L-\delta_P)}{L^2}.
\end{aligned}
\end{equation}
Here independence permits averaging the endpoint projectors separately, and $T$ is an orthogonal projector of rank $L-\delta_P$.
Substitution into Eq.~\ref{eq:supp-eta-definition} proves the claim.
\end{proof}

\subsection{Corresponding Independent Parts}
\begin{corollary}[Partwise pivot capacity]
\label{cor:app-parts}
Let $z=(z_1,\ldots,z_G)$ have independent $z_g\sim\mathcal N(0,I_r)$.
Within each block, set $a_g=Q_{A,g}z_g$, $b_g=Q_{B,g}z_g$, and $p_g=Q_{P,g}z_g$, each $Q_{X,g}$ with orthonormal rows.
All row spaces are mutually independent across modalities and parts and uniformly oriented within their corresponding blocks.
For endpoint ranks $\delta_A,\delta_B>0$ and local-pivot rank $\delta=\min(d,r)$,
\begin{equation}
\eta_{\rm part}=\delta/r=\min\{1,d/r\}.
\end{equation}
For these same stacked endpoints and an independent global rank-$\min(d,Gr)$ pivot, the global fraction remains $\min\{1,d/(Gr)\}$ under random subspace orientations.
\end{corollary}
\begin{proof}
Independence makes the off-part endpoint cross-covariances zero.
The trace calculation in Proposition~\ref{prop:app-global}, now in dimension $r$, gives $\delta_A\delta_B/r$ expected covariance energy and $\delta_A\delta_B(r-\delta)/r^2$ residual energy for each block.
Summing over the $G$ diagonal blocks gives the stated ratio.
For the same stacked endpoints, the endpoint projectors are mutually independent and independent of the global pivot projector, with expectations $(\delta_A/r)I_{Gr}$ and $(\delta_B/r)I_{Gr}$; the preceding trace argument thus gives residual fraction $(Gr-\delta_P)/(Gr)$.
\end{proof}

\subsection{Total Rank and the Scope of the Result}
Equal width $d$ per vector permits $G\min(d,r)$ total local-pivot rank, compared with $\min(d,Gr)$ global rank.
At a fixed total local rank budget $D_{\rm tot}$, block ranks $\delta_g\leq r$ instead give
\begin{equation}
\eta_{\rm part}=\frac{\sum_{g=1}^G\delta_g}{Gr}.
\end{equation}
If $\sum_{g=1}^G\delta_g=\min(D_{\rm tot},Gr)$, this equals the global expression at rank budget $D_{\rm tot}$.
The equal-width result therefore describes preservation of multiple informative subspaces under the stated block model.
The matched-total-direction experiment tests the framework under a separate rank control using only local correlations.

The proposition and its corollary concern noiseless projections, independent block factors, random relative orientation, and expected cross-covariance energy.

\Needspace*{3\baselineskip}

\section{Covariance Estimation and Correlation Scoring}
\label{sec:app-covariance}

\label{sec:estimator-details}
The first three subsections detail multi-granularity covariance regularization (MCR); the last three detail pivot-based spectral relational reasoning (PSRR).

\subsection{Centering and Covariance Estimation}
Each part is centered by its edge-specific training mean and divided by its modality's scalar scale.
For edge $E$ with $n_E$ rows ($n_{AP}=n_A$, $n_{PB}=n_B$), the sample covariance is
\begin{equation}
\widehat\Sigma_{XY,g}^{(E)}
=\frac{1}{n_E}\sum_{i\in E}\bar x_{ig}\bar y_{ig}^{\top}.
\end{equation}
The local scale of modality $X$, with local-vector width $d_X$, is
\begin{equation}
 \upsilon_X=\left[\max\left\{\frac{1}{Gd_X}\sum_{g=1}^G
 \operatorname{tr}(\widehat\Sigma^{\rm raw}_{XX,g}),10^{-12}\right\}\right]^{1/2}.
\end{equation}
The raw moments precede scalar scaling.
Endpoint scales use their respective edges; the pivot scale uses the union of both fitting collections.
Thus $\bar x_{ig}=(x_{ig}-\widehat\mu_{X,g})/\upsilon_X$ shares one scale across local parts while retaining their separate means.
Global vectors have their own scales and means.
A broadcast endpoint uses the same vector at every local part.
All reported fitting collections and banks contain at least two samples for covariance and score moments.

\subsection{Positive-Semidefinite Joint Shrinkage}
For one edge, let $\widehat J_g$ be its joint endpoint--pivot covariance and
$\overline{\widehat J}=G^{-1}\sum_{g=1}^G\widehat J_g$.
The convex combination
$\widetilde J_g=(1-\alpha)\widehat J_g+\alpha\overline{\widehat J}$
preserves positive semidefiniteness for $0\leq\alpha\leq1$.
Its blocks give Eq.~2 of the main text.
The same $\alpha$ is applied to both edges; the global component uses $\alpha=0$.

For centered, scaled vectors on an edge, estimate block variances by
\begin{equation}
\begin{aligned}
\psi_{XX,g}&=\frac{n_E^{-1}\sum_i\|\bar x_{ig}\|^4
-\|\widehat\Sigma_{XX,g}\|_F^2}{n_E},\\
\psi_{XP,g}&=\frac{n_E^{-1}\sum_i\|\bar x_{ig}\|^2\|\bar p_{ig}\|^2
-\|\widehat\Sigma_{XP,g}\|_F^2}{n_E}.
\end{aligned}
\label{eq:supp-blockvar}
\end{equation}
The pivot self-block has the analogous expression.
Negative estimates are clipped to zero.
Let $\mathcal J$ contain the blocks $(E,XY)$ varying across parts on the edges $E\in\{AP,PB\}$, and let
$t_{XY}=2$ for cross-blocks and $1$ for self-blocks.
Then
{\small
\begin{equation}
\alpha=\operatorname{clip}_{[0,1]}
\frac{\sum_{(E,XY)\in\mathcal J}t_{XY}\sum_{g=1}^G \psi^{(E)}_{XY,g}}
{\sum_{(E,XY)\in\mathcal J}t_{XY}\sum_{g=1}^G
\|\widehat\Sigma^{(E)}_{XY,g}-\overline{\widehat\Sigma}{}^{(E)}_{XY}\|_F^2}.
\label{eq:supp-alpha}
\end{equation}}
A zero denominator gives $\alpha=1$.
The weight two counts a cross-block and its transpose in the joint Frobenius norm.
A broadcast endpoint's self-block is constant across parts and is omitted from the shrinkage sums.

\subsection{Shrinkage Risk and the Estimated Target}
With finite fourth moments, covariance shrinkage~\cite{ledoitwolf} motivates controlling Frobenius risk.
For one edge, write $\Delta_g=\widehat J_g-J_g$ and
$D_g=\widehat J_g-\overline{\widehat J}$, where $J_g$ is the population joint covariance of part $g$.
The total Frobenius risk $\sum_{g=1}^G\mathbb E\|\widehat J_g-\alpha D_g-J_g\|_F^2$ is quadratic in $\alpha$.
Its constrained oracle minimizer is
\begin{equation}
\alpha_* =\operatorname{clip}_{[0,1]}
\frac{\sum_{g=1}^G\mathbb E\langle \Delta_g,D_g\rangle_F}
{\sum_{g=1}^G\mathbb E\|D_g\|_F^2}.
\label{eq:supp-alpha-oracle}
\end{equation}
A zero denominator makes every $\alpha\in[0,1]$ a minimizer.
For an unbiased covariance-estimation benchmark, let
$\overline\Delta=G^{-1}\sum_{g=1}^G \Delta_g$.
Then the numerator equals
\begin{equation}
\sum_{g=1}^G\mathbb E\|\Delta_g\|_F^2
-G\mathbb E\|\overline\Delta\|_F^2.
\label{eq:supp-target-correction}
\end{equation}
This follows by expanding $D_g$ and summing
$\langle \Delta_g,\overline\Delta\rangle_F$.
Equation~\ref{eq:supp-alpha} defines the variance-ratio estimator by pooling the leading block-variance terms from Eq.~\ref{eq:supp-blockvar} across both edges.
Both edges contribute their block variances and cross-part dispersion to the shared intensity $\alpha\in[0,1]$.

\subsection{Regularization and Retained Directions}
For block $M$, define the ridge~\cite{hoerl}
$\rho(M)=\lambda\max\{\operatorname{tr}(M)/\dim(M),10^{-12}\}$.
Endpoint inverse square roots clamp negative numerical eigenvalues to zero before adding the ridge.
The pivot inverse uses the equal-edge mean of shrunk pivot covariances.
Equal-edge averaging is consistent for the common pivot covariance when both edge sample sizes grow, including unequal sample sizes.
The relative ridge gives scalar-rescaling equivariance for fixed $\alpha$ away from the numerical floors.
Re-estimating $\alpha$ can alter its block weighting, so the prescribed modality scales remain part of the estimator.

Retain the $k_g\leq\min(256,d_{A,g},d_{B,g})$ directions with the largest singular values per component, where $d_{X,g}$ is the width of component $g$ of $X$.
A single-vector pivot further caps this count at $\max(1,m-1)$, where $m$ counts rounded training pivot states (Sec.~\ref{sec:app-states}).
Exactly $m$ distinct pivot vectors give centered covariance rank at most $m-1$ because their weighted centered sum vanishes.
Rounding can merge nearby vectors, so this is a numerical rank convention.
For $m=1$, retain one direction; an exactly constant pivot gives zero cross-covariance and correlation score.

For each component, let
\Needspace{8\baselineskip}
\begin{equation}
\begin{aligned}
H_g&=\tfrac12(\widetilde\Sigma_{PP,g}^{(AP)}+\widetilde\Sigma_{PP,g}^{(PB)}),\\
K_g&=\widetilde\Sigma_{AP,g}(H_g+\rho_{P,g}I)^{-1}\widetilde\Sigma_{PB,g},\\
W_{X,g}&=(\widetilde\Sigma_{XX,g}+\rho_{X,g}I)^{-1/2},\\
W_{A,g}K_gW_{B,g}&=U_g\operatorname{diag}(\sigma_g)V_g^\top.
\end{aligned}
\label{eq:app-compose}
\end{equation}
With $U_g,V_g,\sigma_g$ restricted to the $k_g$ retained directions, the centered coordinates are $u_g=U_g^\top W_{A,g}\bar a_g$ and $v_g=V_g^\top W_{B,g}\bar b_g$.
The contraction below explains why the retained singular values are bounded despite sampling differences between the observed edges.

\subsection{Singular-Value Bound}
\label{sec:app-bound}
\label{sec:score-proof}
\begin{lemma}[Contraction of equal-edge composition]
\label{lem:contraction}
For a fixed part, suppose the two shrunk joint edge covariances are positive semidefinite.
Let their pivot blocks be $C_P^{(AP)},C_P^{(PB)}$, their endpoint blocks be $C_A,C_B$, and their cross-blocks be $C_{AP},C_{PB}$.
Set $M_P=(C_P^{(AP)}+C_P^{(PB)})/2+\rho_P I\succ0$ and
$W_X=(C_X+\rho_X I)^{-1/2}$ for nonnegative ridges and positive definite regularized endpoint blocks.
Every singular value of
$W_A C_{AP}M_P^{-1}C_{PB}W_B$ is at most one with equal-edge pivot averaging.
\end{lemma}
\begin{proof}
Define
\begin{equation}
\begin{aligned}
\Gamma_A&=W_A C_{AP}M_P^{-1/2},\quad
\Gamma_B=M_P^{-1/2}C_{PB}W_B,\\
\Xi_E&=M_P^{-1/2}C_P^{(E)}M_P^{-1/2},\quad
E\in\{AP,PB\}.
\end{aligned}
\end{equation}
The Schur complements of the edge matrices after adding their endpoint ridges give
$\Gamma_A^\top\Gamma_A\preceq\Xi_{AP}$ and $\Gamma_B\Gamma_B^\top\preceq\Xi_{PB}$.
Also
\begin{equation}
\Xi_{AP}+\Xi_{PB}=2I-2\rho_P M_P^{-1}\preceq2I.
\end{equation}
Using these inequalities and equality of the nonzero spectra of the two Gram products,
\begin{equation}
\begin{aligned}
\|\Gamma_A\Gamma_B\|_2^2
&\leq\lambda_{\max}(\Xi_{PB}^{1/2}\Xi_{AP}\Xi_{PB}^{1/2})\\
&\leq\lambda_{\max}(2\Xi_{PB}-\Xi_{PB}^2)\leq1,
\end{aligned}
\end{equation}
since $2\Xi_{PB}-\Xi_{PB}^2=I-(\Xi_{PB}-I)^2\preceq I$.
\end{proof}

The lemma allows the two empirical pivot covariances to differ.
Convex joint shrinkage ensures its positive-semidefinite premise in exact arithmetic.
Singular-value clipping to $[0,1-10^{-6}]$ keeps $\ell_g$ away from a zero denominator and handles numerical departures.

\subsection{Likelihood Ratio in Singular Coordinates}
\begin{lemma}[Likelihood ratio in singular coordinates]
\label{lem:gaussian}
Let $(A,P,B)$ be jointly Gaussian, with invertible endpoint and pivot covariances and $\Sigma_{AB\cdot P}=0$.
Use population moments, no shrinkage or ridge, and the singular vectors of
$\Sigma_{AA}^{-1/2}\Sigma_{AP}\Sigma_{PP}^{-1}
\Sigma_{PB}\Sigma_{BB}^{-1/2}$.
For retained singular values $0\leq\sigma_k<1$, the log-density ratio of the retained endpoint coordinates against their independent marginal product is
\begin{equation}
\begin{split}
\log\frac{f(u,v)}{f(u)f(v)}
={}&-\frac12\sum_k\log(1-\sigma_k^2)\\
&+\sum_k\frac{\sigma_k u_kv_k-
\tfrac12\sigma_k^2(u_k^2+v_k^2)}{1-\sigma_k^2}.
\end{split}
\label{eq:supp-llr}
\end{equation}
\end{lemma}
\begin{proof}
Zero residual makes the composed cross-covariance equal to $\Sigma_{AB}$.
Population whitening and the singular vectors give
\begin{equation}
\operatorname{Cov}\!\begin{pmatrix}u\\v\end{pmatrix}
=\begin{bmatrix}I&\operatorname{diag}(\sigma)\\\operatorname{diag}(\sigma)&I\end{bmatrix}.
\end{equation}
Reordering coordinates yields independent Gaussian pairs with covariance
$\left[\begin{smallmatrix}1&\sigma_k\\\sigma_k&1\end{smallmatrix}\right]$.
Inverting each two-dimensional covariance and subtracting the independent-marginal log-density gives Eq.~\ref{eq:supp-llr}.
\end{proof}

The likelihood statement applies to the retained coordinates.
At $\sigma_k=1$ the corresponding Gaussian is singular and the ordinary density formula requires a limiting treatment.
For fixed fitted parameters, the determinant term is constant across pairs and is omitted for ranking.

The complete correlation score is $s_C=\ell_0+\omega\sum_{g=1}^G\ell_g$.
Set $\omega=1/G$ when exactly one of $A,P,B$ has multiple local vectors before broadcasting, and $\omega=1$ otherwise.
In the first case, the local routes reuse the other two global vectors; averaging normalizes that reuse.
The separate global component retains its own moments and score.

\Needspace*{3\baselineskip}

\section{Base Scores, Auxiliary Scores, and Task Decisions}
\label{sec:app-pivot}
The covariance score from covariance-weighted pivot maps completes the base scores of multi-granularity score fusion (MSF); the pivot-association and class-semantic scores from repeated states form the auxiliary scores. Their separate definitions lead to the retrieval and classification decisions below after score standardization.

\label{sec:state-details}
\subsection{Covariance-Weighted Pivot Maps}
The covariance maps use each edge's unshrunk moments and its own pivot covariance.
For $E_A=AP$ and $E_B=PB$, define
\begin{equation}
\phi_{X,g}(x)=
\left(\widehat\Sigma_{PP,g}^{(E_X)}+\rho(\widehat\Sigma_{PP,g}^{(E_X)})I\right)^{-1/2}
\widehat\Sigma_{PX,g}^{(E_X)}\bar x_g,
\label{eq:app-covmap}
\end{equation}
with the same relative ridge convention as the correlation route.
Write
$M_A=\widehat\Sigma_{PP}^{(AP)}+\rho(\widehat\Sigma_{PP}^{(AP)})I$ and
$M_B=\widehat\Sigma_{PP}^{(PB)}+\rho(\widehat\Sigma_{PP}^{(PB)})I$ for one part.
The maps' raw inner product is
\begin{equation}
\phi_A(a)^\top\phi_B(b)=
\bar a^\top\widehat\Sigma_{AP}^{(AP)}
M_A^{-1/2}M_B^{-1/2}
\widehat\Sigma_{PB}^{(PB)}\bar b.
\end{equation}
If $M_A=M_B$, this equals the ridge-composed bilinear form using unshrunk moments.
Endpoint whitening is absent, retaining variance weighting and avoiding endpoint precision estimation.
The cosine further normalizes the mapped magnitudes.
This branch's edge-specific inverse square roots and unshrunk moments generally differ from those used in the correlation branch.
Each mapped vector is divided by $\max(\|\phi\|_2,10^{-12})$.
The implementation computes these maps in raw feature units; scalar modality scaling cancels in the cosine away from numerical floors.
All $G$ local cosines and the cosine of the global component have equal weight:
\begin{equation}
s_V(a,b)=\frac{1}{G+1}\sum_{g=0}^{G}
\operatorname{cos}(\phi_{A,g}(a),\phi_{B,g}(b)).
\end{equation}
The empirical repeat index $w$ (Eq.~\ref{eq:app-weights}) controls this complementary score's contribution; its role is assessed by ablation of the covariance score component.

\subsection{Pivot States and Discriminant Predictors}
\label{sec:app-states}
Round each coordinate of the official global pivot embedding to four decimal places and group identical rounded vectors.
The count $m$ includes both fitting collections, and $\mathcal K_\cap$ contains states present on both edges.
State predictors use only rows carrying these shared states.
For each endpoint, the official global vector and, if there are multiple local vectors, their mean form separate voters.
A global endpoint has one voter.

Within each voter, coordinates are centered and divided by one scalar training standard deviation.
Linear discriminant analysis~\cite{fisher,friedman} estimates state means with a common within-state covariance.
That covariance is shrunk toward a scaled identity by a Ledoit--Wolf variance ratio~\cite{ledoitwolf}, with an additional $10^{-6}$ diagonal factor.

For voter $e$, write its standardized descriptor as $\tilde x$, its state means as $\mu_\kappa$, its regularized within-state covariance as $C_e$, and the edge-specific empirical state prior as $\pi_{X\kappa}$.
Its posterior is
\begin{equation}
\begin{aligned}
p_{Xe}(\kappa\mid x)&=\operatorname{softmax}_{\kappa}\bigl[\tilde x^\top C_e^{-1}\mu_\kappa\\
&\quad-\tfrac12\mu_\kappa^\top C_e^{-1}\mu_\kappa+\log\pi_{X\kappa}\bigr].
\end{aligned}
\end{equation}
The regularized covariance is $C_e=(1-\beta)C+(\beta+10^{-6})\theta I$, where $C$ is the covariance estimate before shrinkage, $\theta=\operatorname{tr}(C)/\dim(C)$, and $\beta\in[0,1]$ is the voter-specific Ledoit--Wolf shrinkage intensity.
If the within-state residual energy is at most $10^{-12}$ times the total descriptor energy, or the number of states is at least the number of fitting rows minus one, total covariance is used.
For a constant descriptor, the posterior reduces to the empirical state prior.
The voter-specific $\beta$ is separate from the cross-part intensity $\alpha$.

\subsection{Cross-Fitted Posterior Pooling}
Let $p_{Xe}(\kappa\mid x)$ denote the posterior predicted by voter $e$.
With nonnegative $\gamma_{Xe}$ summing to one, the log-linear pool~\cite{heskes} simplifies as
\begin{equation}
\pi_{X\kappa}\prod_e
\left(\frac{p_{Xe}(\kappa\mid x)}{\pi_{X\kappa}}\right)^{\gamma_{Xe}}
=\prod_e p_{Xe}(\kappa\mid x)^{\gamma_{Xe}}.
\end{equation}
Normalizing over $\kappa\in\mathcal K_\cap$ gives $q_X$.
Two-fold cross-fitting~\cite{wolpert} splits that endpoint's shared pivot-state samples using seed 0.
Each fold contributes held-out rows whose states occur in its fitting fold.
A softmax parameterization of $\gamma_X$ is fitted by minimizing their combined negative log-likelihood with L-BFGS for at most 200 iterations.
A single voter receives weight one; with no eligible held-out rows, multiple voters receive equal weights.
The voters are then refitted on all shared pivot-state samples from that endpoint.

Let $N_{E\kappa}$ count state $\kappa$ on edge $E$ and
$n_E^\cap=\sum_{\kappa\in\mathcal K_\cap}N_{E\kappa}$.
The compatibility prior is
\Needspace{5\baselineskip}
\begin{equation}
\pi_\kappa=
\frac{N_{AP,\kappa}+N_{PB,\kappa}}{n_{AP}^\cap+n_{PB}^\cap}.
\end{equation}
Its denominator includes shared pivot-state samples only.
The pivot-association score $s_T=\log\sum_{\kappa\in\mathcal K_\cap}q_Aq_B/\pi_\kappa$
retains a prior correction after posterior pooling.
Empty shared pivot-state support gives $o=0$ (Eq.~\ref{eq:app-weights}) and skips this pivot-association score component.

Writing $q_X^\star(\kappa\mid x)=\Pr_\pi(\kappa\mid x)$ for the true state posterior under a common prior $\pi$, conditional independence given $\kappa$ yields
\begin{equation}
\begin{aligned}
\frac{f_{\rm match}(a,b)}{f_A(a)f_B(b)}
&=\frac{\sum_\kappa\pi_\kappa f_A(a\mid\kappa)f_B(b\mid\kappa)}{f_A(a)f_B(b)}\\
&=\sum_\kappa\frac{q_A^\star(\kappa\mid a)q_B^\star(\kappa\mid b)}{\pi_\kappa}.
\end{aligned}
\end{equation}
At scoring time, PRAM substitutes the fitted predictors $q_A,q_B$ and evaluates the logarithm of this sum by log-sum-exp as a compatibility statistic.

\subsection{Class-Prompt States and Class Semantics}
\label{sec:app-class-semantics}
\label{sec:app-template}
Class prompts define repeated $A$ states even when their paired pivots have distinct vectors.
Group rounded global $A$ embeddings into states $\nu$ by the same four-decimal convention.
Using $\mathcal D_{AP}$, fit a pivot predictor of these states with the global and mean-local voters defined above.
For each pivot in $\mathcal D_{PB}$, its maximum-posterior prediction assigns a state to the paired $B$ sample.
Let $\mathcal A_\cap$ contain states represented in both the observed $A$ assignments and the predicted $B$ assignments.
On this support, fit endpoint posteriors $q_A^A(\nu\mid a),q_B^A(\nu\mid b)$ using the same discriminant predictors and cross-fitted pooling.
Both endpoint posteriors are fitted, including the posterior for a class prompt.
The frequency $\pi_\nu^A$ pools observed $A$ and predicted $B$ state counts over this shared support, following the convention for $\pi_\kappa$.
The class-semantic score is
\begin{equation}
s_A(a,b)=\log\sum_{\nu\in\mathcal A_\cap}
\frac{q_A^A(\nu\mid a)q_B^A(\nu\mid b)}{\pi_\nu^A}.
\label{eq:app-template}
\end{equation}

Let $m_A$ count the class-prompt states in the $n_A$ training rows of $\mathcal D_{AP}$.
Two-fold pivot-to-state prediction gives held-out accuracy $\mathrm{acc}_{\rm cv}$, using a seed-7 permutation for the outer split and the preceding voter-weight procedure within each fitting fold.
Held-out rows whose states are absent from their fitting fold count as errors.
Define
\begin{equation}
\begin{aligned}
w_A&=\max\{0,1-\log(m_A)/\log(n_A)\},\\
\chi_A&=\operatorname{clip}_{[0,1]}
\left(\frac{\mathrm{acc}_{\rm cv}-1/m_A}{1-1/m_A}\right),\\
\tau_A&=w_A\chi_A.
\end{aligned}
\label{eq:app-template-weight}
\end{equation}
Set $\tau_A=0$ outside class-prompt classification, for $n_A<3$ or $m_A<2$, or if the shared support is empty.
All assignments, reliability estimates, and posterior fits use the fitting collections.
This class-prompt route carries category information through the pivot without requiring repeated pivot samples across the two observed edges.

\subsection{Training Reference Distributions}
Except in the shared/separate-collection control (Sec.~\ref{sec:collection-overlap-control}), each bank contains at most 2,000 rows sampled without replacement from its own endpoint's edge using seed 0.
Correlation and covariance scores use symmetric endpoint-dependent bank standardization.
Means and sample standard deviations are computed against the opposite endpoint's bank, with deviations floored at $10^{-12}$.
Pivot-association and class-semantic scores use one mean and sample standard deviation over the Cartesian bank product.
Each of $s_T$ and $s_A$ has its own overall moments; the affine transformation preserves its ordering.
All fitted parameters and bank samples are fixed before test scoring.

For $s\in\{s_C,s_V\}$, the complete bank transformation is
\begin{equation}
\mathcal Z_s(a,b)=\tfrac12\left[
\frac{s(a,b)-\mu_s^A(a)}{\varsigma_s^A(a)}+
\frac{s(a,b)-\mu_s^B(b)}{\varsigma_s^B(b)}\right],
\label{eq:app-bank}
\end{equation}
where the first mean and sample deviation range over $b'\in\mathcal B_B$ and the second over $a'\in\mathcal B_A$.
Candidate-dependent moments can be cached without using test matches or labels.
Changing the test-gallery size leaves each fixed query--candidate score unchanged because the reference banks remain fixed after sampling from training data.

\subsection{Repeat Index, Coverage, and Task Scores}
The repeat index $w$ describes distinct-state counts relative to sample counts.
The shared-state coverage $o$ is the equal-edge mean fraction of rows carrying a shared state.
Let $n=n_A+n_B$ and $m$ count distinct rounded training pivot states.
With $\mathcal K_\cap$ the states represented on both edges,
\begin{equation}
\begin{aligned}
w&=\max\{0,1-\log(m)/\log(n)\},\\
o&=\tfrac12\sum_{E\in\{AP,PB\}}
\frac{\#\{i\in E:\kappa(p_i)\in\mathcal K_\cap\}}{n_E}.
\end{aligned}
\label{eq:app-weights}
\end{equation}
Here $\kappa(p)$ denotes pivot $p$'s state under the above rounding rule (four decimal places).
Set $w=0$ for $n<3$.
Define the overall standardization $\overline{\mathcal Z}_T=(s_T-\mu_{s_T})/\max(\varsigma_{s_T},10^{-12})$, where $\mu_{s_T}$ and $\varsigma_{s_T}$ are the mean and sample standard deviation over $\mathcal B_A\times\mathcal B_B$.
Define $\overline{\mathcal Z}_A$ analogously for $s_A$. Then
\begin{equation}
\begin{aligned}
R&=\mathcal Z_{s_C}+w\mathcal Z_{s_V},\\
S_0&=(1-o)R+o\overline{\mathcal Z}_T,\\
S&=(1-\tau_A)S_0+\tau_A\overline{\mathcal Z}_A.
\end{aligned}
\label{eq:app-readouts}
\end{equation}
The class-semantic standardization $\overline{\mathcal Z}_A$ uses the same bank convention as $\overline{\mathcal Z}_T$.
The intermediate classification score $S_0$ uses $R$ when $o=0$ and $\overline{\mathcal Z}_T$ when $o=1$.
The class-semantic score affects class-prompt classification through $\tau_A$; retrieval and label transfer retain $R$ and $S_0$, respectively.
Zero-weight components are skipped.

Table~\ref{tab:app-controls} summarizes the fixed numerical and decision conventions for fitting and test scoring.
\begin{table*}[t]
\centering
\caption{\textbf{Fixed Estimator and Decision Conventions.} Each statistic uses its declared fitting collection.}
\label{tab:app-controls}
\fontsize{8.0}{9.6}\selectfont
\setlength{\tabcolsep}{2pt}
\begin{tabularx}{\textwidth}{@{}>{\raggedright\arraybackslash}p{.28\textwidth}*{2}{Y}@{}}
\toprule
Quantity & Value & Use \\
\midrule
Relative Ridge & $\lambda=0.5$ & All Covariance Routes \\
Projection Directions & At Most 256 per Component & Single-Vector-Pivot Rank Cap Also Applies \\
Reference Banks & At Most 2,000 Rows per Endpoint & Seed 0; without Replacement \\
Labeled Gallery & At Most 10,000 Training Rows & Seed 0; Ten-Neighbor Vote \\
Pivot-State Quantization & Four Decimal Places & Official Global Pivot Embeddings \\
Posterior-Weight Fitting & Two Folds; 200 L-BFGS Steps at Most & Seed 0; Nonnegative Unit-Sum Weights \\
Numerical Scale Floor & $10^{-12}$ & Before Square Roots or Divisions \\
Correlation Cap & $1-10^{-6}$ & Avoids Singular Gaussian Coordinates \\
\bottomrule
\end{tabularx}
\end{table*}

\Needspace*{3\baselineskip}

\section{Data and Evaluation Protocols}
\label{sec:full-protocols}
\label{sec:eval-details}
This section specifies the frozen features, modality--pivot fitting data, and decision rules shared by all methods; complete comparisons follow in Secs.~\ref{sec:full-controls}--\ref{sec:app-comparisons}.
Retrieval is scored by Recall@$K$ (R@$K$), the fraction of queries with a positive within rank cutoff $K$, with ties evaluated in expectation (Eqs.~\ref{eq:app-tie-instance} and~\ref{eq:app-tie-category}).

\subsection{Datasets and Available Pairs}
Table~\ref{tab:supp-protocol} lists all 20 main retrieval configurations across ImageBind and LanguageBind.
Counts refer to samples on the two fitting edges and in the held-out evaluation set.
Where aligned training samples are divided between edges, the assignment splits a seed-0 permutation into two index-disjoint halves.
Provided separate collections are retained directly.
All encoders remain frozen for PRAM and the comparison methods during fitting and scoring.

\begin{table*}[t]
\centering
\caption{\textbf{Main Retrieval Configurations.} Counts are samples on the two fitting edges and in the test set. IB: ImageBind; LB: LanguageBind; Im: RGB image; V: video; Au: audio; D: depth; Th: thermal; Tx: text; I: IMU. Dataset short names: BatVision (BV), Ego4D (E4D), MAVD (MAV), UTD-MHAD (UTD), TartanRGBT (TR), VGGSound (VGG), CART (CT), NYUv2 (NYU), UCF101 (UCF), and MSR-VTT (MSR), as in the main text.}
\label{tab:supp-protocol}
\fontsize{8.0}{9.6}\selectfont
\setlength{\tabcolsep}{2pt}
\begin{tabularx}{\textwidth}{@{}c>{\raggedright\arraybackslash}p{.23\textwidth}>{\centering\arraybackslash}p{.15\textwidth}>{\centering\arraybackslash}p{.14\textwidth}r>{\raggedright\arraybackslash}X@{}}
\toprule
Model & Dataset and Endpoints & Pivot & AP / PB & Test & Retained Parts \\
\midrule
IB & BatVision~\cite{batvision} Au--D & RGB Image & 1,268 / 1,268 & 584 & Audio Global $\times$ $7\times7$ Depth Cells \\
IB & Ego4D~\cite{ego4d} Au--I & Video & 569 / 569 & 487 & 5 Time Segments \\
IB & MAVD~\cite{mavd} Au--Th & RGB Image & 323 / 324 & 432 & Audio Global $\times$ $7\times7$ Thermal Cells \\
IB & UTD-MHAD~\cite{utdmhad} D--I & RGB Video & 215 / 216 & 430 & 8 Time Segments \\
IB & TartanRGBT~\cite{anythermal} D--Th & RGB Image & 4,301 / 4,301 & 8,601 & $6\times7$ Cells \\
IB & VGGSound~\cite{vggsound} Tx--Au & Video Frames & 750 / 750 & 1,000 & Text $\times$ 5 Audio Segments \\
IB & Ego4D~\cite{ego4d} Tx--I & Video & 800 / 800 & 686 & Text $\times$ 5 Time Segments \\
IB & TartanRGBT~\cite{anythermal} Tx--Th & RGB Image & 1,663 / 1,663 & 1,791 & Text $\times$ $6\times7$ Thermal Cells \\
IB & CART~\cite{cart} Th--I & RGB Image & 122 / 123 & 245 & IMU Global $\times$ $7\times7$ Thermal Cells \\
IB & NYUv2~\cite{nyuv2} Tx--D & RGB Image & 795 / 47,584 & 654 & Text $\times$ $7\times7$ Depth Cells \\
LB & BatVision~\cite{batvision} D--Au & Scene Prompt & 1,268 / 1,268 & 584 & Audio Global $\times$ $8\times8$ Depth Cells \\
LB & VGGSound~\cite{vggsound} Im--Au & Class Prompt & 750 / 750 & 1,000 & Audio Global $\times$ $8\times8$ Image Cells \\
LB & NYUv2~\cite{nyuv2} Im--D & Scene Prompt & 23,792 / 23,792 & 654 & $8\times8$ Cells \\
LB & TartanRGBT~\cite{anythermal} Im--Th & Trajectory Prompt & 866 / 867 & 436 & $8\times8$ Cells \\
LB & UCF101~\cite{ucf101} Im--V & Action Prompt & 3,613 / 3,613 & 2,874 & $8\times8$ Cells \\
LB & MAVD~\cite{mavd} Th--Au & Drive Prompt & 323 / 324 & 433 & Audio Global $\times$ $8\times8$ Thermal Cells \\
LB & TartanRGBT~\cite{anythermal} Th--D & Trajectory Prompt & 866 / 867 & 436 & $8\times8$ Cells \\
LB & MSR-VTT~\cite{msrvtt} V--Au & Caption & 3,088 / 3,088 & 884 & 4 Time Segments \\
LB & TartanRGBT~\cite{anythermal} V--D & Trajectory Prompt & 866 / 867 & 436 & $8\times8$ Cells \\
LB & TartanRGBT~\cite{anythermal} V--Th & Trajectory Prompt & 866 / 867 & 436 & $8\times8$ Cells \\
\bottomrule
\end{tabularx}
\end{table*}

The two category-retrieval configurations, text--audio on VGGSound and text--thermal on TartanRGBT, use class-consistent positives.
The other main retrieval configurations use paired instances.
Classification evaluates 17 modality relations with category labels.
The main cross-task means average the declared task metrics over all evaluated relations within each binding model.

\subsection{Frozen Feature Construction}
\label{sec:app-features}
ImageBind TartanRGBT depth--thermal uses depth block 6, image block 12, and thermal block 4 for the main random split.
The unseen-trajectory configuration uses depth block 6, image block 12, and thermal block 6.
The text--thermal category configuration uses image block 12 and thermal block 4, with text broadcast across spatial cells.
These selections use validation within the respective modality--pivot edges, with encoders kept frozen.

UTD-MHAD uses eight temporal segments: depth class-token descriptors, RGB block-24 spatial means, and the selected IMU temporal bins.
Ego4D audio--IMU uses five segments: audio class-token descriptors, video block-24 class-token descriptors, and adjacent-pair averaging of the selected gyroscope slots.
The text--IMU configuration broadcasts its text vector across five synchronized segments.
These constructions supply the local vectors of PRAM and its ablations; the sample-level baselines use the official global vectors (Sec.~\ref{sec:comparator-details}).
ImageBind BatVision audio--depth, MAVD audio--thermal, and CART thermal--IMU use $7\times7$ grids for the RGB pivot and spatial endpoint.
ImageBind MAVD depth--thermal (IB MAV D--Th) uses an RGB pivot and $6\times7$ grids with zero registration offsets.
The vision, depth, and thermal features come from blocks 12, 6, and 6, respectively.
Its color-coded depth rendering is inverted to a disparity-like map and normalized per frame between the 1st and 99th percentiles.
In the $7\times7$ configurations, the single-vector audio or IMU endpoint is broadcast across the 49 cells.

LanguageBind spatial configurations use cached block-12 descriptors on an $8\times8$ grid.
The TartanRGBT video descriptors use the spatial orientation of the corresponding depth and thermal samples.
MSR-VTT uses four temporal video and audio segments.
A single-vector endpoint or text pivot is broadcast across the other modality's parts.
Local scores use the structural normalization in Sec.~\ref{sec:estimator-details}.
The official sample-level embeddings are retained separately as the global component with its own moments and score.

\subsection{Spatial Registration}
\label{sec:registration-details}
Each spatial offset is estimated from its own observed edge's fitting frames.
The estimator searches integer shifts within three per axis to maximize normalized cross-covariance strength.
It centers features over at most 2,000 frames sampled with seed 0 on that edge.
TartanRGBT depth maps are the released FoundationStereo~\cite{foundationstereo} estimates from the stereo RGB pair, aligned with the left RGB image.
The MS2 robot localization scenario uses projected LiDAR depth (Sec.~\ref{sec:application-details}).
For ImageBind TartanRGBT, image tokens are first resampled from $16\times16$ to $14\times14$.
The registered overlap retains depth rows 1--13 and image/thermal rows 0--12.
The first twelve overlap rows give a $6\times7$ grid after $2\times2$ averaging.
For IB TR D--Th, $6\times7$ balances accuracy and cost: R@10 75.1, fit 9.4\,s, peak memory 3.62\,GB, versus 75.5, 31.0\,s, 7.04\,GB at $12\times12$ (Sec.~\ref{sec:diagnostic-details}).

\begin{table}[t]
\centering
\caption{\textbf{Registration Offsets.} Offsets $(\Delta y,\Delta x)$ are fitted independently on the two observed edges. All listed pivots are RGB. AT+D uses AnyThermal and DINOv2-B.}
\label{tab:registration-offsets}
\fontsize{8.0}{9.6}\selectfont
\setlength{\tabcolsep}{2pt}
\begin{tabularx}{\linewidth}{@{}l*{2}{Y}@{}}
\toprule
Configuration & $A$--$P$ & $P$--$B$\\
\midrule
IB TR~\cite{anythermal}, Random & $(-1,0)$ & $(0,0)$\\
IB TR~\cite{anythermal}, Trajectory & $(-1,0)$ & $(0,0)$\\
AT+D MS2~\cite{ms2} Depth--Thermal & $(0,0)$ & $(1,0)$\\
AT+D Vehicle~\cite{ms2,vivid} Depth--Thermal & $(0,0)$ & $(2,-1)$\\
\bottomrule
\end{tabularx}
\end{table}

The two TartanRGBT random-split edges contain 4,301 samples each; the trajectory-split edges contain 6,872 each.
Each offset uses 2,000 samples from its respective edge.
MS2 has 489 and 490 fitting samples, all of which enter its offset estimate.
In the vehicle setting, re-estimating the MS2 depth--RGB offset on each condition's fitting edge gives $(0,0)$ in both conditions, while the shared daytime ViViD++ RGB--thermal edge gives $(2,-1)$.
Table~\ref{tab:registration-offsets} records the resulting geometry.

\subsection{Data Splits and NYUv2 Conventions}
The main ImageBind TartanRGBT comparison uses a random sample split.
Its unseen-trajectory control tests on 3,459 samples from trajectories excluded from fitting.
On this split, PRAM reaches 62.7 R@10, and MMBind, the strongest of the 15 comparators, reaches 10.5 (Table~\ref{tab:app-unseen}).
\begin{table}[t]
\centering
\caption{\textbf{Unseen Trajectories.} Bidirectional R@1 and R@10 (\%) on IB TR D--Th for all 15 sample-level baselines, tested on trajectories excluded from fitting. The gray row marks Frozen Cosine as the default reference and is excluded from bold ranking; bold marks the best pivot-only result.}
\label{tab:app-unseen}
\fontsize{8.0}{9.6}\selectfont
\setlength{\tabcolsep}{2pt}
\begin{tabularx}{\linewidth}{@{}l*{2}{Y}@{}}
\toprule
Method & R@1 & R@10 \\
\midrule
\fullrowband{gray!13}
Frozen Cosine~\cite{imagebind} & 0.1 & 0.8 \\
\midrule
Mean Centering & 0.1 & 1.0 \\
Projection Heads & 1.4 & 8.6 \\
\midrule
UCL~\cite{ucl} & 0.0 & 0.3 \\
TextME~\cite{textme} & 0.0 & 0.7 \\
I0T~\cite{i0t} & 0.1 & 0.9 \\
DBSN~\cite{dbsn} & 0.1 & 1.0 \\
C-MCR~\cite{cmcr} & 0.1 & 1.4 \\
Ex-MCR~\cite{exmcr} & 0.3 & 2.0 \\
ASIF~\cite{asif} & 0.4 & 3.5 \\
CMCL~\cite{cmcl} & 0.4 & 4.1 \\
Latent Translation~\cite{latenttranslation} & 0.8 & 5.5 \\
STRUCTURE~\cite{structure} & 1.1 & 8.2 \\
EmergentBridge~\cite{emergentbridge} & 1.2 & 8.9 \\
MMBind~\cite{mmbind} & 1.6 & 10.5 \\
\midrule
\fullrowband{resultshade}
\textbf{PRAM (Ours)} & \textbf{32.4} & \textbf{62.7} \\
\bottomrule
\end{tabularx}
\end{table}
LanguageBind TartanRGBT tests on unseen trajectories throughout.
UCF101 uses the class-balanced release of CMCL~\cite{cmcl}, 100 clips per class, divided by official split 1: source groups 8--25 supply the 7,226 training samples and groups 1--7 the 2,874 test samples.
These groups separate source videos between training and testing.
MAVD fits on the first 60\% of each drive in time order (647 night frames for both backbones) and tests on the remaining frames; the ImageBind test set omits one frame whose depth image is empty, leaving fewer frames than LanguageBind.

ImageBind NYUv2 has two distinct protocols, summarized in Table~\ref{tab:app-nyu}.
Main retrieval uses 795 instance-caption--RGB fitting pairs and 47,584 RGB--depth pairs.
Main classification instead uses scene prompts and 47,584 aligned training samples divided into two 23,792-sample edges.
Both use the same 654 held-out depth images.
Scene-category retrieval in Figure~7 of the main text and caption-based classification in Table~\ref{tab:app-nyu} use their respective positives and decision rules.

\begin{table*}[t]
\centering
\caption{\textbf{ImageBind NYUv2 Conventions.} Both use the same 654 test depth images. Captions define instance positives; scene prompts define category positives. Training collections and classification decisions differ. Acc.: top-1 accuracy; Bal.: balanced accuracy.}
\label{tab:app-nyu}
\fontsize{8.0}{9.6}\selectfont
\setlength{\tabcolsep}{2pt}
\begin{tabularx}{\textwidth}{@{}>{\raggedright\arraybackslash}p{.25\textwidth}*{5}{Y}@{}}
\toprule
Text and Task & AP / PB & R@1 & R@10 & Acc. & Bal. \\
\midrule
Caption Instance & 795 / 47,584 & 3.1 & 18.7 & 60.5 & 42.0 \\
Scene Category & 23,792 / 23,792 & 56.0 & 64.2 & 40.4 & 31.4 \\
\bottomrule
\end{tabularx}
\end{table*}

\subsection{Retrieval Metrics and Tied Scores}
For instance retrieval, each query has one paired positive.
Let $n_>$ count candidates with strictly greater scores than that positive and $n_=$ count candidates tied with it, including the positive.
Under uniform ordering within the tied block, its expected R@$K$ contribution is
\begin{equation}
\operatorname{clip}_{[0,1]}\!\left(\frac{K-n_>}{n_=}\right).
\label{eq:app-tie-instance}
\end{equation}
We average this expected recall over queries and both directions, removing dependence on arbitrary tie ordering.
All means, differences, ratios, and percentage gains use unrounded values and are rounded only for display.

Category retrieval uses any same-class candidate as a positive.
Let the first tied block containing positives have $n_=$ candidates, $n_=^+$ of them positive, and let $n_>$ candidates precede it.
For $K'=\operatorname{clip}_{[0,n_=]}(K-n_>)$, the hit probability from this block is
\begin{equation}
1-\frac{\binom{n_=-n_=^+}{K'}}{\binom{n_=}{K'}},
\label{eq:app-tie-category}
\end{equation}
with probability zero for $K'=0$ and the usual zero numerator when $K'>n_=-n_=^+$.
The implementation evaluates this tied-block expectation directly.

\subsection{Classification and Model Selection}
Category-level text configurations compare one prompt per class using $S$.
The remaining category-transfer tasks use ten-neighbor label transfer from a labeled training gallery in the opposite modality; the two directions are averaged.
The gallery is sampled from both training halves, capped at 10,000 samples with seed 0.
Neighbor ties are broken by a fixed random permutation; tied votes and tied prompt scores are scored by their expected accuracy under random tie-breaking.
Top-1 accuracy reports the fraction of correct predictions; balanced accuracy averages class recall over the evaluated classes.
The same prompt or labeled-gallery decision rule is applied to the corresponding comparator scores on the same test samples.

PRAM fits shared-pivot state predictors and, for class prompts, propagates class-prompt states through the pivot.
The wearable action recognition scenario uses its separate labeled-video and unlabeled video--IMU protocol.

Representation selection for UTD-MHAD, Ego4D audio--IMU, and ImageBind TartanRGBT uses validation within the corresponding modality--pivot edges.
The relative ridge coefficient $\lambda=0.5$ is selected from $\{0.05,0.1,0.25,0.5,1\}$ by held-out retrieval on TartanRGBT modality--pivot pairs, without $A$--$B$ pairs or test data, and used for every relation.
Table~\ref{tab:app-lambda} also reports $\lambda=2$, outside this range, as a sensitivity check.
The projection-direction cap is 256, with the additional single-vector-pivot rank cap defined in Sec.~\ref{sec:estimator-details}.
Table~\ref{tab:app-lambda} varies $\lambda$, the direction cap, the reference-bank size, and the pivot-state decimals one at a time; the other values stay at their defaults.
\begin{table*}[t]
\centering
\caption{\textbf{Hyperparameter Sensitivity.} Mean retrieval R@10 over all 20 settings (Ret.\ Mean) and overall four-column average (Avg., averaging the four one-decimal backbone-task means as in Table~3 of the main text), varying one value at a time; 1k: 1,000 rows. Shaded and bold: default values used throughout.}
\label{tab:app-lambda}
\fontsize{8.0}{9.6}\selectfont
\setlength{\tabcolsep}{0pt}
\begin{tabularx}{\textwidth}{@{}l*{19}{Y}@{}}
\toprule
 & \multicolumn{6}{c}{Relative Ridge $\lambda$} & \multicolumn{4}{c}{Projection Directions} & \multicolumn{4}{c}{Reference Bank} & \multicolumn{5}{c}{Pivot-State Decimals} \\
\cmidrule(lr){2-7}\cmidrule(lr){8-11}\cmidrule(lr){12-15}\cmidrule(lr){16-20}
Metric & 0.05 & 0.1 & 0.25 & \cellcolor{resultshade}\textbf{0.5} & 1 & 2 & 64 & 128 & \cellcolor{resultshade}\textbf{256} & 512 & 500 & 1k & \cellcolor{resultshade}\textbf{2k} & 4k & 2 & 3 & \cellcolor{resultshade}\textbf{4} & 5 & 6 \\
\midrule
Ret.\ Mean & 40.6 & 41.7 & 42.8 & \cellcolor{resultshade}\textbf{43.1} & 42.9 & 42.1 & 43.0 & 43.1 & \cellcolor{resultshade}\textbf{43.1} & 43.1 & 43.1 & 43.1 & \cellcolor{resultshade}\textbf{43.1} & 43.1 & 43.1 & 43.1 & \cellcolor{resultshade}\textbf{43.1} & 43.1 & 43.1 \\
Avg. & 53.1 & 53.9 & 54.3 & \cellcolor{resultshade}\textbf{54.3} & 53.8 & 53.1 & 54.1 & 54.2 & \cellcolor{resultshade}\textbf{54.3} & 54.3 & 54.3 & 54.2 & \cellcolor{resultshade}\textbf{54.3} & 54.2 & 54.3 & 54.2 & \cellcolor{resultshade}\textbf{54.3} & 54.3 & 54.2 \\
\bottomrule
\end{tabularx}
\end{table*}

Main-table runs use fixed settings and seed 0 (seed 7 for $\chi_A$ outer split); pair budgets use seeds 0--1 (0--9 for the ten-subset 2\% check), five-seed runs use 0--4, collection splits use 42--51, and wearable runs average 1--3.

\subsection{Comparator Inputs and Adaptations}

\label{sec:comparator-details}
\noindent\textbf{Sample-Level Baselines.}
All 15 primary baselines use the official global vectors.
Frozen Cosine uses their native coordinates.
Mean centering subtracts training means, and I0T~\cite{i0t} standardizes coordinates before cosine comparison.
DBSN~\cite{dbsn} calibrates the frozen similarity using training banks.
UCL~\cite{ucl} supplies a two-hop probabilistic comparison through training pivots.
Latent Translation~\cite{latenttranslation} fits Procrustes or ridge mappings into pivot coordinates, selecting the mapping and regularization on held-out modality--pivot pairs.
ASIF~\cite{asif} is adapted to separate anchor collections: each endpoint forms a weighted average of the pivot features paired with its own anchors.
The two weighted averages are compared in pivot coordinates.
Training-mean subtraction is applied before normalization.
ASIF selects among six configurations by R@10 on held-out modality--pivot pairs.
C-MCR~\cite{cmcr} selects among four configurations in the same way.
Ex-MCR~\cite{exmcr} extends the overlapping representation through the pivot.
The projection-head baseline fits endpoint heads from the two observed pair collections.
EmergentBridge~\cite{emergentbridge} is evaluated as a head-level adaptation on frozen embeddings.
TextME~\cite{textme} follows the official recipe: for each backbone, a projection head maps text embeddings of 100K COCO~\cite{coco}, AudioCaps~\cite{audiocaps}, and InternVid~\cite{internvid} captions into the Qwen3-Embedding-4B~\cite{qwen3embedding} text space; at test time, each modality is shifted by the difference between its training mean and the text mean before projection, so no pairs are used.
STRUCTURE~\cite{structure}, CMCL~\cite{cmcl}, and MMBind~\cite{mmbind} supply neighborhood-based alignment, sequential contrastive alignment, and pseudo-pair construction, respectively.
Ex-MCR and MMBind are reimplemented for the pivot setting with global vectors.
Other hyperparameter choices use held-out modality--pivot pairs from the two fitting collections.

\Needspace*{3\baselineskip}

\section{Supporting Mechanism Experiments}
\label{sec:full-controls}
This section provides the protocols and numerical results behind Table~4 and Figure~7 of the main text, together with additional analyses of part pairing, misregistered fitting data, and internal states.

\subsection{Representation and Spatial Granularity}
\label{sec:diagnostic-details}
The ImageBind TartanRGBT comparison uses 8,602 training and 8,601 test samples.
Pivot fitting assigns 4,301 samples to each edge; direct fitting uses all 8,602 target pairs with encoders frozen.
The local features use depth block 6, image block 12, and thermal block 4, with the main configuration's relative ridge and direction cap.
Table~\ref{tab:app-representation} compares official global vectors with retained global and local vectors.
The same-layer study in Table~\ref{tab:supp-grid} controls feature construction by varying only the spatial grid, retaining the separate global component with its own score.

\begin{table}[t]
\centering
\caption{\textbf{Representation and Supervision on ImageBind TartanRGBT.} Mean bidirectional R@10 (\%). With encoders frozen, ``Direct target pairs'' fits PRAM's local and global relations directly on true depth--thermal pairs with the same scores and caps, bypassing RGB pivot composition. Shaded: PRAM.}
\label{tab:app-representation}
\fontsize{8.0}{9.6}\selectfont
\renewcommand{\arraystretch}{1.35}
\begin{tabularx}{\linewidth}{@{}l*{2}{Y}@{}}
\toprule
Fitting Data & Global & Global + Local \\
\midrule
\fullrowband{resultshade}
\textbf{Modality--Pivot Pairs} & 11.0 & 75.1 \\
Direct Target Pairs & 19.9 & 79.8 \\
\bottomrule
\end{tabularx}
\end{table}

Each resolution is obtained by adaptive average pooling of the registered $12\times14$ token grid.
The $1\times1$ setting averages all local tokens into one vector; $6\times7$ corresponds to $2\times2$ averaging.
Pooling windows can overlap when the requested grid does not divide the token grid.
The granularity study stores local features in float16.
\begin{table}[t]
\centering
\caption{\textbf{Same-Layer Granularity with the Global Component Retained.} Times come from this study's chunked float16 pipeline on local NVMe storage, not the main pipeline of Sec.~\ref{sec:app-cost}. GB is the peak GPU memory from fitting to scoring all 8,601 test samples; shaded and bold: the main-table grid.}
\label{tab:supp-grid}
\fontsize{8.0}{9.6}\selectfont
\setlength{\tabcolsep}{2pt}
\begin{tabularx}{\linewidth}{@{}l*{6}{Y}@{}}
\toprule
Grid & Parts & R@1 & R@10 & \shortstack{Direct\\R@10} & Fit (s) & GB \\
\midrule
$1\times1$ & 1 & 3.9 & 21.7 & 30.4 & 1.4 & 3.57 \\
$2\times2$ & 4 & 15.5 & 49.6 & 56.6 & 2.0 & 3.58 \\
$3\times3$ & 9 & 27.0 & 61.7 & 67.3 & 3.2 & 3.58 \\
$4\times4$ & 16 & 35.1 & 68.3 & 73.3 & 4.6 & 3.59 \\
$5\times5$ & 25 & 37.0 & 69.3 & 74.0 & 6.3 & 3.60 \\
$6\times6$ & 36 & 43.8 & 73.7 & 78.3 & 8.7 & 3.61 \\
\fullrowband{resultshade}
{\boldmath$6\times7$} & \textbf{42} & 46.8 & 75.1 & 79.8 & 9.4 & 3.62 \\
$7\times7$ & 49 & 45.1 & 74.2 & 78.6 & 11.0 & 3.63 \\
$8\times8$ & 64 & 45.8 & 74.5 & 79.2 & 14.0 & 3.64 \\
$9\times9$ & 81 & 45.3 & 74.3 & 79.0 & 17.5 & 3.96 \\
$10\times10$ & 100 & 46.2 & 74.8 & 79.0 & 22.3 & 4.89 \\
$12\times12$ & 144 & 48.6 & 75.5 & 81.0 & 31.0 & 7.04 \\
$12\times14$ & 168 & 48.4 & 74.1 & 80.6 & 39.8 & 8.21 \\
\bottomrule
\end{tabularx}
\end{table}

\subsection{Retained-Direction Budgets}
Table~\ref{tab:app-direction} gives the eight TartanRGBT settings under the direction-budget protocol; Table~4 of the main text displays five.
All three modalities provide multiple parts, and scoring uses only local correlations.
The ImageBind configurations fix RGB, depth, and thermal blocks at 12, 6, and 4 and exchange their endpoint and pivot roles.
LanguageBind video--thermal and video--depth use RGB pivots.

Coarser grids average adjacent native cells with identical groups in both endpoints and the pivot.
All settings use $6\times7$, $3\times4$, $2\times2$, and $1\times1$ grids; LanguageBind merges its native $8\times8$ cells into these grids.
Ordered groups differ in size by at most one; seven columns split into four groups have sizes 2, 2, 2, and 1.
The total cap distributes at most 256 directions across parts, while the per-part cap permits at most 256 directions in every part.
The ``No Quadratic'' column of Table~\ref{tab:app-direction} reports R@10 under the total cap after removing the quadratic endpoint-magnitude terms from $\ell_g$, as in Sec.~\ref{sec:app-ablations}; the best of 4, 12, and 42 parts still exceeds the 1-part grid in all eight settings by 9.3--45.2 points.
At 42 parts, the per-part cap gains 14.9--18.1 points in the five main-paper Table~4(b) settings; the other three (LB Im--Th, Im--D, Th--D) already reach 92.9--97.4 under the total cap and 98.3--99.7 under the per-part cap.
\begin{table*}[t]
\centering
\caption{\textbf{Direction Budgets.} Eight TartanRGBT settings; Table~4 of the main text shows five. The total cap retains at most 256 directions across parts; the per-part cap retains at most 256 per part. Both exclude the global component. ``No Quadratic'' gives R@10 under the total cap with the endpoint-magnitude terms removed from $\ell_g$. Shaded: per-part cap.}
\label{tab:app-direction}
\fontsize{8.0}{9.6}\selectfont
\setlength{\tabcolsep}{2.2pt}
\renewcommand{\arraystretch}{0.96}

\resizebox{\textwidth}{!}{\begin{tabular}{@{}lc@{\hspace{4pt}}c*{3}{c}*{2}{>{\columncolor{resultshade}}c}@{\hspace{1em}}lc@{\hspace{4pt}}c*{3}{c}>{\columncolor{resultshade}}c>{\columncolor{resultshade}[\tabcolsep][0pt]}c@{}}
\toprule
Relation & Pivot & Parts & \multicolumn{3}{c}{Total Cap} & \multicolumn{2}{c}{Per-Part Cap} & Relation & Pivot & Parts & \multicolumn{3}{c}{Total Cap} & \multicolumn{2}{c@{}}{Per-Part Cap} \\
\cmidrule(lr){4-6}\cmidrule(lr){7-8}\cmidrule(lr){12-14}\cmidrule(l){15-16}
 & & & R@1 & R@10 & \shortstack{No\\Quadratic} & \multicolumn{1}{c}{R@1} & \multicolumn{1}{c@{\hspace{1em}}}{R@10} & & & & R@1 & R@10 & \shortstack{No\\Quadratic} & \multicolumn{1}{c}{R@1} & \multicolumn{1}{c@{}}{R@10} \\
\midrule
\multirow{4}{*}{IB D--Th} & \multirow{4}{*}{RGB} & 1 & 4.4 & 22.9 & 14.0 & 4.4 & 22.9 & \multirow{4}{*}{LB Im--Th} & \multirow{4}{*}{D} & 1 & 44.0 & 87.0 & 79.0 & 44.0 & 87.0 \\
 &  & 4 & 16.1 & 49.8 & 36.3 & 16.3 & 50.4 &  &  & 4 & 76.7 & 97.9 & 94.7 & 77.2 & 98.2 \\
 &  & 12 & 29.5 & 62.7 & 48.6 & 33.3 & 67.1 &  &  & 12 & 81.7 & 97.4 & 94.8 & 87.3 & 98.6 \\
 &  & 42 & 29.2 & 58.2 & 37.8 & 47.2 & 75.3 &  &  & 42 & 74.7 & 96.4 & 84.3 & 90.6 & 99.1 \\
\cmidrule(r{1em}){1-8}\cmidrule{9-16}
\multirow{4}{*}{IB Im--Th} & \multirow{4}{*}{D} & 1 & 6.6 & 30.5 & 18.4 & 6.6 & 30.5 & \multirow{4}{*}{LB Im--D} & \multirow{4}{*}{Th} & 1 & 48.0 & 91.5 & 87.2 & 48.0 & 91.5 \\
 &  & 4 & 22.8 & 60.5 & 45.3 & 23.4 & 61.2 &  &  & 4 & 78.0 & 98.2 & 98.2 & 78.3 & 98.3 \\
 &  & 12 & 38.8 & 72.2 & 56.3 & 43.6 & 77.0 &  &  & 12 & 85.8 & 98.7 & 97.8 & 88.5 & 99.5 \\
 &  & 42 & 37.2 & 66.8 & 41.5 & 59.5 & 83.9 &  &  & 42 & 77.0 & 97.4 & 90.6 & 94.0 & 99.7 \\
\cmidrule(r{1em}){1-8}\cmidrule{9-16}
\multirow{4}{*}{IB Im--D} & \multirow{4}{*}{Th} & 1 & 8.2 & 36.0 & 21.8 & 8.2 & 36.0 & \multirow{4}{*}{LB V--Th} & \multirow{4}{*}{RGB} & 1 & 36.0 & 75.2 & 71.3 & 36.0 & 75.2 \\
 &  & 4 & 29.4 & 71.0 & 55.2 & 30.1 & 71.7 &  &  & 4 & 48.6 & 84.6 & 80.6 & 51.5 & 85.7 \\
 &  & 12 & 47.9 & 81.6 & 67.0 & 55.5 & 86.4 &  &  & 12 & 47.5 & 81.4 & 75.1 & 58.8 & 87.7 \\
 &  & 42 & 43.7 & 74.2 & 47.9 & 71.3 & 92.3 &  &  & 42 & 39.3 & 72.0 & 58.7 & 63.6 & 89.2 \\
\cmidrule(r{1em}){1-8}\cmidrule{9-16}
\multirow{4}{*}{LB Th--D} & \multirow{4}{*}{RGB} & 1 & 35.9 & 84.6 & 77.5 & 35.9 & 84.6 & \multirow{4}{*}{LB V--D} & \multirow{4}{*}{RGB} & 1 & 27.6 & 75.9 & 69.4 & 27.6 & 75.9 \\
 &  & 4 & 66.5 & 96.0 & 93.9 & 68.2 & 96.2 &  &  & 4 & 45.1 & 86.4 & 81.1 & 48.0 & 88.5 \\
 &  & 12 & 71.7 & 96.0 & 93.6 & 78.2 & 97.8 &  &  & 12 & 41.9 & 83.1 & 75.6 & 52.9 & 88.9 \\
 &  & 42 & 63.1 & 92.9 & 83.7 & 84.2 & 98.3 &  &  & 42 & 30.9 & 73.7 & 59.3 & 53.0 & 88.6 \\
\bottomrule
\end{tabular}}
\par\vspace{6pt}
{\caption{\textbf{Changing Pivot Structure at Fixed Endpoints across Thirteen Settings.} LanguageBind uses RGB pivots. Each condition refits PRAM, including $\omega$, retained rank, $w$, and $o$. CT has no state labels. All values are bidirectional recall (\%); shaded: full pivots.}
\label{tab:supp-pivot}
\fontsize{8.0}{9.6}\selectfont
\setlength{\tabcolsep}{2pt}
\renewcommand{\arraystretch}{1.0}
\begin{tabularx}{\textwidth}{@{}>{\raggedright\arraybackslash}p{.29\textwidth}*{8}{Y}@{}}
\toprule
 & \multicolumn{2}{c}{Full Pivot} & \multicolumn{2}{c}{Mean Parts} & \multicolumn{2}{c}{Shuffled Parts} & \multicolumn{2}{c}{State Label} \\
Relation & R@1 & R@10 & R@1 & R@10 & R@1 & R@10 & R@1 & R@10 \\
\midrule
IB TR~\cite{anythermal} D--Th & \cellcolor{resultshade}46.8 & \cellcolor{resultshade}75.1 & 14.8 & 43.7 & 4.9 & 23.5 & 2.9 & 12.9 \\
LB TR~\cite{anythermal} Th--D & \cellcolor{resultshade}85.6 & \cellcolor{resultshade}98.5 & 51.5 & 89.2 & 22.8 & 69.0 & 16.1 & 53.2 \\
LB TR~\cite{anythermal} V--D & \cellcolor{resultshade}55.2 & \cellcolor{resultshade}89.3 & 34.5 & 77.8 & 18.2 & 60.2 & 14.2 & 50.8 \\
LB TR~\cite{anythermal} V--Th & \cellcolor{resultshade}65.7 & \cellcolor{resultshade}89.9 & 45.2 & 79.2 & 27.2 & 64.4 & 25.3 & 59.4 \\
IB UTD~\cite{utdmhad} D--I & \cellcolor{resultshade}2.6 & \cellcolor{resultshade}22.7 & 2.1 & 16.5 & 1.6 & 13.6 & 4.0 & 27.6 \\
IB CT~\cite{cart} Th--I & \cellcolor{resultshade}2.2 & \cellcolor{resultshade}19.8 & 2.0 & 16.9 & 2.0 & 14.5 & -- & -- \\
IB NYU~\cite{nyuv2} Tx--D & \cellcolor{resultshade}3.1 & \cellcolor{resultshade}18.7 & 2.5 & 18.8 & 2.5 & 16.9 & 1.0 & 8.3 \\
IB E4D~\cite{ego4d} Au--I & \cellcolor{resultshade}1.6 & \cellcolor{resultshade}10.9 & 1.9 & 10.7 & 2.0 & 10.1 & 0.8 & 7.8 \\
IB E4D~\cite{ego4d} Tx--I & \cellcolor{resultshade}0.8 & \cellcolor{resultshade}6.9 & 1.0 & 6.7 & 1.0 & 6.6 & 0.5 & 6.0 \\
IB BV~\cite{batvision} Au--D & \cellcolor{resultshade}1.4 & \cellcolor{resultshade}11.1 & 1.2 & 10.9 & 0.8 & 9.9 & 0.8 & 7.3 \\
LB BV~\cite{batvision} D--Au & \cellcolor{resultshade}1.7 & \cellcolor{resultshade}12.2 & 1.6 & 12.4 & 1.3 & 11.6 & 0.6 & 6.8 \\
IB MAV~\cite{mavd} Au--Th & \cellcolor{resultshade}0.6 & \cellcolor{resultshade}5.1 & 0.8 & 5.2 & 0.9 & 4.1 & 0.0 & 2.0 \\
LB MAV~\cite{mavd} Th--Au & \cellcolor{resultshade}0.4 & \cellcolor{resultshade}4.5 & 0.7 & 5.0 & 0.8 & 4.6 & 0.5 & 3.0 \\
\bottomrule
\end{tabularx}
}
\end{table*}

\begin{table}[t]
\centering
\caption{\textbf{LB TR Th--D in Tables~2 and~4(c) of the Main Text.} Both use the same endpoints, split, and test set.}
\label{tab:app-lbtr-config}
\fontsize{8.0}{9.6}\selectfont
\setlength{\tabcolsep}{2pt}
\begin{tabularx}{\linewidth}{@{}l*{2}{Y}@{}}
\toprule
 & Table~2 & Table~4(c) \\
\midrule
Pivot & Trajectory Text & RGB Image \\
Pivot Parts & 1 (Broadcast) & 64 \\
Endpoint Parts & 64 / 64 & 64 / 64 \\
Cross-Part Intensity $\alpha$ & 0.071 & 0.079 \\
Direction Cap & 36 & 256 \\
Repeat Index $w$ & 0.52 & 0 \\
State Coverage $o$ & 0.995 & 0 \\
Split & Unseen Trajectories & Unseen Trajectories \\
R@10 & 54.6 & 98.5 \\
\bottomrule
\end{tabularx}
\end{table}
\subsection{Pivot Structure}
\begin{table*}[t]
\centering
\caption{\textbf{Shared and Separate Fitting Collections.} R@10 averages ten splits and both half assignments. Cos.: Frozen Cosine R@10; $^*$: category retrieval. Brackets give the minimum and maximum drop across splits. Kept is the fraction of the gain over Frozen Cosine; in the mean row, the fraction of the mean gain.}
\label{tab:app-collection-overlap}
\fontsize{8.0}{9.6}\selectfont
\setlength{\tabcolsep}{1.4pt}
\renewcommand{\arraystretch}{1.0}
\begin{tabular*}{\textwidth}{@{\extracolsep{\fill}}llc*{3}{c}cc@{}}
\toprule
Dataset & Pair & Pivot & Cos. & \shortstack{Shared\\Collection} & \shortstack{Separate\\Collections} & Drop [Range] & Kept (\%) \\
\midrule
\multicolumn{8}{l}{\textbf{ImageBind}~\cite{imagebind}} \\
BatVision~\cite{batvision} & Au--D & RGB & 1.97 & 11.10 & 11.03 & 0.07 [\textminus 0.43, 0.55] & 99.2 \\
Ego4D~\cite{ego4d} & Au--I & Video & 1.75 & 11.54 & 9.56 & 1.99 [0.47, 3.03] & 79.7 \\
MAVD~\cite{mavd} & Au--Th & RGB & 1.56 & 4.76 & 4.59 & 0.17 [\textminus 0.35, 0.54] & 94.7 \\
UTD-MHAD~\cite{utdmhad} & D--I & Video & 3.14 & 26.39 & 22.95 & 3.44 [2.15, 4.31] & 85.2 \\
TartanRGBT~\cite{anythermal} & D--Th & RGB & 0.34 & 74.85 & 75.28 & \textminus 0.43 [\textminus 0.58, \textminus 0.28] & 100.6 \\
VGGSound~\cite{vggsound} & Tx--Au$^*$ & Video & 75.95 & 84.70 & 81.97 & 2.73 [2.35, 3.23] & 68.7 \\
Ego4D~\cite{ego4d} & Tx--I & Video & 1.53 & 7.56 & 6.62 & 0.94 [\textminus 0.03, 1.54] & 84.4 \\
TartanRGBT~\cite{anythermal} & Tx--Th$^*$ & RGB & 7.68 & 77.36 & 76.03 & 1.33 [0.61, 2.68] & 98.1 \\
CART~\cite{cart} & Th--I & RGB & 4.08 & 21.99 & 16.84 & 5.15 [1.94, 6.94] & 71.2 \\
NYUv2~\cite{nyuv2} & Tx--D$^*$ & RGB & 57.87 & 64.57 & 64.49 & 0.08 [\textminus 0.15, 0.45] & 98.8 \\
\addlinespace[2pt]
\multicolumn{8}{l}{\textbf{LanguageBind}~\cite{languagebind}} \\
BatVision~\cite{batvision} & D--Au & RGB & 2.65 & 13.35 & 13.12 & 0.23 [\textminus 0.30, 1.24] & 97.8 \\
VGGSound~\cite{vggsound} & Im--Au & Text & 43.60 & 56.91 & 55.19 & 1.72 [0.82, 2.85] & 87.0 \\
NYUv2~\cite{nyuv2} & Im--D & Text & 50.69 & 98.71 & 98.59 & 0.12 [0.00, 0.19] & 99.7 \\
TartanRGBT~\cite{anythermal} & Im--Th & Depth & 36.93 & 99.35 & 99.11 & 0.24 [0.00, 0.52] & 99.6 \\
UCF101~\cite{ucf101} & Im--V & Text & 96.56 & 99.76 & 99.65 & 0.11 [0.07, 0.14] & 96.7 \\
MAVD~\cite{mavd} & Th--Au & RGB & 2.02 & 5.00 & 4.83 & 0.17 [\textminus 0.31, 1.27] & 94.3 \\
TartanRGBT~\cite{anythermal} & Th--D & RGB & 25.23 & 98.50 & 98.41 & 0.08 [\textminus 0.17, 0.23] & 99.9 \\
MSR-VTT~\cite{msrvtt} & V--Au & Text & 6.96 & 30.39 & 26.76 & 3.63 [3.11, 4.61] & 84.5 \\
TartanRGBT~\cite{anythermal} & V--D & RGB & 25.00 & 88.02 & 89.30 & \textminus 1.28 [\textminus 1.78, \textminus 0.69] & 102.0 \\
TartanRGBT~\cite{anythermal} & V--Th & RGB & 37.16 & 89.35 & 89.81 & \textminus 0.46 [\textminus 0.91, 0.34] & 100.9 \\
\addlinespace[2pt]
\midrule
\fullrowband{resultshade}
\multicolumn{3}{l}{\textbf{Mean of the 20 Figure Configurations}} & 24.13 & 53.21 & 52.21 & 1.00 & 96.6 \\
\bottomrule
\end{tabular*}
\end{table*}
Table~\ref{tab:supp-pivot} fixes endpoint features and replaces the pivot with part means, independent within-sample part permutations, or state indicators.
The mean condition replaces every pivot part with the sample's mean vector, retaining its part count.
State indicators provide one vector per sample.
Permutations retain each sample's set of vectors while changing their part indices.
All thirteen settings have multi-part pivots; LanguageBind uses RGB here.
State indicators encode TartanRGBT trajectories, UTD-MHAD actions, BatVision recording locations, MAVD vehicle counts, Ego4D scenarios, and the 24 NYUv2 scene types.
For Ego4D, single-tag scenarios with at least five training videos give 26 audio--IMU and 35 text--IMU states; other samples form an additional state.
MAVD uses Grounding DINO~\cite{groundingdino} base pseudo-labels for 1, 2, 3, or at least 4 cars, with 29 unlabeled samples in an additional state.
CT has no state-label condition.
Category-level text endpoints are excluded because their class names reveal the retrieval target under the state-label condition.

The five LB rows replace their main-table text prompts with RGB pivots, changing both pivot features and direction cap; a single-vector text pivot is capped by its number of states minus one (Sec.~\ref{sec:estimator-details}), giving 36 directions for LB TR Th--D versus 256 with RGB.
Table~\ref{tab:app-lbtr-config} lists the LB TR Th--D configurations in Tables~2 and~4(c) of the main text.

\subsection{Shared and Separate Fitting Collections}
\label{sec:collection-overlap-control}
For each seed from 42 through 51, the fitting samples are randomly divided into equal halves $I_1,I_2$, dropping one sample if necessary.
The \emph{shared collection} condition fits both edges from $I_1$ or both from $I_2$ and averages the two results.
The \emph{separate collections} condition fits $A$--$P$ from $I_1$ and $P$--$B$ from $I_2$, then reverses the assignments and averages the results.
Per-edge sample counts are identical across conditions.
Reference banks always draw $A$ samples from $I_1$ and $B$ samples from $I_2$, using the same seed and cap.
The repeat index $w$ is computed on $I_1\cup I_2$ and held fixed within each split.
Every condition uses the retrieval score $R$ on the same held-out samples.

Table~\ref{tab:app-collection-overlap} specifies the datasets and pivots for all 20 configurations in Figure~7 of the main text.
ImageBind VGGSound text--audio, TartanRGBT text--thermal, and NYUv2 text--depth use category-consistent positives in this experiment.
NYUv2 uses scene prompts here; caption-instance retrieval follows Sec.~\ref{sec:full-protocols}.
In Figure~7 of the main text, six LanguageBind settings with a frame-level image alongside the two endpoints use that image as the pivot: BatVision depth--audio, MAVD thermal--audio, and TartanRGBT thermal--depth, video--depth, and video--thermal use RGB; TartanRGBT image--thermal uses depth.
Their main-table pivot is a text prompt shared by all frames of a recording, so a shared collection would link $A$ and $B$ through the same frames; a frame-level image instead varies across individual frames of the recording.
The random halves split the same recordings, so text prompts recur in both halves; frame-level images rarely repeat.
All remaining pivots follow the main-table configurations.
All main-table settings fit the two modality--pivot edges on index-disjoint samples (Sec.~\ref{sec:full-protocols}), giving the separate-collection construction with the main-table pivots.

\subsection{Pairing of Corresponding Parts}
\label{sec:app-part-pairing}
\begin{figure*}[t]
\centering
\includegraphics[width=\textwidth]{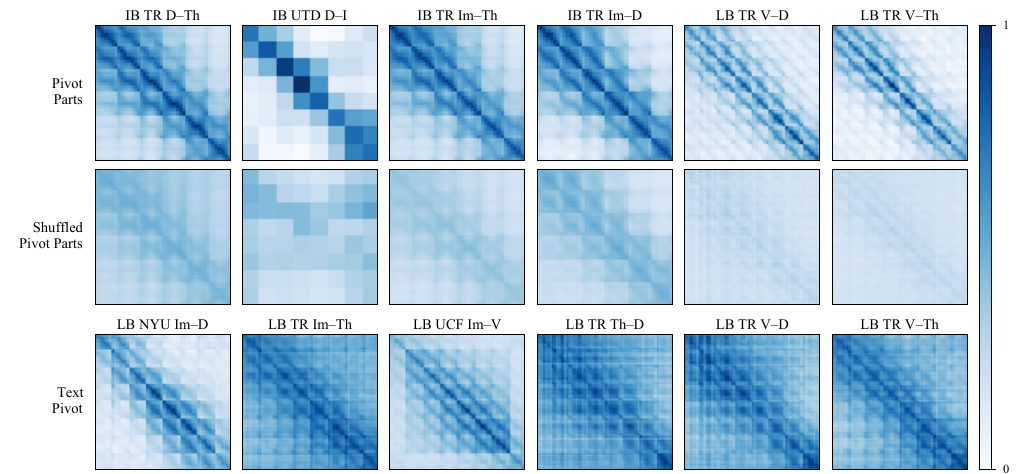}
\caption{\textbf{Pairing of Corresponding Parts.} Rows index parts of $A$ and columns parts of $B$; darker cells pair more strongly, with negative values shown as 0 and each top or bottom panel scaled to its maximum. Top: multi-part pivots; middle: the same settings with pivot parts shuffled before fitting, on the same scale; bottom: LanguageBind main-table settings with text pivots.}
\label{fig:supp-pairing}
\end{figure*}
\begin{table*}[t]
\centering
\caption{\textbf{Pairing Statistics.} Top: share of $A$ parts whose strongest partner is the corresponding $B$ part. Strength: mean corresponding and non-corresponding pairing (nats). R@10: mean single-part bidirectional R@10 (\%). Shaded: Top. --: single-vector pivot.}
\label{tab:supp-pairing}
\fontsize{8.0}{9.6}\selectfont
\setlength{\tabcolsep}{1.4pt}
\renewcommand{\arraystretch}{1.1}
\begin{tabular*}{\textwidth}{@{\extracolsep{\fill}}llcccccccc@{}}
\toprule
 & & & & \multicolumn{4}{c}{Pivot as Given} & \multicolumn{2}{c}{Pivot Parts Shuffled} \\
\cmidrule(lr){5-8}\cmidrule(lr){9-10}
Dataset & Pair & Pivot & Parts & Top (\%) & Strength & R@10 & Chance & Strength & R@10 \\
\midrule
\multicolumn{10}{l}{\textbf{ImageBind}~\cite{imagebind}} \\
TartanRGBT~\cite{anythermal} & D--Th & RGB & 42 & \cellcolor{resultshade}100 & 9.28 / 4.57 & 15.6 & 0.12 & 4.84 / 3.68 & 8.1 \\
UTD-MHAD~\cite{utdmhad} & D--I & Video & 8 & \cellcolor{resultshade}100 & 1.04 / 0.33 & 11.8 & 2.33 & 0.49 / 0.40 & 7.2 \\
TartanRGBT~\cite{anythermal} & Im--Th & Depth & 42 & \cellcolor{resultshade}100 & 9.63 / 4.76 & 24.1 & 0.12 & 4.32 / 3.16 & 9.7 \\
TartanRGBT~\cite{anythermal} & Im--D & Thermal & 42 & \cellcolor{resultshade}100 & 9.02 / 4.19 & 28.1 & 0.12 & 4.80 / 3.27 & 11.8 \\
\addlinespace[2pt]
\multicolumn{10}{l}{\textbf{LanguageBind}~\cite{languagebind}} \\
TartanRGBT~\cite{anythermal} & V--D & RGB & 64 & \cellcolor{resultshade}98 & 11.03 / 3.87 & 48.1 & 2.29 & 4.16 / 3.12 & 22.4 \\
TartanRGBT~\cite{anythermal} & V--Th & RGB & 64 & \cellcolor{resultshade}100 & 12.39 / 3.83 & 53.7 & 2.29 & 4.29 / 3.23 & 25.1 \\
NYUv2~\cite{nyuv2} & Im--D & Text & 64 & \cellcolor{resultshade}92 & 1.08 / 0.45 & 21.7 & 1.53 & -- & -- \\
TartanRGBT~\cite{anythermal} & Im--Th & Text & 64 & \cellcolor{resultshade}80 & 8.52 / 5.83 & 25.5 & 2.29 & -- & -- \\
UCF101~\cite{ucf101} & Im--V & Text & 64 & \cellcolor{resultshade}100 & 10.06 / 5.01 & 87.3 & 0.35 & -- & -- \\
TartanRGBT~\cite{anythermal} & Th--D & Text & 64 & \cellcolor{resultshade}59 & 7.70 / 5.43 & 16.9 & 2.29 & -- & -- \\
TartanRGBT~\cite{anythermal} & V--D & Text & 64 & \cellcolor{resultshade}61 & 11.09 / 7.38 & 21.3 & 2.29 & -- & -- \\
TartanRGBT~\cite{anythermal} & V--Th & Text & 64 & \cellcolor{resultshade}77 & 11.45 / 7.49 & 26.1 & 2.29 & -- & -- \\
\bottomrule
\end{tabular*}
\end{table*}
\begin{table*}[t]
\centering
\caption{\textbf{Existing Methods on PRAM's Local Vectors.} R@10 (\%) for retrieval; top-1 accuracy (\%) for classification; all columns read PRAM's local vectors. LT: Latent Translation; LI: late interaction; Heads: projection heads. Bold: best in each row; $^*$: class names on the text side, as in Tables~1 and~2 of the main text.}
\label{tab:app-local-methods}
\fontsize{6.5}{8.1}\selectfont
\setlength{\tabcolsep}{2pt}
\renewcommand{\arraystretch}{1.0}
\begin{tabularx}{\textwidth}{@{}>{\raggedright\arraybackslash}p{.21\textwidth}>{\centering\arraybackslash}p{.08\textwidth}>{\centering\arraybackslash}p{.075\textwidth}Y>{\centering\arraybackslash}p{.085\textwidth}>{\centering\arraybackslash}p{.125\textwidth}>{\centering\arraybackslash}p{.125\textwidth}Y@{}}
\toprule
Relation & PRAM & Heads & Partwise ASIF~\cite{asif} & LT~\cite{latenttranslation} & LT+LI~\cite{latenttranslation,colbert} & STRUCTURE~\cite{structure} & Partwise MMBind~\cite{mmbind} \\
\midrule
\multicolumn{8}{l}{\textbf{Retrieval R@10}} \\
IB BV~\cite{batvision} Au--D & \cellcolor{resultshade}\textbf{11.1} & 9.0 & 8.3 & 7.3 & 7.1 & 6.0 & 9.5 \\
IB E4D~\cite{ego4d} Au--I & \cellcolor{resultshade}\textbf{10.9} & 8.3 & 5.5 & 3.9 & 4.2 & 7.3 & 8.7 \\
IB MAV~\cite{mavd} Au--Th & \cellcolor{resultshade}5.1 & \textbf{5.9} & 3.9 & 4.6 & 4.5 & 4.2 & 4.2 \\
IB UTD~\cite{utdmhad} D--I & \cellcolor{resultshade}\textbf{22.7} & 16.4 & 7.2 & 6.3 & 4.8 & 17.9 & 16.6 \\
IB TR~\cite{anythermal} D--Th & \cellcolor{resultshade}\textbf{75.1} & 66.4 & 67.5 & 73.2 & 69.3 & 69.0 & 68.5 \\
IB VGG$^*$~\cite{vggsound} Tx--Au & \cellcolor{resultshade}\textbf{82.1} & 77.8 & 71.1 & 76.1 & 77.5 & 78.4 & 76.0 \\
IB E4D~\cite{ego4d} Tx--I & \cellcolor{resultshade}\textbf{6.9} & 4.3 & 4.1 & 3.8 & 4.0 & 5.4 & 5.9 \\
IB TR$^*$~\cite{anythermal} Tx--Th & \cellcolor{resultshade}\textbf{76.3} & 71.6 & 67.1 & 73.3 & 73.5 & 64.3 & 73.8 \\
IB CT~\cite{cart} Th--I & \cellcolor{resultshade}19.8 & 5.9 & \textbf{21.4} & 11.0 & 11.0 & 4.1 & 4.2 \\
IB NYU~\cite{nyuv2} Tx--D & \cellcolor{resultshade}\textbf{18.7} & 8.4 & 9.3 & 8.7 & 8.8 & 10.1 & 12.9 \\
\textit{IB Mean} & \cellcolor{resultshade}\textbf{32.9} & 27.4 & 26.5 & 26.8 & 26.5 & 26.7 & 28.0 \\
LB BV~\cite{batvision} D--Au & \cellcolor{resultshade}6.3 & 4.7 & 6.9 & 6.6 & \textbf{7.5} & 4.5 & \textbf{7.5} \\
LB VGG~\cite{vggsound} Im--Au & \cellcolor{resultshade}\textbf{53.9} & 26.3 & 22.8 & 17.6 & 18.5 & 25.7 & 28.5 \\
LB NYU~\cite{nyuv2} Im--D & \cellcolor{resultshade}\textbf{98.5} & 35.6 & 48.1 & 37.3 & 25.7 & 40.1 & 22.9 \\
LB TR~\cite{anythermal} Im--Th & \cellcolor{resultshade}\textbf{76.1} & 39.3 & 43.0 & 41.6 & 34.6 & 46.9 & 32.8 \\
LB UCF~\cite{ucf101} Im--V & \cellcolor{resultshade}\textbf{99.6} & 99.5 & 91.8 & 88.3 & 59.3 & 99.5 & 95.7 \\
LB MAV~\cite{mavd} Th--Au & \cellcolor{resultshade}\textbf{5.3} & 4.0 & 4.0 & 4.2 & 4.4 & 4.2 & 4.6 \\
LB TR~\cite{anythermal} Th--D & \cellcolor{resultshade}\textbf{54.6} & 20.8 & 22.0 & 21.0 & 20.1 & 26.7 & 20.1 \\
LB MSR~\cite{msrvtt} V--Au & \cellcolor{resultshade}\textbf{26.4} & 17.0 & 14.3 & 14.3 & 13.7 & 18.9 & 14.0 \\
LB TR~\cite{anythermal} V--D & \cellcolor{resultshade}\textbf{51.8} & 26.0 & 25.1 & 18.2 & 19.3 & 34.1 & 26.3 \\
LB TR~\cite{anythermal} V--Th & \cellcolor{resultshade}\textbf{61.5} & 33.0 & 35.3 & 27.6 & 27.0 & 40.7 & 22.6 \\
\textit{LB Mean} & \cellcolor{resultshade}\textbf{53.4} & 30.6 & 31.3 & 27.7 & 23.0 & 34.1 & 27.5 \\
\addlinespace[2pt]
\multicolumn{8}{l}{\textbf{Classification Top-1}} \\
IB BV~\cite{batvision} Au--D & \cellcolor{resultshade}74.4 & 73.2 & 68.9 & 73.8 & 73.7 & 65.7 & \textbf{77.1} \\
IB E4D~\cite{ego4d} Au--I & \cellcolor{resultshade}\textbf{30.6} & 29.6 & 23.3 & 15.3 & 16.7 & 29.9 & 27.5 \\
IB UTD~\cite{utdmhad} D--I & \cellcolor{resultshade}\textbf{50.7} & 27.3 & 28.1 & 18.4 & 13.4 & 31.1 & 35.6 \\
IB TR~\cite{anythermal} D--Th & \cellcolor{resultshade}\textbf{51.9} & 46.0 & 51.1 & 50.7 & 51.0 & 47.2 & 45.9 \\
IB VGG$^*$~\cite{vggsound} Tx--Au & \cellcolor{resultshade}\textbf{64.7} & 55.6 & 47.2 & 55.2 & 58.0 & 57.8 & 53.0 \\
IB E4D~\cite{ego4d} Tx--I & \cellcolor{resultshade}26.4 & 23.1 & 18.5 & 21.2 & 21.1 & 22.9 & \textbf{29.1} \\
IB TR$^*$~\cite{anythermal} Tx--Th & \cellcolor{resultshade}\textbf{65.7} & 45.5 & 37.2 & 47.5 & 47.9 & 37.5 & 49.8 \\
IB NYU$^*$~\cite{nyuv2} Tx--D & \cellcolor{resultshade}\textbf{40.4} & 30.7 & 10.6 & 30.3 & 32.0 & 14.1 & \textbf{40.4} \\
\textit{IB Mean} & \cellcolor{resultshade}\textbf{50.6} & 41.4 & 35.6 & 39.0 & 39.2 & 38.3 & 44.8 \\
LB BV~\cite{batvision} D--Au & \cellcolor{resultshade}\textbf{75.8} & 66.4 & 68.0 & 71.6 & 70.9 & 60.4 & 73.9 \\
LB VGG~\cite{vggsound} Im--Au & \cellcolor{resultshade}\textbf{68.8} & 51.6 & 46.6 & 46.3 & 46.2 & 48.9 & 49.9 \\
LB NYU~\cite{nyuv2} Im--D & \cellcolor{resultshade}\textbf{72.2} & 49.0 & 59.6 & 60.7 & 60.0 & 50.8 & 59.3 \\
LB TR~\cite{anythermal} Im--Th & \cellcolor{resultshade}\textbf{93.4} & 91.6 & 92.5 & 92.0 & 92.1 & 92.4 & 91.2 \\
LB UCF~\cite{ucf101} Im--V & \cellcolor{resultshade}\textbf{91.6} & 67.6 & 65.0 & 67.4 & 67.8 & 66.5 & 65.8 \\
LB TR~\cite{anythermal} Th--D & \cellcolor{resultshade}91.7 & 92.1 & 91.7 & 92.0 & \textbf{92.4} & 91.9 & 91.7 \\
LB MSR~\cite{msrvtt} V--Au & \cellcolor{resultshade}\textbf{42.6} & 40.0 & 40.4 & 32.9 & 34.5 & 40.7 & 38.0 \\
LB TR~\cite{anythermal} V--D & \cellcolor{resultshade}92.4 & 92.2 & 91.9 & 92.2 & 92.2 & \textbf{93.5} & 93.0 \\
LB TR~\cite{anythermal} V--Th & \cellcolor{resultshade}\textbf{92.9} & 91.5 & 92.7 & 91.7 & 91.7 & 92.4 & 91.4 \\
\textit{LB Mean} & \cellcolor{resultshade}\textbf{80.2} & 71.3 & 72.1 & 71.9 & 72.0 & 70.8 & 72.7 \\
\bottomrule
\end{tabularx}
\end{table*}
This check examines whether the part relations fitted from modality--pivot pairs match each part of one endpoint with the corresponding part of the other.
For the relation of part $g$, the pairing strength of $A$ part $g$ and $B$ part $g'$ is
\begin{equation}
M_{gg'}=\mathbb{E}_{i}\,\ell_g(a_{ig},b_{ig'})-\mathbb{E}_{i\neq j}\,\ell_g(a_{ig},b_{jg'}),
\end{equation}
where $B$ part $g'$ is projected by the part-$g$ relation and the expectations run over held-out pairs.
The second term uses other samples at the same position, so similarity explained by position alone cancels; the quadratic terms of $\ell_g$ cancel as well, and $M_{gg'}$ is in nats.
Single-part retrieval ranks all held-out samples at one position by $\ell_g$ alone.
We report the main-table configurations whose two endpoints are both divided into corresponding parts and whose cross-part intensity $\alpha$ is below one, together with four TartanRGBT settings with multi-part pivots.
For multi-part pivots, a control shuffles pivot parts within each sample before fitting, keeping features, partition, and estimator fixed.

Figure~\ref{fig:supp-pairing} and Table~\ref{tab:supp-pairing} show the corresponding part as the strongest partner for 59--100\% of parts in every setting.
Shuffling pivot parts reduces the corresponding-part strength to 35--53\% and the single-part R@10 to 40--61\% of their values.
A single ImageBind TartanRGBT cell retrieves its depth--thermal counterpart among 8,601 samples with 15.6\% R@10, against 0.12\% by chance; a single LanguageBind UCF101 cell reaches 87.3\% among 2,874 samples.
The fitted part relations therefore pair corresponding parts, and with multi-part pivots this pairing depends on part correspondence in the pivot.

\subsection{Misregistered Fitting Data}
We shift only the depth tokens of the depth--RGB modality--pivot fitting edge of IB TR D--Th by $(\Delta y,\Delta x)$ tokens on the registered $12\times14$ grid with replicated boundary padding, then apply the main-table $2\times2$ averaging to obtain $6\times7$ parts.
One token is half a part cell; test samples stay registered.
Latent Translation (LT), LT with late interaction (LT+LI), and Partwise MMBind use PRAM's local vectors (Table~\ref{tab:app-local-methods}); all four are refitted per shift.
Table~\ref{tab:app-misregistration} shows that R@10 decreases for all four methods as shift magnitude increases along each tested direction, with larger declines under vertical shifts than horizontal shifts of the same size.
PRAM achieves the highest R@10 at all ten offsets and retains 92.1\%, 74.2\%, and 62.4\% of its unshifted R@10 at $(0,1)$, $(1,0)$, and $(0,3)$, compared with maxima of 86.6\% and 62.8\% at $(0,1)$ and $(1,0)$ and 43.8\% at $(0,3)$ among LT, LT+LI, and Partwise MMBind.
All four methods show lower R@10 under diagonal shifts than under either corresponding single-axis shift, with PRAM remaining highest at every diagonal offset.
At $(1,1)$, PRAM retains 63.8\% of its unshifted R@10 versus LT's 53.8\%, the highest among the other methods.
PRAM's R@10 lead grows from 1.9 points at $(0,0)$ to 16.9 at $(0,3)$.
The registration estimator (Sec.~\ref{sec:registration-details}) is not re-run; all nine nonzero shifts lie within its three-token per-axis radius.
\begin{table*}[t]
\centering
\caption{\textbf{Misregistered Fitting Data.} Bidirectional R@10 (\%) on IB TR D--Th after shifting the depth tokens of the fitting depth--RGB pairs by $(\Delta y,\Delta x)$ tokens before pooling; one token is half a part cell, and test samples stay registered. All methods read PRAM's local vectors, as in Table~\ref{tab:app-local-methods}; bold marks the best result in each column.}
\label{tab:app-misregistration}
\fontsize{8.0}{9.6}\selectfont
\renewcommand{\arraystretch}{1.0}
\setlength{\tabcolsep}{3pt}
\begin{tabular*}{\textwidth}{@{\extracolsep{\fill}}l*{10}{c}@{}}
\toprule
& \multicolumn{10}{c}{Shift $(\Delta y,\Delta x)$ in Tokens} \\
\cmidrule(l){2-11}
Method & $(0,0)$ & $(0,1)$ & $(1,0)$ & $(1,1)$ & $(0,2)$ & $(2,0)$ & $(2,2)$ & $(0,3)$ & $(3,0)$ & $(3,3)$ \\
\midrule
\fullrowband{resultshade}
\textbf{PRAM (Ours)} & \textbf{75.1} & \textbf{69.2} & \textbf{55.7} & \textbf{47.9} & \textbf{57.8} & \textbf{24.1} & \textbf{14.7} & \textbf{46.9} & \textbf{11.7} & \textbf{5.4} \\
LT~\cite{latenttranslation} & 73.2 & 63.4 & 46.0 & 39.4 & 45.3 & 15.4 & 9.1 & 29.1 & 6.9 & 3.0 \\
LT+LI~\cite{latenttranslation,colbert} & 69.3 & 57.7 & 39.3 & 33.7 & 41.9 & 15.5 & 9.7 & 27.9 & 9.4 & 3.6 \\
Partwise MMBind~\cite{mmbind} & 68.5 & 57.1 & 39.2 & 31.0 & 38.2 & 11.4 & 7.2 & 30.0 & 7.2 & 3.8 \\
\bottomrule
\end{tabular*}
\end{table*}
\begin{table*}[t]
\centering
\caption{\textbf{Modality--Pivot Pair Budgets.} Bidirectional R@10 (\%). Each row lists the available pair count $n$; budget $q$ uses $\max\{50,\operatorname{round}(qn)\}$ training samples. Reduced budgets give the mean and [minimum, maximum] over seeds 0 and 1; 100\% uses one fit. Test samples are fixed. Shaded: the 2\% budget.}
\label{tab:app-budget-scores}
\fontsize{8.0}{9.6}\selectfont
\renewcommand{\arraystretch}{1.02}
\setlength{\tabcolsep}{2.5pt}
\resizebox{\textwidth}{!}{\begin{tabular}{@{}lr*{8}{c}@{}}
\toprule
Configuration & $n$ & 2\% & 5\% & 10\% & 20\% & 35\% & 50\% & 70\% & 100\% \\
\midrule
IB TR~\cite{anythermal} D--Th & $8{,}602$ & \cellcolor{resultshade}59.0\,{\tiny [58.8, 59.3]} & 63.7\,{\tiny [63.6, 63.9]} & 69.2\,{\tiny [68.3, 70.0]} & 72.8\,{\tiny [72.3, 73.3]} & 74.8\,{\tiny [74.7, 75.0]} & 75.1\,{\tiny [75.1, 75.2]} & 75.0\,{\tiny [75.0, 75.1]} & 75.1 \\
IB TR~\cite{anythermal} D--Th (Unseen) & $13{,}744$ & \cellcolor{resultshade}46.4\,{\tiny [45.7, 47.0]} & 51.7\,{\tiny [51.5, 52.0]} & 57.7\,{\tiny [57.1, 58.2]} & 60.4\,{\tiny [60.2, 60.5]} & 61.9\,{\tiny [61.8, 62.1]} & 62.5\,{\tiny [62.2, 62.8]} & 62.4\,{\tiny [62.1, 62.6]} & 62.7 \\
LB TR~\cite{anythermal} Th--D (RGB) & $1{,}733$ & \cellcolor{resultshade}94.0\,{\tiny [93.9, 94.2]} & 95.6\,{\tiny [95.5, 95.6]} & 96.6\,{\tiny [96.1, 97.1]} & 98.5\,{\tiny [97.9, 99.0]} & 98.3\,{\tiny [98.2, 98.5]} & 98.2\,{\tiny [98.0, 98.3]} & 98.5\,{\tiny [98.4, 98.5]} & 98.5 \\
IB MAV~\cite{mavd} D--Th & $647$ & \cellcolor{resultshade}11.7\,{\tiny [10.5, 12.8]} & 11.7\,{\tiny [10.5, 12.8]} & 13.6\,{\tiny [11.5, 15.7]} & 16.1\,{\tiny [15.3, 17.0]} & 22.3\,{\tiny [21.9, 22.8]} & 21.5\,{\tiny [21.2, 21.9]} & 24.2\,{\tiny [24.1, 24.3]} & 24.7 \\
IB UTD~\cite{utdmhad} D--I & $431$ & \cellcolor{resultshade}4.0\,{\tiny [3.5, 4.5]} & 4.0\,{\tiny [3.5, 4.5]} & 4.0\,{\tiny [3.5, 4.5]} & 5.6\,{\tiny [5.3, 5.9]} & 15.2\,{\tiny [14.3, 16.2]} & 16.0\,{\tiny [13.4, 18.6]} & 20.6\,{\tiny [19.4, 21.9]} & 22.7 \\
LB VGG~\cite{vggsound} Im--Au & $1{,}500$ & \cellcolor{resultshade}3.0\,{\tiny [2.3, 3.8]} & 6.0\,{\tiny [4.5, 7.5]} & 11.9\,{\tiny [10.5, 13.2]} & 25.8\,{\tiny [25.5, 26.1]} & 39.8\,{\tiny [38.1, 41.5]} & 45.9\,{\tiny [44.5, 47.2]} & 51.1\,{\tiny [48.9, 53.4]} & 53.9 \\
LB UCF~\cite{ucf101} Im--V & $7{,}226$ & \cellcolor{resultshade}67.3\,{\tiny [59.4, 75.2]} & 97.9\,{\tiny [97.6, 98.2]} & 99.2\,{\tiny [99.2, 99.2]} & 99.5\,{\tiny [99.5, 99.5]} & 99.6\,{\tiny [99.6, 99.6]} & 99.6\,{\tiny [99.6, 99.6]} & 99.6\,{\tiny [99.6, 99.6]} & 99.6 \\
IB BV~\cite{batvision} Au--D & $2{,}536$ & \cellcolor{resultshade}2.0\,{\tiny [1.8, 2.1]} & 5.0\,{\tiny [4.6, 5.4]} & 6.5\,{\tiny [5.3, 7.7]} & 8.6\,{\tiny [6.8, 10.3]} & 9.0\,{\tiny [8.3, 9.8]} & 9.1\,{\tiny [7.7, 10.4]} & 10.9\,{\tiny [10.6, 11.2]} & 11.1 \\
IB E4D~\cite{ego4d} Au--I & $1{,}138$ & \cellcolor{resultshade}2.5\,{\tiny [1.9, 3.1]} & 2.8\,{\tiny [2.4, 3.2]} & 1.9\,{\tiny [1.9, 1.9]} & 4.6\,{\tiny [4.1, 5.0]} & 6.7\,{\tiny [6.5, 6.9]} & 7.3\,{\tiny [6.9, 7.7]} & 8.5\,{\tiny [6.9, 10.1]} & 10.9 \\
LB MSR~\cite{msrvtt} V--Au & $6{,}176$ & \cellcolor{resultshade}2.5\,{\tiny [2.5, 2.5]} & 4.6\,{\tiny [4.0, 5.3]} & 8.0\,{\tiny [7.5, 8.6]} & 13.6\,{\tiny [13.3, 13.9]} & 17.0\,{\tiny [16.5, 17.5]} & 21.2\,{\tiny [21.1, 21.2]} & 22.8\,{\tiny [21.6, 24.1]} & 26.4 \\
\bottomrule
\end{tabular}}
\end{table*}
\begin{table}[t]
\centering
\caption{\textbf{Cross-Part Shrinkage under Pair Budgets.} Bidirectional R@10 (\%) of PRAM with and without cross-part shrinkage ($\alpha=0$); 2\% gives the mean of the two subsets; shaded: PRAM.}
\label{tab:app-budget-shrinkage}
\fontsize{8.0}{9.6}\selectfont
\setlength{\tabcolsep}{1pt}
\begin{tabularx}{\linewidth}{@{}>{\raggedright\arraybackslash}p{.46\linewidth}*{4}{Y}@{}}
\toprule
 & \multicolumn{2}{c}{2\%} & \multicolumn{2}{c}{100\%} \\
\cmidrule(lr){2-3}\cmidrule(l){4-5}
Configuration & Full & $\alpha=0$ & Full & $\alpha=0$ \\
\midrule
IB TR~\cite{anythermal} D--Th & \cellcolor{resultshade}59.0 & 21.7 & \cellcolor{resultshade}75.1 & 75.2 \\
IB TR~\cite{anythermal} D--Th (Unseen) & \cellcolor{resultshade}46.4 & 25.0 & \cellcolor{resultshade}62.7 & 62.8 \\
LB TR~\cite{anythermal} Th--D (RGB) & \cellcolor{resultshade}94.0 & 41.6 & \cellcolor{resultshade}98.5 & 98.2 \\
IB MAV~\cite{mavd} D--Th & \cellcolor{resultshade}11.7 & 7.6 & \cellcolor{resultshade}24.7 & 20.6 \\
IB UTD~\cite{utdmhad} D--I & \cellcolor{resultshade}4.0 & 2.8 & \cellcolor{resultshade}22.7 & 19.7 \\
LB VGG~\cite{vggsound} Im--Au & \cellcolor{resultshade}3.0 & 2.8 & \cellcolor{resultshade}53.9 & 53.1 \\
LB UCF~\cite{ucf101} Im--V & \cellcolor{resultshade}67.3 & 75.2 & \cellcolor{resultshade}99.6 & 99.6 \\
IB BV~\cite{batvision} Au--D & \cellcolor{resultshade}2.0 & 2.8 & \cellcolor{resultshade}11.1 & 10.4 \\
IB E4D~\cite{ego4d} Au--I & \cellcolor{resultshade}2.5 & 1.9 & \cellcolor{resultshade}10.9 & 9.9 \\
LB MSR~\cite{msrvtt} V--Au & \cellcolor{resultshade}2.5 & 2.4 & \cellcolor{resultshade}26.4 & 26.0 \\
\bottomrule
\end{tabularx}
\end{table}
\begin{table}[t]
\centering
\caption{\textbf{Ten Subsets at the 2\% Budget.} Mean drop in PRAM's bidirectional R@10 (percentage points) without cross-part shrinkage ($\alpha=0$), with a 95\% $t$-interval over ten shared subsets and the number of subsets showing a drop.}
\label{tab:app-budget-ten}
\fontsize{7.2}{8.8}\selectfont
\setlength{\tabcolsep}{1pt}
\begin{tabularx}{\linewidth}{@{}>{\raggedright\arraybackslash}p{.41\linewidth}>{\centering\arraybackslash}p{.29\linewidth}Y@{}}
\toprule
Configuration & Drop [95\% Interval] & Subsets with a Drop \\
\midrule
IB TR~\cite{anythermal} D--Th & 39.5 [37.8, 41.2] & 10 \\
IB TR~\cite{anythermal} D--Th (Unseen) & 22.9 [21.9, 23.8] & 10 \\
LB TR~\cite{anythermal} Th--D (RGB) & 49.8 [47.1, 52.5] & 10 \\
IB MAV~\cite{mavd} D--Th & 5.0 [3.6, 6.4] & 10 \\
IB UTD~\cite{utdmhad} D--I & 1.7 [0.5, 2.8] & 8 \\
LB VGG~\cite{vggsound} Im--Au & 0.2 [0.0, 0.4] & 7 \\
LB UCF~\cite{ucf101} Im--V & \textminus 4.3 [\textminus 7.4, \textminus 1.3] & 2 \\
IB BV~\cite{batvision} Au--D & 0.5 [\textminus 0.4, 1.3] & 6 \\
IB E4D~\cite{ego4d} Au--I & \textminus 0.1 [\textminus 1.0, 0.8] & 6 \\
LB MSR~\cite{msrvtt} V--Au & \textminus 0.1 [\textminus 0.5, 0.4] & 5 \\
\bottomrule
\end{tabularx}
\end{table}
\begin{table*}[t]
\centering
\caption{\textbf{Full ImageBind Retrieval.} Bidirectional R@1 and R@10 (\%) for all 15 sample-level baselines. The gray row marks Frozen Cosine as the default reference and is excluded from bold ranking; bold marks the best pivot-only result. $^*$: class names on the text side.}
\label{tab:supp-ret-ib}
\fontsize{8.0}{9.6}\selectfont
\setlength{\tabcolsep}{2.0pt}
\renewcommand{\arraystretch}{1.12}
\resizebox{\textwidth}{!}{\begin{tabular}{@{}l*{20}{c}@{}}
\toprule
Method & \multicolumn{2}{c}{\shortstack{BV~\cite{batvision}\\Au--D}} & \multicolumn{2}{c}{\shortstack{E4D~\cite{ego4d}\\Au--I}} & \multicolumn{2}{c}{\shortstack{MAV~\cite{mavd}\\Au--Th}} & \multicolumn{2}{c}{\shortstack{UTD~\cite{utdmhad}\\D--I}} & \multicolumn{2}{c}{\shortstack{TR~\cite{anythermal}\\D--Th}} & \multicolumn{2}{c}{\shortstack{VGG$^*$~\cite{vggsound}\\Tx--Au}} & \multicolumn{2}{c}{\shortstack{E4D~\cite{ego4d}\\Tx--I}} & \multicolumn{2}{c}{\shortstack{TR$^*$~\cite{anythermal}\\Tx--Th}} & \multicolumn{2}{c}{\shortstack{CT~\cite{cart}\\Th--I}} & \multicolumn{2}{c}{\shortstack{NYU~\cite{nyuv2}\\Tx--D}} \\
\cmidrule(lr){2-3}\cmidrule(lr){4-5}\cmidrule(lr){6-7}\cmidrule(lr){8-9}\cmidrule(lr){10-11}\cmidrule(lr){12-13}\cmidrule(lr){14-15}\cmidrule(lr){16-17}\cmidrule(lr){18-19}\cmidrule(lr){20-21}
 & R@1 & R@10 & R@1 & R@10 & R@1 & R@10 & R@1 & R@10 & R@1 & R@10 & R@1 & R@10 & R@1 & R@10 & R@1 & R@10 & R@1 & R@10 & R@1 & R@10 \\
\midrule
\fullrowband{gray!13}
Frozen Cosine~\cite{imagebind} & 0.2 & 2.0 & 0.1 & 1.7 & 0.1 & 1.6 & 0.3 & 3.1 & 0.0 & 0.3 & 65.0 & 76.0 & 0.3 & 1.5 & 2.1 & 7.7 & 0.4 & 4.1 & 1.7 & 14.4 \\
\midrule
Mean Centering & 0.3 & 2.7 & 0.2 & 2.3 & 0.0 & 1.8 & 0.0 & 2.4 & 0.1 & 0.4 & 67.8 & 76.8 & 0.1 & 2.0 & 3.4 & 12.8 & 0.0 & 2.7 & 2.3 & 14.8 \\
Projection Heads & 0.9 & 8.3 & 0.5 & 6.9 & 0.1 & 1.8 & 0.5 & 4.8 & 2.0 & 12.9 & 63.4 & 76.4 & 0.5 & 4.7 & 54.7 & 72.4 & 0.6 & 6.1 & 2.8 & 17.1 \\
\midrule
I0T~\cite{i0t} & 0.1 & 2.4 & 0.1 & 2.3 & 0.1 & 1.7 & 0.6 & 2.8 & 0.0 & 0.4 & 66.0 & 76.5 & 0.1 & 1.7 & 3.2 & 9.9 & 0.2 & 2.2 & 2.1 & 14.8 \\
DBSN~\cite{dbsn} & 0.2 & 2.3 & 0.5 & 3.1 & 0.1 & 2.1 & 0.1 & 2.4 & 0.0 & 0.4 & 73.5 & 78.5 & 0.1 & 1.5 & 3.1 & 8.7 & 0.4 & 2.0 & 2.8 & 14.8 \\
UCL~\cite{ucl} & 0.3 & 2.0 & 0.3 & 2.5 & 0.1 & 1.8 & 0.2 & 2.2 & 0.0 & 0.1 & 50.5 & 66.5 & 0.1 & 1.7 & 4.8 & 15.7 & 0.4 & 4.2 & 0.6 & 5.8 \\
TextME~\cite{textme} & 0.3 & 2.5 & 0.2 & 2.0 & 0.3 & 1.2 & 0.2 & 2.3 & 0.0 & 0.3 & 56.7 & 73.2 & 0.1 & 2.0 & 2.6 & 10.4 & 0.3 & 2.4 & 1.5 & 9.9 \\
C-MCR~\cite{cmcr} & 0.3 & 3.7 & 0.7 & 2.9 & 0.2 & 2.3 & 0.5 & 2.8 & 0.1 & 0.5 & 73.5 & 82.0 & 0.1 & 1.9 & 7.7 & 22.2 & 0.4 & 3.9 & 2.2 & 12.9 \\
CMCL~\cite{cmcl} & 0.3 & 3.9 & 0.3 & 4.5 & 0.1 & 3.0 & 0.1 & 3.7 & 0.4 & 3.3 & 68.3 & 77.8 & 0.4 & 2.9 & 25.8 & 52.6 & 0.8 & 4.6 & 2.3 & 14.1 \\
Ex-MCR~\cite{exmcr} & 0.3 & 4.5 & 0.2 & 2.8 & 0.4 & 3.9 & 0.2 & 2.7 & 0.3 & 2.4 & 63.6 & 75.6 & 0.4 & 2.8 & 10.8 & 40.5 & 0.4 & 4.5 & 2.2 & 12.4 \\
EmergentBridge~\cite{emergentbridge} & 0.4 & 8.3 & 1.0 & 5.7 & 0.2 & 2.8 & 0.5 & 4.7 & 1.7 & 11.4 & 67.3 & 78.8 & 0.1 & 4.1 & 54.2 & 72.1 & 0.6 & 6.5 & 2.5 & 14.4 \\
Latent Translation~\cite{latenttranslation} & 0.5 & 7.3 & 0.6 & 5.0 & 0.2 & 4.5 & 0.5 & 3.7 & 1.2 & 8.0 & 63.7 & 78.7 & 0.3 & 3.5 & 46.7 & 70.4 & 0.6 & 5.9 & 1.6 & 13.5 \\
MMBind~\cite{mmbind} & 0.7 & 8.0 & 0.4 & 5.4 & 0.3 & 4.1 & 0.6 & 5.9 & 2.9 & 17.5 & 64.5 & 77.5 & 0.4 & 4.5 & \textbf{65.3} & 74.2 & 0.4 & 4.1 & \textbf{3.1} & 14.4 \\
STRUCTURE~\cite{structure} & 1.2 & 8.3 & 0.7 & 5.6 & 0.2 & 4.0 & 0.9 & 4.2 & 1.3 & 9.3 & 63.3 & 77.8 & 0.3 & 4.5 & 46.0 & 68.5 & 1.0 & 8.8 & 2.7 & 16.7 \\
ASIF~\cite{asif} & 1.3 & 8.2 & 0.7 & 5.7 & 0.4 & 4.0 & 0.5 & 3.8 & 1.6 & 9.6 & 52.6 & 71.1 & 0.7 & 4.7 & 40.8 & 64.9 & 1.0 & 6.5 & 1.1 & 10.7 \\
\midrule
\fullrowband{resultshade}
\textbf{PRAM (Ours)} & \textbf{1.4} & \textbf{11.1} & \textbf{1.6} & \textbf{10.9} & \textbf{0.6} & \textbf{5.1} & \textbf{2.6} & \textbf{22.7} & \textbf{46.8} & \textbf{75.1} & \textbf{75.1} & \textbf{82.1} & \textbf{0.8} & \textbf{6.9} & 64.0 & \textbf{76.3} & \textbf{2.2} & \textbf{19.8} & \textbf{3.1} & \textbf{18.7} \\
\bottomrule
\end{tabular}}
\par\vspace{8pt}
{\caption{\textbf{Full LanguageBind Retrieval.} Bidirectional R@1 and R@10 (\%) for all 15 sample-level baselines. The gray row marks Frozen Cosine as the default reference and is excluded from bold ranking; bold marks the best pivot-only result.}
\label{tab:supp-ret-lb}
\fontsize{8.0}{9.6}\selectfont
\setlength{\tabcolsep}{2.0pt}
\renewcommand{\arraystretch}{1.12}
\resizebox{\textwidth}{!}{\begin{tabular}{@{}l*{20}{c}@{}}
\toprule
Method & \multicolumn{2}{c}{\shortstack{BV~\cite{batvision}\\D--Au}} & \multicolumn{2}{c}{\shortstack{VGG~\cite{vggsound}\\Im--Au}} & \multicolumn{2}{c}{\shortstack{NYU~\cite{nyuv2}\\Im--D}} & \multicolumn{2}{c}{\shortstack{TR~\cite{anythermal}\\Im--Th}} & \multicolumn{2}{c}{\shortstack{UCF~\cite{ucf101}\\Im--V}} & \multicolumn{2}{c}{\shortstack{MAV~\cite{mavd}\\Th--Au}} & \multicolumn{2}{c}{\shortstack{TR~\cite{anythermal}\\Th--D}} & \multicolumn{2}{c}{\shortstack{MSR~\cite{msrvtt}\\V--Au}} & \multicolumn{2}{c}{\shortstack{TR~\cite{anythermal}\\V--D}} & \multicolumn{2}{c}{\shortstack{TR~\cite{anythermal}\\V--Th}} \\
\cmidrule(lr){2-3}\cmidrule(lr){4-5}\cmidrule(lr){6-7}\cmidrule(lr){8-9}\cmidrule(lr){10-11}\cmidrule(lr){12-13}\cmidrule(lr){14-15}\cmidrule(lr){16-17}\cmidrule(lr){18-19}\cmidrule(lr){20-21}
 & R@1 & R@10 & R@1 & R@10 & R@1 & R@10 & R@1 & R@10 & R@1 & R@10 & R@1 & R@10 & R@1 & R@10 & R@1 & R@10 & R@1 & R@10 & R@1 & R@10 \\
\midrule
\fullrowband{gray!13}
Frozen Cosine~\cite{languagebind} & 0.0 & 2.7 & 8.9 & 43.6 & 18.7 & 50.7 & 11.2 & 36.9 & 46.8 & 96.6 & 0.2 & 2.0 & 7.5 & 25.2 & 1.2 & 7.0 & 5.6 & 25.0 & 10.8 & 37.2 \\
\midrule
Mean Centering & 0.2 & 1.6 & 11.7 & 51.5 & 36.5 & 71.2 & 17.9 & 46.4 & 53.7 & 98.1 & 0.1 & 1.6 & 10.8 & 32.3 & 2.3 & 11.3 & 8.8 & 30.5 & 12.8 & 42.3 \\
Projection Heads & 0.3 & 4.2 & 7.7 & 47.7 & 1.7 & 13.8 & 4.0 & 22.1 & 26.1 & 83.5 & 0.3 & 3.9 & 1.4 & 16.9 & 2.3 & 16.8 & 2.8 & 18.5 & 2.9 & 19.0 \\
\midrule
TextME~\cite{textme} & 0.0 & 2.0 & 8.2 & 41.3 & 20.3 & 54.7 & 10.2 & 36.1 & 42.7 & 94.5 & 0.1 & 1.7 & 7.7 & 24.9 & 1.2 & 8.4 & 6.2 & 23.7 & 9.5 & 33.4 \\
DBSN~\cite{dbsn} & 0.0 & 2.5 & \textbf{13.2} & \textbf{56.0} & 45.9 & 81.9 & 21.0 & 56.3 & 57.0 & 99.2 & 0.1 & 1.3 & 12.7 & 35.7 & 2.1 & 10.4 & 11.1 & 38.3 & 15.0 & 48.7 \\
Latent Translation~\cite{latenttranslation} & 0.1 & 4.0 & 7.7 & 49.5 & 2.1 & 14.1 & 4.7 & 22.0 & 20.8 & 74.3 & 0.5 & 3.6 & 2.4 & 16.7 & 2.8 & 15.0 & 2.1 & 14.8 & 3.3 & 20.1 \\
I0T~\cite{i0t} & 0.2 & 1.8 & 10.4 & 51.2 & 37.5 & 73.0 & 18.5 & 49.9 & 53.5 & 98.1 & 0.2 & 1.8 & 12.0 & 33.4 & 2.3 & 10.2 & 10.0 & 34.2 & 13.0 & 44.3 \\
CMCL~\cite{cmcl} & 0.2 & 3.8 & 9.8 & 46.3 & 16.6 & 50.8 & 11.2 & 42.3 & 31.0 & 88.0 & 0.5 & 3.7 & 3.2 & 17.3 & 1.3 & 8.5 & 6.1 & 31.1 & 5.7 & 29.5 \\
UCL~\cite{ucl} & 0.3 & 2.5 & 4.0 & 27.4 & 0.8 & 9.1 & 0.7 & 5.6 & 2.1 & 22.6 & 0.3 & 2.3 & 0.6 & 5.2 & 1.1 & 6.2 & 0.5 & 5.6 & 0.7 & 7.3 \\
Ex-MCR~\cite{exmcr} & 0.3 & 3.2 & 8.6 & 53.8 & 1.7 & 14.4 & 2.3 & 15.1 & 11.8 & 57.4 & 0.3 & 3.1 & 0.8 & 10.0 & 2.7 & 15.3 & 1.9 & 13.5 & 1.4 & 13.5 \\
STRUCTURE~\cite{structure} & 0.3 & 4.0 & 7.8 & 49.1 & 3.7 & 19.9 & 3.8 & 21.1 & 22.5 & 80.2 & 0.3 & 4.1 & 2.4 & 15.4 & 3.3 & 18.4 & 2.4 & 17.4 & 1.5 & 20.0 \\
ASIF~\cite{asif} & 0.3 & 4.7 & 6.0 & 48.1 & 1.9 & 12.1 & 2.8 & 20.9 & 4.5 & 38.3 & 0.4 & 3.9 & 2.2 & 13.8 & 2.3 & 13.4 & 2.8 & 18.0 & 2.3 & 19.2 \\
MMBind~\cite{mmbind} & 0.3 & 5.4 & 7.3 & 47.7 & 2.0 & 12.9 & 2.9 & 21.4 & 10.7 & 59.1 & \textbf{0.9} & 5.2 & 1.6 & 16.1 & 2.5 & 17.0 & 1.6 & 17.0 & 2.9 & 18.8 \\
EmergentBridge~\cite{emergentbridge} & 0.3 & 5.7 & 9.0 & 50.5 & 3.4 & 20.2 & 5.2 & 27.9 & 22.9 & 77.9 & 0.4 & 3.0 & 2.6 & 18.3 & 4.0 & 20.2 & 3.0 & 17.3 & 2.2 & 15.1 \\
C-MCR~\cite{cmcr} & 0.5 & 2.8 & 9.2 & 53.2 & 1.9 & 12.8 & 1.3 & 12.0 & 27.2 & 85.4 & 0.1 & 1.6 & 0.9 & 10.0 & 2.8 & 14.4 & 1.0 & 10.2 & 1.5 & 9.9 \\
\midrule
\fullrowband{resultshade}
\textbf{PRAM (Ours)} & \textbf{0.9} & \textbf{6.3} & 8.2 & 53.9 & \textbf{93.2} & \textbf{98.5} & \textbf{46.8} & \textbf{76.1} & \textbf{78.6} & \textbf{99.6} & 0.2 & \textbf{5.3} & \textbf{18.7} & \textbf{54.6} & \textbf{4.8} & \textbf{26.4} & \textbf{15.5} & \textbf{51.8} & \textbf{26.6} & \textbf{61.5} \\
\bottomrule
\end{tabular}}
}
\end{table*}
\subsection{Internal States}
\label{sec:app-internal-states}
For Figure~4 of the main text, we use IB TR depth--thermal with the seed-0 main-table split and fitted models, taking the recording with the most test samples (851 test frames in time order).
Panel (a) jointly embeds the depth and thermal states of the middle 120 consecutive test frames (240 points) with t-SNE~\cite{tsne} (perplexity 10, PCA initialization, seed 0) and highlights frames 7, 52, and 97 of the 120 as D1--D3 and T1--T3.
Frozen Cosine uses the official ImageBind embeddings and MMBind its head outputs in the shared space; both are normalized and compared by cosine distance.
PRAM concatenates the singular coordinates of its 42 parts and the global component, each scaled by $\sqrt{\sigma/(1-\sigma^2)}$ with its singular value $\sigma$, and uses Euclidean distance.
Panel (b) shows heat maps for the middle 40 consecutive frames of the same trajectory, within the 120, with depth queries as rows and thermal candidates as columns; PRAM uses its retrieval score $R$ with reference-bank standardization, Frozen Cosine and MMBind use cosine similarities, and each row is standardized to zero mean and unit standard deviation.
The same-frame thermal sample scores highest in 35 of 40 rows for PRAM, 6 for MMBind, and 2 for Frozen Cosine with depth as the query.

\Needspace*{3\baselineskip}

\section{Ablations and Paired-Data Requirements}
\label{sec:app-ablations}

\subsection{Representation and Score Components}
This subsection defines each ablation in Table~3 of the main text.
Global-only PRAM fits the complete framework on the official global vectors; ``No Global'' removes the separate global component from all score routes, using local parts for correlation composition and covariance mapping and the local-mean predictor for state and class-semantic routes while retaining broadcast descriptors for single-vector modalities.
``No Shrinkage'' sets the cross-part shrinkage intensity $\alpha=0$, so each part keeps its own sample covariance.
Partwise ASIF applies the ASIF adaptation of Sec.~\ref{sec:comparator-details} to the local vectors and selects among its six configurations on held-out modality--pivot pairs.
``No Quadratic'' removes the endpoint-magnitude terms from $\ell_g$.
``No Covariance'' and ``No Pivot Associations'' set $w=0$ and $o=0$, removing the covariance and pivot-association scores; ``No Standardization'' removes the reference-bank transformation.
``No Class Semantics'' removes the class-semantic route by setting $\tau_A=0$.
All remaining feature, split, and decision conventions are fixed.
Pivot-association and class-semantic scores contribute only to classification decisions, so these removals leave retrieval by $R$ unchanged.
Removing an inactive component also retains the full-model result.

Table~\ref{tab:app-local-methods} applies existing methods to PRAM's local vectors with the main-table grids, layers, parts, broadcasting, test rows, metrics, and classification rules.
Pivot-only training uses the two modality--pivot fitting collections with no $A$--$B$ pair; held-out modality--pivot pairs select all choices; Partwise ASIF follows Sec.~\ref{sec:comparator-details}.
LT maps $A,B$ parts to their pivot parts via orthogonal Procrustes or affine ridge (one type for all parts), scored by mean cosine in pivot space.
LT+LI uses these maps with the late interaction of ColBERT~\cite{colbert}, $\tfrac12[\operatorname{mean}_g\max_h c(a_g,b_h)+\operatorname{mean}_h\max_g c(a_g,b_h)]$, with $c$ the mapped-part cosine.
Heads, STRUCTURE, and Partwise MMBind each use one shared model trained on part-level modality--pivot pairs within their row and step caps and scored by mean part cosine.
PRAM has the highest mean in all four blocks: IB R@10 32.9 (best other mean 28.0), LB R@10 53.4 (34.1), IB top-1 50.6 (44.8), and LB top-1 80.2 (72.7).

\begin{table*}[t]
\centering
\caption{\textbf{Full ImageBind Classification.} Top-1 and balanced accuracy (\%) for every baseline. The gray row marks Frozen Cosine as the default reference and is excluded from bold ranking; bold marks the best pivot-only result. $^*$: class names on the text side.}
\label{tab:supp-class-ib}
\fontsize{7.2}{8.8}\selectfont
\setlength{\tabcolsep}{0pt}
\renewcommand{\arraystretch}{0.945}
\begin{tabular*}{\textwidth}{@{\extracolsep{\fill}}l*{8}{c}@{}}
\toprule
Method & \multicolumn{8}{c}{Top-1 Accuracy (\%) $\uparrow$} \\
\cmidrule(lr){2-9}
& \shortstack{BV~\cite{batvision}\\Au--D} & \shortstack{E4D~\cite{ego4d}\\Au--I} & \shortstack{UTD~\cite{utdmhad}\\D--I} & \shortstack{TR~\cite{anythermal}\\D--Th} & \shortstack{VGG$^*$~\cite{vggsound}\\Tx--Au} & \shortstack{E4D~\cite{ego4d}\\Tx--I} & \shortstack{TR$^*$~\cite{anythermal}\\Tx--Th} & \shortstack{NYU$^*$~\cite{nyuv2}\\Tx--D} \\
\midrule
\fullrowband{gray!13}
Frozen Cosine~\cite{imagebind} & 24.7 & 10.4 & 2.8 & 5.8 & 55.9 & 5.8 & 2.0 & 26.6 \\
\midrule
Mean Centering & 30.0 & 9.8 & 3.4 & 7.8 & 56.5 & 7.9 & 2.3 & 29.5 \\
Projection Heads & 69.8 & 24.6 & 16.6 & 48.1 & 53.8 & 23.6 & 44.8 & 25.5 \\
\midrule
UCL~\cite{ucl} & 18.3 & 14.7 & 3.7 & 6.8 & 51.0 & 10.5 & 2.9 & 27.7 \\
TextME~\cite{textme} & 22.6 & 10.9 & 3.1 & 8.5 & 49.4 & 8.7 & 2.9 & 22.0 \\
C-MCR~\cite{cmcr} & 27.1 & 13.4 & 3.5 & 11.3 & 64.0 & 9.5 & 7.2 & 22.3 \\
I0T~\cite{i0t} & 28.2 & 11.1 & 3.3 & 7.0 & 54.9 & 7.6 & 2.0 & 29.5 \\
DBSN~\cite{dbsn} & 28.8 & 11.6 & 3.3 & 7.7 & 58.6 & 6.5 & 1.9 & 27.5 \\
CMCL~\cite{cmcl} & 36.7 & 16.7 & 5.8 & 23.3 & 57.6 & 8.7 & 22.9 & 6.4 \\
Ex-MCR~\cite{exmcr} & 39.5 & 10.3 & 6.7 & 21.2 & 54.2 & 10.6 & 12.7 & 32.0 \\
ASIF~\cite{asif} & 67.2 & 24.3 & 11.5 & 44.0 & 47.6 & 23.6 & 31.4 & 25.9 \\
EmergentBridge~\cite{emergentbridge} & 68.3 & 23.7 & 16.9 & 46.3 & 58.5 & 24.5 & 44.1 & 35.8 \\
STRUCTURE~\cite{structure} & 68.6 & 23.0 & 17.5 & 42.3 & 56.5 & 21.7 & 41.4 & 28.3 \\
Latent Translation~\cite{latenttranslation} & 71.6 & 22.3 & 13.7 & 42.2 & 57.3 & 17.6 & 40.8 & \textbf{41.9} \\
MMBind~\cite{mmbind} & 72.0 & 23.4 & 17.0 & 50.4 & 54.9 & 23.5 & 48.5 & 37.6 \\
\midrule
\fullrowband{resultshade}
\textbf{PRAM (Ours)} & \textbf{74.4} & \textbf{30.6} & \textbf{50.7} & \textbf{51.9} & \textbf{64.7} & \textbf{26.4} & \textbf{65.7} & 40.4 \\
\bottomrule
\end{tabular*}
\par\medskip
\begin{tabular*}{\textwidth}{@{\extracolsep{\fill}}l*{8}{c}@{}}
\toprule
Method & \multicolumn{8}{c}{Balanced Accuracy (\%) $\uparrow$} \\
\cmidrule(lr){2-9}
& \shortstack{BV~\cite{batvision}\\Au--D} & \shortstack{E4D~\cite{ego4d}\\Au--I} & \shortstack{UTD~\cite{utdmhad}\\D--I} & \shortstack{TR~\cite{anythermal}\\D--Th} & \shortstack{VGG$^*$~\cite{vggsound}\\Tx--Au} & \shortstack{E4D~\cite{ego4d}\\Tx--I} & \shortstack{TR$^*$~\cite{anythermal}\\Tx--Th} & \shortstack{NYU$^*$~\cite{nyuv2}\\Tx--D} \\
\midrule
\fullrowband{gray!13}
Frozen Cosine~\cite{imagebind} & 19.2 & 3.7 & 2.8 & 4.3 & 55.9 & 2.6 & 2.0 & 13.5 \\
\midrule
Mean Centering & 22.5 & 3.3 & 3.4 & 4.5 & 56.5 & 3.4 & 2.3 & 15.9 \\
Projection Heads & 66.0 & 11.5 & 16.6 & 34.8 & 53.8 & 10.1 & 45.1 & 15.8 \\
\midrule
UCL~\cite{ucl} & 15.7 & 5.5 & 3.7 & 2.8 & 51.0 & 2.8 & 2.9 & 11.1 \\
TextME~\cite{textme} & 16.7 & 4.2 & 3.1 & 4.0 & 49.4 & 3.9 & 2.9 & 13.7 \\
C-MCR~\cite{cmcr} & 20.2 & 6.2 & 3.5 & 5.0 & 64.0 & 3.1 & 7.1 & 9.1 \\
I0T~\cite{i0t} & 21.8 & 4.8 & 3.3 & 4.4 & 54.9 & 3.6 & 1.9 & 15.3 \\
DBSN~\cite{dbsn} & 21.7 & 6.8 & 3.2 & 4.6 & 58.6 & 2.6 & 2.0 & 13.5 \\
CMCL~\cite{cmcl} & 28.1 & 5.9 & 5.8 & 13.5 & 57.6 & 4.1 & 23.5 & 6.6 \\
Ex-MCR~\cite{exmcr} & 32.0 & 6.1 & 6.8 & 13.0 & 54.2 & 4.5 & 12.5 & 15.9 \\
ASIF~\cite{asif} & 57.6 & 11.1 & 11.6 & 31.9 & 47.6 & 10.8 & 30.8 & 15.5 \\
EmergentBridge~\cite{emergentbridge} & 63.9 & 11.9 & 16.9 & 33.2 & 58.5 & 10.3 & 44.3 & 22.2 \\
STRUCTURE~\cite{structure} & 62.3 & 11.1 & 17.6 & 30.2 & 56.5 & 11.0 & 41.6 & 17.4 \\
Latent Translation~\cite{latenttranslation} & 64.4 & 9.5 & 13.7 & 27.4 & 57.3 & 8.5 & 40.2 & 21.5 \\
MMBind~\cite{mmbind} & \textbf{67.9} & 12.8 & 17.1 & 37.3 & 54.9 & 11.0 & 48.5 & 20.6 \\
\midrule
\fullrowband{resultshade}
\textbf{PRAM (Ours)} & 66.5 & \textbf{18.2} & \textbf{50.8} & \textbf{40.0} & \textbf{64.7} & \textbf{14.2} & \textbf{65.3} & \textbf{31.4} \\
\bottomrule
\end{tabular*}
\vspace{-3.5pt}
\end{table*}
\begin{table*}[t]
\centering
\caption{\textbf{Full LanguageBind Classification.} Top-1 and balanced accuracy (\%) for every baseline. The gray row marks Frozen Cosine as the default reference and is excluded from bold ranking; bold marks the best pivot-only result.}
\label{tab:supp-class-lb}
\fontsize{7.0}{8.6}\selectfont
\setlength{\tabcolsep}{0pt}
\renewcommand{\arraystretch}{0.90}
\begin{tabular*}{\textwidth}{@{\extracolsep{\fill}}l*{9}{c}@{}}
\toprule
Method & \multicolumn{9}{c}{Top-1 Accuracy (\%) $\uparrow$} \\
\cmidrule(lr){2-10}
& \shortstack{BV~\cite{batvision}\\D--Au} & \shortstack{VGG~\cite{vggsound}\\Im--Au} & \shortstack{NYU~\cite{nyuv2}\\Im--D} & \shortstack{TR~\cite{anythermal}\\Im--Th} & \shortstack{UCF~\cite{ucf101}\\Im--V} & \shortstack{TR~\cite{anythermal}\\Th--D} & \shortstack{MSR~\cite{msrvtt}\\V--Au} & \shortstack{TR~\cite{anythermal}\\V--D} & \shortstack{TR~\cite{anythermal}\\V--Th} \\
\midrule
\fullrowband{gray!13}
Frozen Cosine~\cite{languagebind} & 29.9 & 44.8 & 61.0 & 89.9 & 82.8 & 78.1 & 21.9 & 86.0 & 92.3 \\
\midrule
Mean Centering & 37.4 & 51.8 & 65.5 & 92.4 & 86.4 & 86.4 & 25.4 & 89.7 & 91.8 \\
Projection Heads & 68.9 & 60.7 & 65.5 & 90.9 & 90.9 & 91.4 & 37.1 & 91.7 & 91.2 \\
\midrule
TextME~\cite{textme} & 32.7 & 44.0 & 61.9 & 89.9 & 81.3 & 83.8 & 23.0 & 87.3 & 90.3 \\
UCL~\cite{ucl} & 34.0 & 39.5 & 56.2 & 70.0 & 64.2 & 71.5 & 20.9 & 71.8 & 73.0 \\
DBSN~\cite{dbsn} & 38.4 & 54.1 & 60.6 & 89.9 & 85.3 & 83.5 & 24.4 & 86.3 & 89.2 \\
I0T~\cite{i0t} & 38.9 & 51.0 & 65.6 & 91.7 & 86.7 & 86.0 & 25.2 & 88.3 & 90.8 \\
Ex-MCR~\cite{exmcr} & 43.9 & 60.9 & 64.7 & 91.6 & 86.1 & 77.5 & 33.4 & 82.6 & 91.3 \\
C-MCR~\cite{cmcr} & 45.2 & 59.3 & 53.8 & 91.3 & 86.1 & 84.7 & 28.7 & 85.0 & 91.6 \\
CMCL~\cite{cmcl} & 50.6 & 50.5 & 65.8 & 91.7 & 85.8 & 89.6 & 22.5 & 89.3 & 90.8 \\
ASIF~\cite{asif} & 61.1 & 57.0 & 69.1 & 92.7 & 88.0 & 91.2 & 38.3 & 92.1 & 92.8 \\
STRUCTURE~\cite{structure} & 62.4 & 59.7 & 68.9 & 92.8 & 89.7 & \textbf{91.8} & 37.2 & \textbf{93.2} & 92.8 \\
Latent Translation~\cite{latenttranslation} & 68.7 & 63.2 & 72.1 & 93.0 & 89.7 & 91.6 & 33.0 & 91.9 & 92.5 \\
EmergentBridge~\cite{emergentbridge} & 69.0 & 65.9 & 71.8 & 92.3 & 90.7 & 91.7 & 37.7 & 92.3 & 92.3 \\
MMBind~\cite{mmbind} & 70.2 & 64.4 & 70.0 & 92.1 & 91.0 & 90.1 & 38.9 & 90.4 & 91.5 \\
\midrule
\fullrowband{resultshade}
\textbf{PRAM (Ours)} & \textbf{75.8} & \textbf{68.8} & \textbf{72.2} & \textbf{93.4} & \textbf{91.6} & 91.7 & \textbf{42.6} & 92.4 & \textbf{92.9} \\
\bottomrule
\end{tabular*}
\par\medskip
\begin{tabular*}{\textwidth}{@{\extracolsep{\fill}}l*{9}{c}@{}}
\toprule
Method & \multicolumn{9}{c}{Balanced Accuracy (\%) $\uparrow$} \\
\cmidrule(lr){2-10}
& \shortstack{BV~\cite{batvision}\\D--Au} & \shortstack{VGG~\cite{vggsound}\\Im--Au} & \shortstack{NYU~\cite{nyuv2}\\Im--D} & \shortstack{TR~\cite{anythermal}\\Im--Th} & \shortstack{UCF~\cite{ucf101}\\Im--V} & \shortstack{TR~\cite{anythermal}\\Th--D} & \shortstack{MSR~\cite{msrvtt}\\V--Au} & \shortstack{TR~\cite{anythermal}\\V--D} & \shortstack{TR~\cite{anythermal}\\V--Th} \\
\midrule
\fullrowband{gray!13}
Frozen Cosine~\cite{languagebind} & 23.7 & 44.8 & 34.5 & 75.1 & 83.0 & 63.4 & 16.2 & 62.2 & 80.4 \\
\midrule
Mean Centering & 29.2 & 51.8 & 40.8 & 76.4 & 86.5 & 66.2 & 19.8 & 66.5 & 75.5 \\
Projection Heads & 63.9 & 60.7 & 41.0 & 76.5 & 90.8 & 76.4 & 28.6 & 75.5 & 77.7 \\
\midrule
TextME~\cite{textme} & 25.6 & 44.0 & 36.3 & 74.5 & 81.4 & 66.7 & 17.0 & 66.4 & 75.6 \\
UCL~\cite{ucl} & 25.0 & 39.5 & 32.3 & 60.1 & 64.4 & 44.5 & 16.7 & 58.7 & 55.8 \\
DBSN~\cite{dbsn} & 28.4 & 54.1 & 35.2 & 76.3 & 85.3 & 70.7 & 18.7 & 70.6 & 76.1 \\
I0T~\cite{i0t} & 30.6 & 51.0 & 41.0 & 76.0 & 86.8 & 66.7 & 19.1 & 64.9 & 74.8 \\
Ex-MCR~\cite{exmcr} & 33.7 & 60.9 & 43.5 & 72.1 & 86.2 & 59.5 & 26.2 & 65.7 & 72.5 \\
C-MCR~\cite{cmcr} & 31.1 & 59.3 & 30.6 & 67.7 & 86.1 & 64.6 & 22.8 & 69.3 & 69.3 \\
CMCL~\cite{cmcl} & 40.7 & 50.5 & 41.5 & 76.3 & 85.9 & 72.9 & 16.5 & 75.3 & 75.7 \\
ASIF~\cite{asif} & 56.0 & 57.0 & \textbf{47.3} & 76.8 & 88.0 & 71.6 & 31.1 & 72.8 & 76.2 \\
STRUCTURE~\cite{structure} & 59.1 & 59.7 & 47.2 & 78.9 & 89.6 & 76.2 & 28.1 & \textbf{78.6} & 77.5 \\
Latent Translation~\cite{latenttranslation} & 63.7 & 63.2 & 46.4 & 79.3 & 89.6 & 75.5 & 24.9 & 76.6 & \textbf{78.7} \\
EmergentBridge~\cite{emergentbridge} & 63.7 & 65.9 & 46.5 & 77.6 & 90.6 & \textbf{76.6} & 29.0 & 75.0 & 76.7 \\
MMBind~\cite{mmbind} & 64.8 & 64.4 & 45.3 & 76.9 & 90.9 & 74.5 & 29.6 & 75.7 & 73.7 \\
\midrule
\fullrowband{resultshade}
\textbf{PRAM (Ours)} & \textbf{72.5} & \textbf{68.8} & 43.1 & \textbf{80.3} & \textbf{91.5} & 75.4 & \textbf{33.5} & 77.1 & 77.8 \\
\bottomrule
\end{tabular*}
\end{table*}
\subsection{Modality--Pivot Pair Budgets}
Table~\ref{tab:app-budget-scores} gives the numerical results behind Figure~6 of the main text for all ten evaluated configurations across both backbones, of which the figure plots six.
IB MAV D--Th, specified in Sec.~\ref{sec:app-features}, differs from the main-table audio--thermal pair (IB MAV Au--Th).
For a budget fraction $q\in\{0.02,0.05,0.1,0.2,0.35,0.5,0.7,1\}$ and $n$ available pairs, the total subset contains $\max\{50,\operatorname{round}(qn)\}$ samples and is divided between the two observed edges.
Reduced budgets use subset seeds 0 and 1; the full-data point uses one fit.
The reported ranges describe the two subsets, with test samples fixed across budgets.
Table~\ref{tab:app-budget-shrinkage} repeats the 2\% and full-data fits without cross-part shrinkage ($\alpha=0$): the 2\% results fall to 21.7 on IB TR D--Th, 25.0 on its unseen trajectories, and 41.6 on LB TR Th--D, while the mean of the ten full-data results changes from 48.5 to 47.5.
At the 2\% budget, ten subsets (seeds 0--9) are drawn at random from the training samples (individual frames for TartanRGBT), and PRAM and the $\alpha=0$ variant use the same subsets and test samples.
Table~\ref{tab:app-budget-ten} gives the mean drop without shrinkage with a 95\% $t$-interval over subsets and the number of subsets with a drop; all ten subsets show a drop for IB TR D--Th (39.5 [37.8, 41.2]), its unseen trajectories (22.9 [21.9, 23.8]), and LB TR Th--D (49.8 [47.1, 52.5]).
The ten configurations include unseen TartanRGBT trajectories and ImageBind MAVD depth--thermal; LanguageBind TartanRGBT uses an RGB pivot in this study, as in Figure~6 and Table~4(c) of the main text.
IB TR D--Th, its unseen trajectories, LB TR Th--D, and LB UCF Im--V reach at least 96\% of their full-data R@10 with 20\% of pairs.
LB MSR V--Au reaches 21.2 with 50\% of pairs, exceeding the best full-data sample-level baseline of 20.2 in Table~2 of the main text.

\Needspace*{3\baselineskip}

\section{Complete Benchmark Comparisons}
\label{sec:app-comparisons}
\label{sec:full-classification}
Tables~\ref{tab:supp-ret-ib}--\ref{tab:supp-class-lb} report all fifteen sample-level baselines and PRAM under Sec.~\ref{sec:comparator-details}.
Tables~1 and~2 of the main text include the baselines with the highest mean classification top-1 accuracy and retrieval R@10 for each backbone (MMBind for ImageBind classification and retrieval, EmergentBridge for LanguageBind classification, and DBSN for LanguageBind retrieval) and both 2026 baselines (TextME, EmergentBridge).
With all fifteen baselines included, PRAM ranks first or second in all 17 classification settings by top-1 accuracy and all 20 retrieval settings by R@10.
Classification includes balanced accuracy.
Boldface in the comparison tables identifies the strongest result among methods using only modality--pivot pairs for fitting~and~model~selection.
Each table separates the gray Frozen Cosine reference and two generic baselines from the twelve published methods.

\subsection{Five Seeds}
Seeds 0--4 vary the training-edge split, gallery subsample, tie-breaking, four ImageBind layer choices, and all random initializations; test rows stay fixed, seed 0 gives the main tables, and Table~\ref{tab:app-seeds} summarizes backbone-task means.
PRAM ranks first on all five seeds in every benchmark.
\begin{table}[t]
\centering
\caption{\textbf{Five-Seed Summary.} Backbone-task means $\pm$ standard deviations (\%) over seeds 0--4 for PRAM and the strongest baseline among 15 sample-level comparators; \#1 counts seeds where PRAM ranks first.}
\label{tab:app-seeds}
\fontsize{7.0}{8.6}\selectfont
\setlength{\tabcolsep}{1.5pt}
\begin{tabular*}{\linewidth}{@{\extracolsep{\fill}}lcccc@{}}
\toprule
Benchmark & PRAM & Comparator & Mean & \#1 \\
\midrule
IB Classification & \cellcolor{resultshade}\textbf{50.7}$\,\pm\,$0.5 & MMBind~\cite{mmbind} & 41.2$\,\pm\,$0.3 & 5/5 \\
LB Classification & \cellcolor{resultshade}\textbf{79.8}$\,\pm\,$0.4 & EmergentBridge~\cite{emergentbridge} & 77.7$\,\pm\,$0.4 & 5/5 \\
IB Retrieval & \cellcolor{resultshade}\textbf{32.5}$\,\pm\,$0.3 & MMBind~\cite{mmbind} & 21.5$\,\pm\,$0.1 & 5/5 \\
LB Retrieval & \cellcolor{resultshade}\textbf{53.6}$\,\pm\,$0.5 & DBSN~\cite{dbsn} & 43.0$\,\pm\,$0.1 & 5/5 \\
\bottomrule
\end{tabular*}
\end{table}

\section{Computational Cost}
\label{sec:app-cost}
We report measured statistical fitting and test-scoring costs, followed by their scaling with feature dimension and gallery size, with encoder inference excluded.

\subsection{Measured Fitting and Scoring}
Costs in Table~\ref{tab:supp-cost} use one NVIDIA RTX 4070 SUPER, PyTorch 2.6.0, and CUDA 12.4.
Fitting includes relation and scoring statistics and active state-posterior weight fitting.
Scoring computes the full test similarity matrix, including reference-bank standardization.
Feature loading and encoder extraction are separate costs.
Each listed relation has one timing run for fitting and scoring.

\begin{table}[t]
\centering
\caption{\textbf{Measured Fitting and Scoring Cost.} Scoring includes reference-bank standardization and all test pairs. Direction counts include local and global components. Fit GB excludes scoring; including scoring, IB TR D--Th peaks at 3.62\,GB.}
\label{tab:supp-cost}
\fontsize{8.0}{9.6}\selectfont
\renewcommand{\arraystretch}{1.0}
\setlength{\tabcolsep}{2pt}
\begin{tabularx}{\linewidth}{@{}>{\raggedright\arraybackslash}p{.45\linewidth}*{4}{Y}@{}}
\toprule
Relation & \shortstack{Fit\\(s)} & \shortstack{Score\\(s)} & \shortstack{Fit\\GB} & Dirs. \\
\midrule
IB TR~\cite{anythermal} D--Th & 14.04 & 4.06 & 3.22 & 11,008 \\
IB UTD~\cite{utdmhad} D--I & 1.27 & 0.02 & 0.55 & 2,304 \\
LB TR~\cite{anythermal} Th--D & 17.18 & 0.63 & 3.94 & 2,340 \\
LB NYU~\cite{nyuv2} Im--D & 174.98 & 4.74 & 3.94 & 1,430 \\
\bottomrule
\end{tabularx}
\end{table}

For ImageBind TartanRGBT, PRAM retains 11,008 singular coordinates across 42 local components and one global component.
Its fitting takes 14.04 seconds and full $8{,}601\times8{,}601$ test scoring takes 4.06 seconds.
The 42 local components hold 10,752 of these coordinates and the global component holds 256 singular coordinates.

\subsection{Scaling with Dimensions and Gallery Size}
For one component with dimensions $d_A,d_P,d_B$, dense moment accumulation has leading cost
\begin{equation}
\begin{aligned}
\mathcal O\big(&n_A(d_A^2+d_Ad_P+d_P^2)\\
&+n_B(d_B^2+d_Bd_P+d_P^2)\big).
\end{aligned}
\end{equation}
The endpoint eigendecompositions and pivot solve additionally depend cubically on their respective feature dimensions in a dense implementation.
These costs are summed over the retained components; the global component has its own dimensions.
Chunked accumulation controls sample-batch memory while covariance matrices still scale quadratically with feature dimension.

Let $k_\Sigma=\sum_{g=0}^G k_g$.
With cached singular coordinates, the correlation-score matrix for $Q$ queries and $N$ candidates costs $\mathcal O(QNk_\Sigma)$.
Computing both sides' bank moments adds $\mathcal O((Q+N)N_{\rm b}k_\Sigma)$ for bank size $N_{\rm b}$, with reuse for repeated queries or candidates.
Covariance maps and state posteriors add their active feature and state dimensions.
Scoring can be chunked over query--candidate blocks, preserving the defined score while reducing peak memory.
These expressions describe the dense statistical fitting and scoring operations and exclude encoder inference.
On IB TR, peak memory grows with the number of local components, from 3.57\,GB with one and 3.62\,GB with 42 to 8.21\,GB with 168, and fitting time from 1.4 to 39.8 seconds (Table~\ref{tab:supp-grid}), with fixed feature layers.

\Needspace*{3\baselineskip}

\section{Sensing-Scenario Protocols and Results}
\label{sec:application-details}
\label{sec:app-applications}
The three sensing scenarios test the same statistical construction with task-specific frozen features.
Table~\ref{tab:app-applications} reports the multi-metric results in the three sensing scenarios.
Tables~\ref{tab:app-domain-localization} and~\ref{tab:app-domain-wearable} complete the domain-method comparisons for Figure~5 of the main text.
The following subsections specify the sensors, data splits, and supervision used for these localization and wearable comparisons.
\begin{table*}[t]
\centering
\caption{\textbf{Sensing Scenarios.} Localization success and IMU-only classification accuracy (\%). Rows include ASIF, MMBind, and the best domain method for each localization column and each wearable dataset's top-1 accuracy. Robot R@10 is bidirectional exact-frame recall; wearable results are three-seed means. DAv2: Depth Anything V2; CA: contrastive alignment; CZU: CZU-MHAD. Tables~\ref{tab:app-domain-localization} and~\ref{tab:app-domain-wearable} list all domain methods on the main-paper metrics.}
\label{tab:app-applications}
\fontsize{8.0}{9.6}\selectfont
\setlength{\tabcolsep}{2.4pt}
\renewcommand{\arraystretch}{0.94}
\resizebox{\textwidth}{!}{\begin{tabular}{@{}l*{16}{c}@{}}
\toprule
 & \multicolumn{4}{c}{Robot: MS2~\cite{ms2}} & \multicolumn{4}{c}{Robot: OBV~\cite{obv}} & \multicolumn{4}{c}{Vehicle: ViViD++~\cite{vivid} $\to$ MS2~\cite{ms2}} & \multicolumn{4}{c}{Wearable} \\
\multirow{2}{*}{Method} & \multirow{2}{*}{5\,m} & \multirow{2}{*}{15\,m} & \multirow{2}{*}{25\,m} & \multirow{2}{*}{R@10} & \multirow{2}{*}{0.25\,m} & \multirow{2}{*}{0.5\,m} & \multirow{2}{*}{1\,m} & \multirow{2}{*}{R@10} & \multicolumn{3}{c}{Map} & \multirow{2}{*}{\shortstack{Traversal\\5\,m}} & \multicolumn{2}{c}{UTD~\cite{utdmhad}} & \multicolumn{2}{c}{CZU~\cite{czumhad}} \\
 &  &  &  &  &  &  &  &  & 5\,m & 15\,m & 25\,m &  & Top-1 & Top-3 & Top-1 & Top-3 \\
\midrule
ASIF~\cite{asif} & 13.3 & 24.0 & 30.6 & 4.0 & 6.2 & 12.9 & 24.5 & 7.0 & 16.9 & 36.0 & 46.7 & 8.3 & 31.0 & 54.7 & 61.5 & 82.2 \\
MMBind~\cite{mmbind} & 5.7 & 12.6 & 19.1 & 5.1 & 8.0 & 14.5 & 23.3 & 6.7 & 7.9 & 19.0 & 25.1 & 2.5 & 63.6 & 90.3 & 70.1 & 94.0 \\
LIP-Loc~\cite{liploc} & 19.5 & 35.9 & 42.9 & 13.6 & 7.0 & 13.3 & 23.3 & 9.4 & 4.6 & 7.5 & 10.6 & 1.3 & -- & -- & -- & -- \\
ThermalGen~\cite{thermalgen} & 8.6 & 16.9 & 24.2 & 4.6 & 3.4 & 10.0 & 17.7 & 2.9 & 9.6 & 20.1 & 25.1 & 3.5 & -- & -- & -- & -- \\
DAv2 + AnyLoc~\cite{depthanythingv2,anyloc} & 16.0 & 27.3 & 33.4 & 9.8 & 15.7 & 32.3 & 48.2 & 17.2 & 1.4 & 2.5 & 3.2 & 0.4 & -- & -- & -- & -- \\
MINIMA~\cite{minima} & 33.4 & 46.6 & 51.9 & 12.4 & 23.1 & 34.1 & 47.6 & 19.1 & 2.6 & 3.4 & 4.2 & 1.8 & -- & -- & -- & -- \\
C3T~\cite{c3t} & -- & -- & -- & -- & -- & -- & -- & -- & -- & -- & -- & -- & 52.7 & 82.6 & 73.9 & 93.7 \\
CA~\cite{c3t} & -- & -- & -- & -- & -- & -- & -- & -- & -- & -- & -- & -- & 44.2 & 69.8 & 85.9 & 97.1 \\
\fullrowband{resultshade}
\multicolumn{1}{l}{\textbf{PRAM (Ours)}} & \textbf{55.3} & \textbf{74.6} & \textbf{84.8} & \textbf{51.9} & \textbf{31.9} & \textbf{49.0} & \textbf{64.3} & \textbf{35.9} & \textbf{26.2} & \textbf{43.9} & \textbf{50.3} & \textbf{15.0} & \textbf{80.6} & \textbf{95.0} & \textbf{90.8} & \textbf{99.4} \\
\bottomrule
\end{tabular}}
\end{table*}

\subsection{Domain-Method Comparisons}
The localization comparison covers seven method families: MINIMA~\cite{minima}, Depth Anything V2 (DAv2)~\cite{depthanythingv2} with AnyLoc~\cite{anyloc}, LIP-Loc~\cite{liploc}, ThermalGen~\cite{thermalgen}, DINOv2-G~\cite{dinov2} with AnyLoc, Omnivorous-DINO~\cite{omnivorous_dino}, and LC$^2$~\cite{lc2}.
The depth-estimation cascade and DINOv2-G each choose between two fields of view for AnyLoc-VLAD by R@10 on held-out pairs; MINIMA reranks the cascade's shortlist.
The Omnivorous comparison combines AnyThermal with the released Omnivorous-DINO encoder.
LC$^2$ is evaluated with released weights and with adaptation on the depth--RGB edge.
Table~\ref{tab:app-domain-localization} reports the localization metrics plotted in Figure~5 of the main text.

The wearable action recognition comparison covers ten methods: C3T~\cite{c3t}, its contrastive-alignment (CA) and student--teacher (ST) baselines, IMU-Video-OOD-HAR (IMU-Video-OOD)~\cite{imu_video_ood}, PRIMUS~\cite{primus}, Babel~\cite{babel}, COMODO~\cite{comodo}, IMU2CLIP~\cite{imu2clip}, AnyMo~\cite{anymo}, and UniMTS~\cite{unimts}.
C3T has no public code.
We reimplement C3T, CA, and ST from the paper's description using the same splits, seeds 1--3, and test clips as PRAM.
PRIMUS is evaluated both frozen and after continued paired pretraining.
PRIMUS, IMU2CLIP, COMODO, IMU-Video-OOD, and Babel use video class prototypes for the label-free IMU decision.
IMU-Video-OOD, ST, continued PRIMUS, COMODO, and IMU2CLIP train on unlabeled video--IMU pairs and, like Babel, select epochs on held-out pairs, never test clips; Babel realigns its IMU branch with paired data, and its pretraining includes UTD-MHAD IMU data.
AnyMo and UniMTS use their text-based zero-shot decisions and their documented sensor-placement conventions.
Figure~5 of the main text shows the four highest-scoring domain methods in each panel, ranked by mean score on MS2 at 5\,m, OBV at 0.25\,m, and the vehicle map condition at 5\,m for panel (a), and by mean IMU-only test accuracy on UTD and CZU for panel (b).
\begin{table}[t]
\centering
\caption{\textbf{Localization Domain Methods.} Success rates (\%) of seven localization domain methods for the metrics in Figure~5 of the main text: MS2 and vehicle map at 5\,m, OBV at 0.25\,m. Released and fine-tuned LC$^2$ variants are reported separately.}
\label{tab:app-domain-localization}
\fontsize{7.4}{9.0}\selectfont
\setlength{\tabcolsep}{2.5pt}
\resizebox{\linewidth}{!}{\begin{tabular}{@{}l*{3}{c}@{}}
\toprule
Method & \shortstack{MS2~\cite{ms2}\\(5\,m)} & \shortstack{OBV~\cite{obv}\\(0.25\,m)} & \shortstack{Vehicle~\cite{vivid,ms2}\\(5\,m)} \\
\midrule
MINIMA~\cite{minima} & 33.4 & 23.1 & 2.6 \\
DAv2 + AnyLoc~\cite{depthanythingv2,anyloc} & 16.0 & 15.7 & 1.4 \\
LIP-Loc~\cite{liploc} & 19.5 & 7.0 & 4.6 \\
ThermalGen~\cite{thermalgen} & 8.6 & 3.4 & 9.6 \\
DINOv2-G + AnyLoc~\cite{dinov2,anyloc} & 6.5 & 11.0 & 0.2 \\
Omnivorous-DINO~\cite{omnivorous_dino} & 2.5 & 7.0 & 0.2 \\
LC$^2$ (Released)~\cite{lc2} & 3.0 & 3.6 & 0.1 \\
LC$^2$ (Fine-Tuned)~\cite{lc2} & 10.0 & 2.4 & 0.8 \\
\midrule
\fullrowband{resultshade}
\textbf{PRAM (Ours)} & \textbf{55.3} & \textbf{31.9} & \textbf{26.2} \\
\bottomrule
\end{tabular}}
\end{table}

\begin{table}[t]
\centering
\caption{\textbf{Wearable Domain Methods.} Mean top-1 accuracy (\%) $\pm$ standard deviation over seeds 1--3 on UTD and CZU for ten wearable action recognition methods; PRIMUS appears both frozen and with continued pretraining.}
\label{tab:app-domain-wearable}
\fontsize{8.0}{9.6}\selectfont
\renewcommand{\arraystretch}{1.0}
\setlength{\tabcolsep}{3pt}
\begin{tabularx}{\linewidth}{@{}>{\raggedright\arraybackslash}p{.43\linewidth}*{2}{Y}@{}}
\toprule
Method & UTD~\cite{utdmhad} & CZU~\cite{czumhad} \\
\midrule
CA~\cite{c3t} & 44.2$\,\pm\,$7.6 & 85.9$\,\pm\,$4.0 \\
C3T~\cite{c3t} & 52.7$\,\pm\,$11.9 & 73.9$\,\pm\,$10.0 \\
IMU-Video-OOD~\cite{imu_video_ood} & 45.3$\,\pm\,$5.3 & 67.2$\,\pm\,$7.9 \\
ST~\cite{c3t} & 45.7$\,\pm\,$2.9 & 49.1$\,\pm\,$0.9 \\
PRIMUS (Frozen)~\cite{primus} & 5.0$\,\pm\,$2.7 & 7.2$\,\pm\,$2.6 \\
PRIMUS (Continued)~\cite{primus} & 29.5$\,\pm\,$9.7 & 49.1$\,\pm\,$4.8 \\
Babel~\cite{babel} & 27.1$\,\pm\,$3.7 & 48.0$\,\pm\,$4.1 \\
COMODO~\cite{comodo} & 28.7$\,\pm\,$4.7 & 45.7$\,\pm\,$7.1 \\
IMU2CLIP~\cite{imu2clip} & 25.2$\,\pm\,$1.3 & 35.9$\,\pm\,$2.8 \\
AnyMo~\cite{anymo} & 14.3$\,\pm\,$1.3 & 20.7$\,\pm\,$0.9 \\
UniMTS~\cite{unimts} & 15.5$\,\pm\,$1.8 & 11.8$\,\pm\,$1.3 \\
\midrule
\fullrowband{resultshade}
\textbf{PRAM (Ours)} & \textbf{80.6}$\,\pm\,$4.7 & \textbf{90.8}$\,\pm\,$3.3 \\
\bottomrule
\end{tabularx}
\end{table}

\subsection{Robot Localization}
MS2~\cite{ms2} uses 979 test samples and projected LiDAR depth.
OBV~\cite{obv} uses 502 dark-sequence samples, with separate bright-sequence halves for fitting.
The thermal query's retrieved depth sample is counted as localized if its pose lies within the specified distance.
Generic comparison methods use the same AnyThermal and DINOv2-B features as the corresponding PRAM configuration.
LIP-Loc is trained through RGB pairs; ThermalGen synthesizes thermal data from the depth--RGB edge for downstream retrieval training.
MINIMA reranks the top 20 candidates from a depth-estimation and place-recognition cascade.
Its bidirectional exact-frame recall uses the cascade scores for the reverse depth-to-thermal direction.

\subsection{Cross-Device Vehicle Localization}
The queries are 851 nighttime thermal frames from ViViD++~\cite{vivid} campus\_night.
The query vehicle's RGB--thermal edge contains 3,555 daytime frames from campus\_day1.
The map condition uses 3,474 MS2 depth-map frames and a depth--RGB edge of 10,421 moving frames from that mapping traversal.
Thus, map samples are represented in this edge.
The separate-traversal condition uses a 3,365-frame map from another traversal of the same route, with map frames excluded from the fitting edge.
Approximately 69\% of nighttime queries have a map sample within 5\,m, and 88\% within 15\,m.
The corresponding random localization rates are 0.11\% and 0.43\%.
ASIF and LC$^2$ configurations and LIP-Loc loss variants are selected by R@10 on held-out pairs.
The initial training adaptations of ThermalGen and fine-tuned LC$^2$ subsample every third depth--RGB edge frame (giving 9.6\% and 0.6\% map-condition 5\,m success, respectively), while LIP-Loc uses all edge frames.
Retraining ThermalGen and fine-tuned LC$^2$ on all edge frames gives 4.4\% and 0.8\%; Table~\ref{tab:app-domain-localization} reports each baseline's stronger regime (9.6\% and 0.8\%), favoring the comparators.

\subsection{Wearable Action Recognition}
The C3T~\cite{c3t} protocol provides labeled video and unlabeled paired video--IMU samples, with only IMU supplied at test time.
PRAM connects category-indicator vectors to IMU through the video pivot.
The video representation is chosen from ImageBind frames and the ResNet18 representation used by C3T.
The IMU candidates are ImageBind, PRIMUS~\cite{primus}, and the AnyMo graph encoder~\cite{anymo}, selected using the paired training edges without IMU labels.
UTD-MHAD selects ImageBind video with AnyMo IMU; CZU-MHAD~\cite{czumhad} selects ResNet18 video with ImageBind IMU.
ASIF and MMBind use the same video and IMU features as PRAM on each dataset.
ASIF selects its configuration for each seed by R@10 on held-out pivot pairs, without test data or labeled validation data.
For these PRAM features, the AnyMo latent sequence is averaged into eight time bins.
Every UTD clip uses the right-hand sensor node, independently of its action label.
CZU depth-video preprocessing uses the dataset's foreground crop for all video-based comparators; its ten inertial sensors are concatenated with side identities preserved.
Table~\ref{tab:app-applications} reports top-1 and top-3 means over seeds 1--3.
Table~\ref{tab:app-domain-wearable} adds the standard deviations over seeds; PRAM has the highest top-1 accuracy in each of the three seeds on both datasets.
For each of seeds 1--3, the 861 UTD-MHAD clips are split with class stratification, pooling the eight subjects, into 86 IMU-only test clips, 86 IMU-only validation clips, 344 labeled videos, and 345 unlabeled video--IMU pairs, following the 10/10/40/40 fractions of C3T~\cite{c3t}.
On these splits, PRAM with ImageBind instead of AnyMo IMU features reaches 79.5$\,\pm\,$1.8\% top-1, versus C3T's reported 62.5\% on its own splits.

\section{Limitations}
PRAM uses registered spatial cells or synchronized time segments as parts; a single global vector is broadcast across the other modality's parts.
Relations use covariances of features from frozen encoders.
Extensions include learned or adaptive correspondence for unregistered inputs and integration with fine-tuned encoders.

\FloatBarrier
\phantomsection
\addcontentsline{toc}{section}{References}
{\footnotesize\setlength{\bibsep}{1.8pt plus 1pt minus 1pt}

\end{document}